%% file: main.tex
\documentclass{article}
\pdfoutput=1

\PassOptionsToPackage{numbers, compress, sort}{natbib}

\usepackage[preprint]{neurips_2022}

\usepackage[utf8]{inputenc} %
\usepackage[T1]{fontenc}    %
\usepackage[hidelinks]{hyperref}       %
\usepackage{url}            %
\usepackage{booktabs}       %
\usepackage{amsfonts}       %
\usepackage{nicefrac}       %
\usepackage{microtype}      %
\usepackage{xcolor}         %
\usepackage{graphicx}
\usepackage{caption}
\usepackage{multirow}
\usepackage{algorithm}
\usepackage{algpseudocode}
\usepackage{subcaption}
\usepackage[acronym]{glossaries}

\usepackage{amsmath}
\usepackage{amssymb}
\usepackage{mathtools}
\usepackage{amsthm}
\usepackage{amsfonts}
\usepackage{bm}
\usepackage{enumitem}
\usepackage{dsfont}

\graphicspath{ {./tex/figures/} }

\input{math_commands.tex}

\DeclarePairedDelimiterX{\infdivx}[2]{(}{)}{%
  #1\;\delimsize\|\;#2%
}
\newcommand{\infdiv}{\KL\infdivx}
\theoremstyle{plain}
\newtheorem{theorem}{Theorem}[section]

\theoremstyle{definition}

\theoremstyle{remark}

\newcommand{\mn}{PSLD }
\newcommand{\nsmn}{PSLD}
\newcommand{\fmn}{Phase Space Langevin Diffusion}

\newacronym{SGM}{SGM}{Score-based Generative Model}
\newacronym{\nsmn}{\nsmn}{\fmn}
\newacronym{\mn}{\mn}{\fmn}
\newacronym{CLD}{CLD}{Critically Damped Langevin Diffusion}

\title{A Complete Recipe for Diffusion Generative Models}

\author{%
  Kushagra Pandey
  \\
  Department of Computer Science\\
  University of California, Irvine\\
  \texttt{pandeyk1@uci.edu} \\
  \And
  Stephan Mandt \\
  Department of Computer Science \\
  University of California, Irvine \\
  \texttt{mandt@uci.edu} \\
}
\begin{document}

\maketitle

\begin{abstract}
\glspl{SGM} have demonstrated exceptional synthesis outcomes across various tasks. However, the current design landscape of the forward diffusion process remains largely untapped and often relies on physical heuristics or simplifying assumptions. Utilizing insights from the development of scalable Bayesian posterior samplers, we present a complete recipe for formulating forward processes in \glspl{SGM}, ensuring convergence to the desired target distribution. Our approach reveals that several existing \glspl{SGM} can be seen as specific manifestations of our framework. Building upon this method, we introduce \gls{\nsmn}, which relies on score-based modeling within an augmented space enriched by auxiliary variables akin to physical phase space. Empirical results exhibit the superior sample quality and improved speed-quality trade-off of \gls{\nsmn} compared to various competing approaches on established image synthesis benchmarks. Remarkably, \gls{\nsmn} achieves sample quality akin to state-of-the-art \glspl{SGM} (FID: \textbf{2.10} for unconditional CIFAR-10 generation). Lastly, we demonstrate the applicability of \gls{\nsmn} in conditional synthesis using pre-trained score networks, offering an appealing alternative as an \gls{SGM} backbone for future advancements. Code and model checkpoints can be accessed at \url{https://github.com/mandt-lab/PSLD}.
\end{abstract}

\input{tex/introduction.tex}
\input{tex/recipe.tex}
\input{tex/psld.tex}
\input{tex/exps.tex}

\input{tex/related.tex}
\input{tex/conclusion.tex}
\input{tex/additional}
\bibliography{ref2}
\bibliographystyle{unsrtnat}

\newpage
\appendix

\input{tex/appendix/App_A.tex}
\input{tex/appendix/App_B.tex}
\input{tex/appendix/App_C.tex}
\input{tex/appendix/App_D.tex}

\end{document}

%% file: math_commands.tex
\def\eqref#1{equation~\ref{#1}}

\def\1{\bm{1}}

\def\eps{{\epsilon}}

\def\rvf{{\mathbf{f}}}

\def\rvu{{\mathbf{i}}}

\def\rvm{{\mathbf{m}}}

\def\rvu{{\mathbf{u}}}

\def\rvw{{\mathbf{w}}}
\def\rvx{{\mathbf{x}}}
\def\rvy{{\mathbf{y}}}
\def\rvz{{\mathbf{z}}}

\def\brvx{\bar {\mathbf{x}}}
\def\brvm{\bar {\mathbf{m}}}
\def\brvz{\bar {\mathbf{z}}}

\def\lop{\hat{\mathcal{L}}}

\def\vzero{{\bm{0}}}

\def\vmu{{\bm{\mu}}}
\def\vtheta{{\bm{\theta}}}

\def\vf{{\bm{f}}}

\def\vh{{\bm{h}}}

\def\vk{{\bm{k}}}

\def\vm{{\bm{m}}}

\def\vs{{\bm{s}}}

\def\vx{{\bm{x}}}

\def\vz{{\bm{z}}}

\def\mD{{\bm{D}}}

\def\mF{{\bm{F}}}
\def\mG{{\bm{G}}}

\def\mI{{\bm{I}}}

\def\mL{{\bm{L}}}

\def\mQ{{\bm{Q}}}

\def\mSigma{{\bm{\Sigma}}}

\DeclareMathAlphabet{\mathsfit}{\encodingdefault}{\sfdefault}{m}{sl}
\SetMathAlphabet{\mathsfit}{bold}{\encodingdefault}{\sfdefault}{bx}{n}

\def\gB{{\mathcal{B}}}

\def\gN{{\mathcal{N}}}

\def\gU{{\mathcal{U}}}

\newcommand{\E}{\mathbb{E}}

\newcommand{\R}{\mathbb{R}}

\newcommand{\KL}{D_{\mathrm{KL}}}

%% file: tex/introduction.tex
\section{Introduction}
\label{sec:introduction}
\noindent
\acrlongpl{SGM} \citep{sohl2015deep, song2019generative, ho2020denoising, songscore} are a class of explicit-likelihood based generative models that have recently demonstrated impressive performance on various synthesis benchmarks, such as image generation \citep{dhariwal2021diffusion, ho2022cascaded, rombach2022high, https://doi.org/10.48550/arxiv.2204.06125, sahariaphotorealistic}, video synthesis \citep{https://doi.org/10.48550/arxiv.2203.09481, hovideo} and 3D shape generation \citep{luo2021diffusion, zhou20213d}. \glspl{SGM} employ a forward stochastic process to add noise to data incrementally, transforming the data-generating distribution to a tractable prior distribution that enables sampling. Subsequently, a learnable reverse process transforms the prior distribution back to the data distribution using a parametric estimator of the gradient field of the log probability density of the data (a.k.a score).

However, a principled framework for extending the current design space of diffusion processes is still missing. Although some studies have proposed augmenting the forward diffusion process with auxiliary variables \citep{dockhornscore} to improve sample quality, their design is primarily motivated by physical intuition and non-obvious how to generalize. Therefore, a principled framework is required to explore the space of possible diffusion processes better.

\begin{figure}[t]
    \centering
    \includegraphics[width=1.0\linewidth]{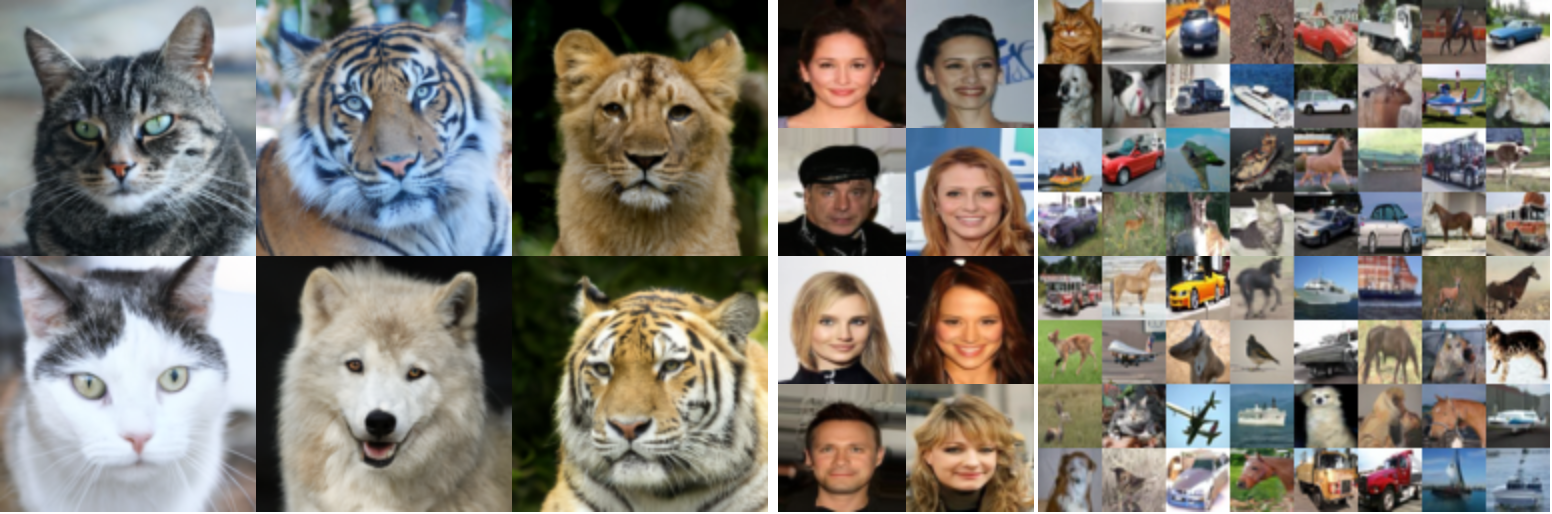}
    \caption{Unconditional \mn generated samples. AFHQv2 128 x 128 (Top), CelebA-64 (Bottom Left, FID=2.01) and CIFAR-10 (Bottom Right, FID=2.10)}
    \label{fig:main_0}
\end{figure}

In this work, we propose a \textit{complete} recipe for the design of diffusion processes, motivated by the design of stochastic gradient MCMC samplers \citep{10.5555/3104482.3104568,chen2014stochastic, ma2015complete}. Our recipe leads to a flexible parameterization of the forward diffusion process without requiring physical intuition. Moreover, under the proposed parameterization, the forward process is guaranteed to converge to a prior distribution of interest. We show that several existing \glspl{SGM} can be cast under our diffusion process parameterization. Furthermore, using our proposed recipe, we introduce \gls{\nsmn},
a novel \gls{SGM} which performs diffusion in the joint space of data and auxiliary variables. We demonstrate that \gls{\nsmn} generalizes Critically Damped Langevin Diffusion (CLD) \citep{dockhornscore} and outperforms existing baselines on several empirical settings on standard image synthesis benchmarks such as CIFAR-10 \citep{krizhevsky2009learning} and CelebA-64 \citep{liu2015faceattributes}. More specifically, we make the following theoretical and empirical contributions:

\begin{enumerate}
    \item \textbf{A Complete Recipe for SGM Design:} We propose a specific parameterization of the forward process, guaranteed to converge asymptotically to a desired stationary ``prior'' distribution. The proposed recipe is \textit{complete} in the sense that it subsumes all possible Markovian stochastic processes which converge to this distribution. We show that several existing \glspl{SGM} \citep{songscore, dockhornscore} can be cast as specific instantiations 
    of our recipe.

    \item \textbf{\fmn (PSLD):} To exemplify the proposed diffusion parameterization concretely, we propose \gls{\nsmn}: a novel \gls{SGM} which performs diffusion in the phase space by adding noise in \textbf{both} data and the momentum space.

    \item \textbf{Superior Sample Quality and Speed-Quality Tradeoffs:} Using ablation experiments on standard image synthesis benchmarks like CIFAR-10 and CelebA-64, we demonstrate the benefits of adding stochastic noise in both the data and the momentum space on overall sample quality and the speed quality trade-offs associated with \gls{\nsmn}. Furthermore, using similar score network architectures, our proposed method outperforms existing diffusion baselines on both criteria across different sampler settings.

    \item \textbf{State-of-the-Art Sample Quality:} We show that \gls{\nsmn} outperforms competing baselines and achieves competitive perceptual sample quality to other state-of-the-art methods. Our model achieves an FID \citep{heusel2017gans} score of \textbf{2.10}, an IS score \citep{salimans2016improved} of \textbf{9.93} on unconditional CIFAR-10 and an FID score of \textbf{2.01} on CelebA-64.

    \item \textbf{Conditional synthesis:} We show that pre-trained unconditional \gls{\nsmn} models can be used for conditional synthesis tasks like class-conditional generation and image inpainting.
\end{enumerate}
Overall, given the superior performance of \gls{\nsmn} on several tasks, we present an attractive alternative to existing \gls{SGM} backbones for further development. We organize the rest of our work as follows: Section \ref{sec:recipe} presents some background on \glspl{SGM} and our proposed \textit{recipe} for \gls{SGM} design. Section \ref{sec:es3sde} presents the construction of our novel \gls{\nsmn} model. Section \ref{sec:exps} presents our empirical findings. Lastly, Section \ref{sec:related} compares our proposed contributions to several existing works while we present some directions for future work in Section \ref{sec:conclusion}.

%% file: tex/recipe.tex
\section{A Complete Recipe for SGM Design}
\label{sec:recipe}

\subsection{Background}
\label{sec:background}
\noindent
Consider the following forward process SDE for converting data $\rvx_t \in \R^d$ to noise,
\begin{equation*}
    d\rvx_t = \vf(\rvx_t, t) \, dt + \mG(t) \, d\rvw_t, \quad t \in [0, T],
\end{equation*}
with continuous time variable $t \in [0, T]$, a standard Wiener process $\rvw_t$, drift coefficient $\vf \colon \R^d \times [0, T] \to \R^d$, and diffusion coefficient $\mG \colon [0, T] \to \R^{d \times d}$.
Given this forward process, the corresponding reverse-time diffusion process \cite{songscore, ANDERSON1982313} that generates data from noise can be specified as follows,
\begin{equation} \label{eq:reverse_time_diffusion}
    d \rvx_t = \left[\vf(\rvx_t, t) - \mG(t) \mG(t)^\top \nabla_{\rvx_t} \log p_t(\rvx_t)\right] \, dt + \mG(t) d\bar \rvw_t,
\end{equation}
Given an estimate of the \textit{score} $\nabla_{\rvx_t} \log p_t(\rvx_t)$ of the marginal distribution over $\rvx_t$ at time $t$, the reverse SDE can then be simulated to recover the original data samples from noise. In practice, the score is intractable to compute and is approximated using a parametric estimator $s_{\theta}(\rvx_t, t)$, trained using denoising score matching \cite{song2019generative, ho2020denoising, songscore, 6795935}:
\begin{equation*}
\min_\vtheta\E_{t}\E_{p(\rvx_0)}\E_{p_t(\rvx_t|\rvx_0)}\left[\lambda(t)\Vert\mathbf{s}_\vtheta(\rvx_t,t) - \nabla_{\rvx_t} \log p_{t}(\rvx_t|\rvx_0)\Vert_2^2\right]. 
\end{equation*}
Above, the time $t$ is usually sampled from a uniform distribution $\mathcal{U}(0, T)$. Given an appropriate choice of $\vf$ and $\mG$, the perturbation kernel $p(\rvx_t|\rvx_0)$ can frequently be computed analytically (e.g., it is typically Gaussian). Consequently, samples $\rvx_t$ can be generated in constant time, allowing for fast stochastic gradient updates. The choice of the weighting schedule $\lambda(t)$ plays an essential role during training and can be selected to optimize for likelihood \citep{song2021maximum} or sample quality \citep{songscore}. The forward SDE asymptotically converges to an equilibrium distribution (usually a standard isotropic Gaussian) which can be used as a prior to initialize the reverse SDE, which can be simulated using numerical solvers.

\subsection{A General Recipe for Constructing Stochastic Forward Processes}
\label{subsec:2_3}
\noindent
As has been shown in the MCMC literature \citep{2011, chen2014stochastic}, it is often beneficial to extend the sampling space
into an \emph{augmented} space according to $\rvz = [\rvx, \rvm]^T \in \mathbb{R}^{d_z}$,
where $\rvx \in \mathbb{R}^{d_x}$ is the original state space variable and $\rvm \in \mathbb{R}^{d_m}$ corresponds to some additional auxiliary dimensions. Simulating the dynamics of the variable $\rvz$ may have desirable properties, such as faster mixing. Inspired by the naming conventions in statistical physics, we call $\rvx$ the \emph{position} variable and $\rvm$ the \emph{momentum} variable. Accordingly, we denote
their joint space as \emph{augmented space} (or \emph{phase space if $\rvx$ and $\rvm$} have equal dimensions). Note that our notation also captures the scenario where $\rvm$ is absent (zero-dimensional). 
We now consider the following form of the stochastic process:
\begin{equation}
    d\rvz = \vf(\rvz) dt + \sqrt{2\mD(\rvz)}d\rvw_t,
    \label{eqn:3_1}
\end{equation}
with drift term $\vf(\rvz) \in \mathbb{R}^{d_z}$ and diffusion coefficient $\mD(\rvz) \in \mathbb{R}^{d_z \times d_z}$. 
We assume a desired stationary state distribution $p_s(\rvz)$ specified as
\begin{equation*}
    p_s(\rvz) \propto \exp(-H(\rvz)),
\end{equation*}
\begin{equation}
    H(\rvz) = H(\rvx, \rvm) = U(\rvx) + \frac{\rvm^T M^{-1} \rvm}{2}
    \label{eqn:invariant_1},
\end{equation}
where $H$ represents the Hamiltonian associated with $p_s(\rvz)$. The first term in $H(\rvz)$ represents the potential energy $U(\rvx)$ associated with the configuration $\rvx$ while the second term represents the kinetic energy associated with the auxiliary (or momentum) variables $\rvm$ and mass matrix $M\mI_{d_m}$. In the context of Bayesian inference, \citep{ma2015complete} propose a framework to elucidate the design space of possible MCMC samplers that sample from $p_s(\rvz)$. In this framework, the drift $\vf(\rvz)$ can be parameterized as
\begin{equation}
    \vf(\rvz) = -(\mD(\rvz) + \mQ(\rvz))\nabla H + \tau(\rvz),
    \label{eqn:3_2}
\end{equation}
\begin{equation*}
\tau_i(\rvz) = \sum_{j=1}^d \frac{\partial}{\partial \rvz_j}\left(\mD_{ij}(\rvz) + \mQ_{ij}(\rvz)\right),
\end{equation*}
where $\mQ(\rvz)$ represents a skew-symmetric curl matrix. Furthermore, the following result holds:
\begin{theorem}[Yin et. al. \citep{Yin_2006}]
For the dynamics defined in Eqn. \ref{eqn:3_1}, if $\vf(\rvz)$ is parameterized as in Eqn. \ref{eqn:3_2} with $\mD(\rvz)$ positive semidefinite and $\mQ(\rvz)$ skew-symmetric, then the distribution $p_s(\rvz) \propto \exp{(-H(\rvz))}$ is a stationary distribution for the dynamics.
\label{theorem:1}
\end{theorem}
\noindent
Theorem \ref{theorem:1} implies that for a specific choice of matrices $\mD(\rvz)$ and $\mQ(\rvz)$, the process defined in Eqn. \ref{eqn:3_1} always asymptotically samples from the target distribution $p_s(\rvz)$. Moreover, \citep{ma2015complete} showed in the context of MCMC that the parameterization defined in Eqn. \ref{eqn:3_2} is \textit{complete} as follows: 

\begin{theorem}
    [Ma et. al. \citep{ma2015complete}] 
    Assume the stochastic process in Eqn. \ref{eqn:3_1} converges to a unique stationary distribution $p_s(\rvz)$. Then, under mild regularity assumptions, there exists a corresponding skew-symmetric matrix $\mQ(\rvz)$, such that $\vf(\rvz)$ assumes the form of Eqn. \ref{eqn:3_2}.
    \label{theorem:2}
\end{theorem}
\noindent
We include the proofs for Theorems \ref{theorem:1} and \ref{theorem:2} in Appendix \ref{app:A_1} for completeness. These results provide a general recipe for designing forward processes in \gls{SGM}s.

For the \gls{SGM} to be a useful forward process, we need it to converge to a simple factorized distribution that serves as the initialization point of the backwards (generative) process. Consequently, we consider the following form of the stationary distribution $p_s(\rvz)$:
\begin{equation}
    p_s(\rvz) = \mathcal{N}(\rvx; \mathbf{0}_{d_x}, \mI_{d_x})\mathcal{N}(\mathbf{0}_{d_m}, M\mI_{d_m}).
    \label{eqn:invariant}
\end{equation}
This form results from setting $U(\rvx) = \frac{\rvx^T\rvx}{2}$ in Eqn. \ref{eqn:invariant_1}.
Therefore, for a positive semidefinite matrix $\mD(\rvz)$ and a skew-symmetric matrix $\mQ(\rvz)$, the most general class of forward processes which lead to an invariant distribution $p_s(\rvz)$ can be specified by substituting the form of $\nabla H(\rvz)$ (corresponding to $p_s(\rvz)$ defined in Eqn. \ref{eqn:invariant})  in Eqn. \ref{eqn:3_2}.
A similar characterization of forward processes has also been explored in a concurrent work by \citep{singhal2023where} in the context of likelihood estimation (see 
Section \ref{sec:related}).

\subsection{Additional constraints on D and Q}
\noindent
Theorems \ref{theorem:1} and \ref{theorem:2} show that the proposed forward process parameterization is complete upon specifying the target distribution  $p_s(\rvz)$ (such as Eqn. \ref{eqn:invariant}). However, we need additional requirements for the resulting generative model and the corresponding training objective to be tractable.  Specifically, when using the denoting score matching objective \citep{6795935}, we require the perturbation kernel $p(\rvz_t|\rvz_0)$ to be computable in closed form. In practice, this restricts our possible choices for $\mD(\rvz)$ and $\mQ(\rvz)$ to constant matrices (i.e., independent of the state variable $\rvz$). Yet, even with this requirement, the framework provides a large design space of models. We provide several examples of existing \gls{SGM}s that can be understood as special cases of our recipe in Appendix \ref{app:A_3}. We stress that training paradigms other than denoting score matching (e.g., such as Sliced Score matching \citep{song2020sliced}) may enable a wider range of possible models with non-constant matrices $D$ and $Q$.

%% file: tex/psld.tex
\section{\fmn}
We next use the proposed recipe to construct a specific SGM with favorable properties.
\label{sec:es3sde}

\subsection{Model Definition}

\noindent
We restrict the family of forward processes considered in this work by constraining $\mD(\rvz)$ and $\mQ(\rvz)$ as constant matrices, i.e., independent of state $\rvz$. Moreover, we assume that $\rvx$ and $\rvm$ have the same dimension d, i.e. $\rvz \in \mathbb{R}^{2d}$. Consequently, the drift $f(\rvz)$ becomes affine in $\rvz$ and the perturbation kernel $p(\rvz_t|\rvz_0)$ can be computed analytically \citep{sarkka2019applied}. Among the possible samplers, we choose a specific form involving $d-$dimensional position and momentum coordinates, $\rvz_t = [\rvx_t, \rvm_t]^T$ where $\rvx_t \in \mathbb{R}^d$, $\rvm_t \in \mathbb{R}^d$.
Our choice for $\mD(\rvz)$ and $\mQ(\rvz)$ is as follows:
\begin{equation}
    \mD \coloneqq \frac{\beta}{2}\left(\begin{pmatrix} \Gamma & 0 \\ 0 & M\nu \end{pmatrix} \otimes \mI_d\right), \;\quad
    \mQ \coloneqq \frac{\beta}{2}\left(\begin{pmatrix} 0 & -1 \\ 1 & 0 \end{pmatrix}\otimes \mI_d\right).
    \label{eqn:choice}
\end{equation}
Above, $\Gamma$, $M$, $\nu$ and $\beta$ are positive scalars. Along with these choices of $\mD$ and $\mQ$, we have $\tau(\rvz) = \bf{0}$. 
The resulting forward process is given by:
\begin{equation}
    d\rvz_t = \vf(\rvz_t)dt + \mG(t)d\rvw_t,
    \label{eqn:4_1}
\end{equation}
\begin{equation*}
\vf(\rvz_t) =  \left(\frac{\beta}{2}\begin{pmatrix} -\Gamma & M^{-1}\\ -1 & - \nu \end{pmatrix} \otimes \mI_d\right) \,\rvz_t, \quad \mG(t) = \sqrt{2D(\rvz_t)} = \begin{pmatrix} \sqrt{\Gamma\beta} &0 \\ 0 &\sqrt{M\nu\beta} \end{pmatrix} \otimes \mI_d.
\end{equation*}

\noindent
We denote the form of the SDE in Eqn. \ref{eqn:4_1} as the \textit{\acrlong{\mn}\,
 (\gls{\nsmn})}. Note that \gls{\nsmn} generalizes \acrlong{CLD} (\gls{CLD}) proposed in \cite{dockhornscore}, which can be obtained by setting $\Gamma = 0$, $\bar{\nu} = M\nu$, and $\bar{\beta} = \frac{\beta}{2}$. Like \gls{CLD}, the parameter $M^{-1}$ couples the data space state $\rvx_t$ with the auxiliary state $\rvm_t$. The parameters $\beta$, $\Gamma$, and $\nu$ control the amount of noise in the forward SDE. Without loss of generality, we use a time-independent $\beta$. However, unlike \gls{CLD} or any physical system, \gls{\nsmn} adds stochastic noise in the data space \emph{in addition} to the noise injected into the momentum component of phase space. While we are not aware of any physical system that displays such behavior, it is a valid stochastic process compatible with our framework. Our experiments reveal the strong benefits of having these two independent noise sources. 

Furthermore, CLD \citep{dockhornscore} proposes setting $\bar{\nu}^2 = 4M$, corresponding to critical damping in a physical system. Under critical damping, an ideal balance is achieved between the oscillatory Hamiltonian dynamics and the noise-injecting Ohrnstein-Uhlenbeck (OU) term, leading to faster convergence to equilibrium.
We generalize this line of argument in Appendix \ref{app:B_1}, where we derive $(\Gamma - \nu)^2 = 4M^{-1}$ as the equivalent condition for critical damping in \gls{\nsmn}. Throughout this work, we choose $\Gamma$, $\nu$, and $M^{-1}$ such that the critical damping condition in \gls{\nsmn} is satisfied.

\subsection{PSLD Training}
\noindent
Since the drift coefficient in \gls{\nsmn} is affine, the perturbation kernel $p(\rvz_t| \rvz_0)$ of \gls{\nsmn} can be computed analytically. We can then use DSM to learn the score function $\vs_{\theta}(\rvz_t, t)$. More specifically, following the derivation in \citep{song2021maximum}, it can be shown that the Maximum-Likelihood (ML) based DSM objective for \gls{\nsmn} can be specified as (Proof in Appendix \ref{app:B_2_1})

\begin{align}
\min_\vtheta\E_{t}\E_{p(\rvz_0)}\E_{p_t(\rvz_t|\rvz_0)}\Big[\Gamma \beta \mathcal{L}_x(\theta, \rvz_t, \rvz_0) + M\nu\beta \mathcal{L}_m(\theta, \rvz_t, \rvz_0)\Big],
\label{eqn:4_1_1}
\end{align}
\begin{equation}
\mathcal{L}_x = \Vert\mathbf{s}_\vtheta(\rvz_t,t)|_{0:d} - \nabla_{\rvx_t} \log p_{t}(\rvz_t|\rvz_0)\Vert_2^2, \quad
\mathcal{L}_m = \Vert\mathbf{s}_\vtheta(\rvz_t,t)|_{d:2d} - \nabla_{\rvm_t} \log p_{t}(\rvz_t|\rvz_0)\Vert_2^2,
\end{equation}
where $\vs_\vtheta(\rvz_t,t)|_{0:d}$ and $\vs_\vtheta(\rvz_t,t)|_{d:2d}$ represent the first and the last $d$ components of the vector $\vs_{\theta}(\rvz_t, t)$ respectively. In the above DSM objective, the perturbation kernel $p(\rvz_t|\rvz_0) = \mathcal{N}(\vmu_t, \mSigma_t)$ is a multivariate Gaussian while $p(\rvz_0) = p(\rvx_0)\mathcal{N}(\rvm_0; 0, M\gamma\mI_d)$, where $p(\rvx_0)$ is the data distribution. In this work, we reformulate the DSM objective in Eqn. \ref{eqn:4_1_1} as follows (also see Appendix \ref{app:B_2_1}):
\begin{equation*}
\min_\vtheta\E_{t}\E_{p(\rvz_0)}\E_{p_t(\rvz_t|\rvz_0)}\Big[\lambda(t)\Vert\mathbf{s}_\vtheta(\rvz_t,t) - \nabla_{\rvz_t} \log p_{t}(\rvz_t|\rvz_0)\Vert_2^2 \Big].
\end{equation*}

Furthermore, due to its gradient variance reduction properties, we instead use the Hybrid Score Matching (HSM) objective \citep{dockhornscore} by marginalizing out the momentum variables $\rvm_0$ as $p(\rvz_t | \rvx_0) = \int p(\rvz_t | \rvx_0, \rvm_0)p(\rvm_0)d\rvm_0$. Since both distributions $p(\rvz_t | \rvx_0, \rvm_0)$ and $p(\rvm_0)$ are Gaussian, $p(\rvz_t | \rvx_0)$ will also be a Gaussian.

\noindent
\textbf{Score Network Parameterization:} Since the perturbation kernel $p(\rvz_t|\rvx_0)$ in the HSM objective is also a multivariate Gaussian, we have $p(\rvz_t|\rvx_0) \sim \gN(\vmu_t, \mSigma_t)$. Furthermore, let $\mSigma_t = \mL_t\mL_t^T$ be the Cholesky factorization of the matrix $\mSigma_t$. We have
\begin{equation}    \nabla_{\rvz_t}\log{p(\rvz_t|\rvx_0)} = -\mSigma_t^{-1}(\rvz_t - \vmu_t) = -\mL_t^{-T}\bm{\epsilon},
\label{eqn:4_1_2}
\end{equation}
\noindent
where $\mL_t^{-T}$ is the transposed inverse of the $\mL_t$ and $\bm{\eps} \sim \mathcal{N}(\vzero_{2d}, \mI_{2d})$. Therefore, we parameterize our score function estimator as $\vs_{\theta}(\vz_t, t) = -\mL_t^{-T}\bm{\eps}_{\theta}(\rvz_t, t)$. Although alternative parameterizations of the score network $s_{\theta}(\rvz_t, t)$ like \textit{mixed score} can be possible \citep{dockhornscore, vahdat2021score, karraselucidating}, we do \textbf{not} explore such parameterizations in this work and leave further exploration to future work. We provide additional details on the score network parameterization in \gls{\nsmn} in Appendix \ref{app:B_2_2} and the analytical form of the perturbation kernel $p(\rvz_t|\rvx_0)$ in Appendix \ref{app:B_3}.\\

\noindent
\textbf{Final Training Objective:} 
Using our score parameterization from Eqn. \ref{eqn:4_1_2} with $\lambda(t) = \frac{1}{\Vert \mL_t^{-T} \Vert_2^2}$, we get the following \textit{epsilon-prediction} form of the HSM objective (See Appendix \ref{app:B_2_3} for a complete derivation):
\begin{align}
\min_\vtheta\E_{t \sim \mathcal{U}(0, T)}\E_{p(\rvx_0)}\E_{\bm{\eps} \sim \mathcal{N}(0, \mI_d)}\left[\Vert\bm{\eps}_\vtheta(\vmu_t + \mL_t\bm{\eps},t) - \bm{\eps}\Vert_2^2 \right]. \label{eqn:4_1_4}
\end{align}
The epsilon-prediction objective has been shown to generate superior sample quality \citep{ho2020denoising, songscore, dockhornscore}. In this work, we optimize for sample quality and therefore use this objective for training all models. One key difference between the objective in Eqn. \ref{eqn:4_1_4} and the HSM objective in \gls{CLD} is that, unlike \gls{CLD}, we predict the full 2d-dimensional $\bm{\eps}$ due to the structure of our diffusion coefficient $\mG(t)$ (see Appendix \ref{app:B_2} for more details). Therefore, for a non-zero $\Gamma$, the neural-net-based score predictor in \mn has twice the number of output channels as in \gls{CLD}. However, the increase in parameters due to this architectural update is negligible.

\subsection{PSLD Sampling}
\label{subsec:sampling}
\noindent
Following the result from \citep{songscore}, the reverse process SDE corresponding to the forward process SDE defined in Eqn. \ref{eqn:4_1_1} can be formulated as follows:
\begin{equation}
    d\bar{\rvz}_t = \bar{\vf}(\bar{\rvz}_t)dt + \mG(T-t)d\bar{\rvw}_t
    \label{eqn:4_2_1}
\end{equation}
\begin{equation}
    \bar{\vf}(\bar{\rvz}_t) = \frac{\beta}{2}\begin{pmatrix}
        \Gamma \bar{\rvx}_t - M^{-1}\bar{\rvm}_t + 2\Gamma \vs_{\theta}(\bar{\rvz}_t, T-t)|_{0:d}\\
        \bar{\rvx}_t + \nu \bar{\rvm}_t + 2M\nu \vs_{\theta}(\bar{\rvz}_t, T-t)|_{d:2d})
    \end{pmatrix},\quad
    \mG(T-t) = \begin{pmatrix} \sqrt{\Gamma\beta} &0 \\ 0 &\sqrt{M\nu\beta} \end{pmatrix} \otimes \mI_d
\end{equation}
where $\bar{\rvz}_t = \rvz_{T - t}$, $\bar{\rvx}_t = \rvx_{T - t}$, $\bar{\rvm}_t = \rvm_{T - t}$. We can simulate this reverse process SDE using standard numerical SDE solvers like the Euler-Maruyama (EM) sampler \cite{Kloeden1992}. As an alternative, \citet{dockhornscore} propose SSCS: a symmetric splitting-based integrator and show that SSCS exhibits a better speed-sample quality tradeoff than EM. Consequently, we extend SSCS for \gls{\nsmn} by using the following splitting formulation:
\begin{equation}
    \begin{pmatrix}
        d\bar{\rvx}_t \\
        d\bar{\rvm}_t
    \end{pmatrix} = \underbrace{\frac{\beta}{2}\begin{pmatrix}
        - \Gamma \bar{\rvx}_t -M^{-1}\bar{\rvm}_t  \\
        \bar{\rvx}_t - \nu\bar{\rvm}_t
    \end{pmatrix} dt + \mG(T-t)d\bar{\rvw}_t}_{\text{Analytical-term}} + \underbrace{\beta\begin{pmatrix}
         \Gamma \bar{\rvx}_t + \Gamma\vs_{\theta}(\bar{\rvz}_t, T-t)|_{0:d} \\
        \nu\bar{\rvm}_t + M\nu\vs_{\theta}(\bar{\rvz}_t, T-t)|_{d:2d}
    \end{pmatrix}dt}_{\text{Score-term}}
    \label{eqn:4_2_2}
\end{equation}

\noindent
Indeed for $\Gamma=0$, the sampler in Eqn. \ref{eqn:4_2_2} resembles the SSCS sampler proposed in \citep{dockhornscore}. It is worth noting that despite an updated formulation, the order of the SSCS sampler, as analyzed in \citep{dockhornscore}, remains unchanged. We discuss the exact solution of the analytical part of the Modified-SSCS sampler in Eqn. \ref{eqn:4_2_2} and other relevant details in Appendix \ref{app:B_4_2}.

%% file: tex/exps.tex
\section{Experiments}
\label{sec:exps}

\noindent
\textbf{Datasets}: We run experiments on three datasets: CIFAR-10 \citep{krizhevsky2009learning}, CelebA \citep{liu2015faceattributes} at 64 x 64 resolution and the AFHQv2 \citep{choi2020stargan} dataset at 128 x 128 resolution.

\noindent
\textbf{Baselines}: We primarily compare \gls{\nsmn} with two popular \gls{SGM} baselines: VP-SDE \citep{songscore} and CLD \citep{dockhornscore} (a particular case of \gls{\nsmn} with $\Gamma=0$). For \gls{\nsmn} and CLD, unless specified otherwise, we operate in the critical damping regime with a fixed $M^{-1}=4$ and therefore choose $\Gamma$ and $\nu$ accordingly ($\nu=2\sqrt{M^{-1}} + \Gamma$, $\nu \geq 0$, $\Gamma \geq 0$).

\noindent
\textbf{Metrics}: We use the FID \citep{heusel2017gans} score for quantitatively assessing sample quality, while we use NFE (Number of Function Evaluations) to assess the sampling efficiency of all methods.

\noindent
We provide full implementation details in Appendix \ref{app:C}. The rest of our experimental section is organized as follows: Firstly, we compare the state-of-the-art performance of \gls{\nsmn} with popular SGM baselines on unconditional image generation. We show that \gls{\nsmn} outperforms competing baselines for similar compute budgets. Secondly, as an ablation experiment, we empirically and theoretically analyze the impact of the SDE parameters $\Gamma$ and $\nu$ on downstream sample quality in \gls{\nsmn}. Furthermore, we analyze the speed-quality trade-off in \gls{\nsmn} and show that \gls{\nsmn} yields better sample quality than competing baselines across four different sampler settings. Lastly, we show that pre-trained unconditional \gls{\nsmn} models can be used for downstream tasks like class-conditional image synthesis and image inpainting.

\subsection{State-of-the-art Comparisons}
\label{sec:exps_sota}

\begin{table}[!t]
\begin{minipage}[b]{0.49\linewidth}
\centering
\scriptsize
\begin{tabular}{@{}lccc@{}}
\toprule
Method & Size  & NFE  & FID@50k ($\downarrow$) \\ \midrule
\multicolumn{4}{@{}l}{\textbf{Ours (Baseline)}}                                \\ \midrule
CLD (w/o MS)     & 97M   & 1000 & 2.41                   \\ \midrule
\multicolumn{4}{@{}l}{\textbf{Ours (Proposed)}}                                \\ \midrule
\mn ($\Gamma$=0.02)        & 39M   & 1000 & 2.80                   \\
\mn ($\Gamma$=0.01)        & 55M   & 1000 & 2.34                   \\
\mn ($\Gamma$=0.02)        & 55M   & 1000 & 2.30                   \\
\mn ($\Gamma$=0.01)  & 97M   & 1000 & 2.23                   \\
\mn ($\Gamma$=0.02)  & 97M   & 1000 & \textbf{2.21}                   \\ \midrule
CLD (w/ MS) \citep{dockhornscore}                        & 108M  & 1000 & 2.27                   \\
$\text{VPSDE (deep)}^{\dagger}$ \citep{songscore}          & 108M  & 1000 & 2.46                   \\
$\text{VESDE (deep)}^{\dagger}$ \citep{songscore}               & 108M  & 1000 & 2.43                   \\
DDPM \citep{ho2020denoising}                      & 35.7M & 1000 & 3.17                   \\
iDDPM \citep{nichol2021improved}                      & - & 1000 & 2.90                   \\
DiffuseVAE \citep{pandeydiffusevae}       & 35.7M & 1000 & 2.80                   \\
NCSNv2 \citep{song2020improved}      & -     & -    & 10.87                  \\
NCSN \citep{song2019generative}                       & -     & 1000 & 25.32                  \\
VDM \citep{kingma2021variational}                       & -     & 1000    & 7.41                    \\\bottomrule
\end{tabular}
\caption{PSLD (SDE) sample quality comparisons for CIFAR-10. PSLD outperforms competing SDE baselines for a similar sampling budget. FID computed using 50k samples. MS: Mixed Score $\dagger$: Results from \citep{dockhornscore}.}
\label{table:main_4}
\end{minipage}
\hfill
\begin{minipage}[b]{0.49\linewidth}
\centering
\scriptsize
\begin{tabular}{@{}lccc@{}}
\toprule
\multicolumn{1}{@{}l}{Method}    & Size & NFE & FID@50k ($\downarrow$) \\ \midrule
\multicolumn{4}{@{}l}{\textbf{Ours (Baseline)}}                                 \\ \midrule
CLD (w/o MS)        & 97M  &  352   & 2.80                       \\\midrule
\multicolumn{4}{@{}l}{\textbf{Ours (Proposed)}}                                 \\\midrule
\mn ($\Gamma$=0.01)           & 55M  & 243 & 2.41                   \\
\mn ($\Gamma$=0.02)           & 55M  & 232 & 2.40                    \\
\mn ($\Gamma$=0.01)     & 97M  & 246 & \textbf{2.10}                 \\
\mn ($\Gamma$=0.02)     & 97M  & 231 & 2.31                   \\ \midrule
\multicolumn{4}{@{}l}{\textbf{Ours (Proposed)}}                                 \\\midrule
\mn ($\Gamma$=0.01)           & 97M  & 159 & 2.13                   \\
\mn ($\Gamma$=0.02)           & 97M  & 159 & 2.34                   \\\midrule
$\text{LSGM}^{\dagger}$ \citep{vahdat2021score} & 100M & 131 & 4.60                    \\
LSGM  \citep{vahdat2021score}                        & 476M & 138 & \textbf{2.10}                    \\
$\text{VPSDE}^{\dagger}$  \citep{songscore}                & 108M & 141 & 2.76                   \\
CLD (w/ MS) \citep{dockhornscore}                  & 108M & 147 & 2.71                   \\
CLD (w/ MS)  \citep{dockhornscore}                 & 108M & 312 & 2.25                   \\
ScoreFlow (VP) \citep{song2021maximum}          & 108M & -   & 5.34                   \\
Flow Matching (w/ OT) \citep{lipman2023flow}         & -    & 142 & 6.35                   \\
$\text{DDIM (VPSDE)}^{\dagger}$ \citep{songdenoising}       & 108M & 150 & 3.15                   \\ \bottomrule
\end{tabular}
\caption{PSLD (ODE) sample quality comparisons for CIFAR-10. PSLD outperforms most competing ODE baselines. FID computed using 50k samples. MS: Mixed Score. $\dagger$: From \citep{dockhornscore}.}
\label{table:main_5}
\end{minipage}
\end{table}

\noindent
\textbf{Setup}: We now compare the sample quality of our proposed method with existing popular SGM methods on the CIFAR-10 and CelebA-64 datasets for unconditional image synthesis. We use \gls{\nsmn} with $\Gamma \in \{0.01, 0.02\}$ for CIFAR-10 and \gls{\nsmn} with $\Gamma=0.005$ for CelebA-64 for state-of-the-art (SOTA) comparisons (See Section \ref{subsec:exp_impact} for a theoretical and empirical justification of these choices). Moreover, we use the training objective in Eqn. \ref{eqn:4_1_4} \textbf{without} any alternative score parameterizations (like \textit{mixed score}\citep{vahdat2021score, dockhornscore}) to train our models for SOTA comparisons. Unless specified otherwise, we perform sampling using the EM sampler with Uniform Striding (US) for CIFAR-10 and Quadratic Striding (QS) for CelebA-64 for the SDE setup and report FID scores on 50k samples (denoted as FID@50k). We include full details on our SDE and ODE solver setup for SOTA analysis in Appendix \ref{app:C_5}. We report the FID scores for most competing methods for a maximum sampling budget of N=1000 while reporting model sizes whenever available for CIFAR-10 for fair comparisons.\\

\noindent
\textbf{Main Observations}: Table \ref{table:main_4} compares CIFAR-10 sample quality between different methods using stochastic sampling. Our proposed method with $\Gamma=0.02$ and 39M parameters achieves an FID score of 2.80, outperforming the DDPM \citep{ho2020denoising} baseline while performing comparably with DiffuseVAE \citep{pandeydiffusevae}, for similar model sizes. It is worth noting that DiffuseVAE refines samples generated from a VAE \citep{kingma2013auto} using a DDPM backbone and is complementary to our work. \textit{Furthermore, our larger \gls{\nsmn} model achieves an FID of \textbf{2.21}, which is better than \gls{CLD} \citep{dockhornscore} (with or without the Mixed Score (MS) parameterization) and (VP/VE)-SDE baselines for similar NFE budget and model sizes}. For CelebA-64 (Table \ref{table:celeba64_sota}), \gls{\nsmn} outperforms the VP/VE-SDE baselines by a significant margin while requiring only 250 NFEs.

We next analyze ODE sample quality in \gls{\nsmn}. Table \ref{table:main_5} compares CIFAR-10 sample quality between different methods using ODE-based samplers. \gls{\nsmn} with $\Gamma=0.01$ achieves an FID score of 2.10 and outperforms most competing methods except LSGM \citep{vahdat2021score}. Though the original LSGM model is more than four times the size of our SOTA model, \gls{\nsmn} performs comparably with LSGM. When scaled to a similar size, LSGM performs much worse than \gls{\nsmn} (FID: 4.60 for LSGM-100M vs. 2.10 for \gls{\nsmn}). We note that EDM \citep{karraselucidating} achieves an FID of 2.05 on unconditional CIFAR-10 generation (without data augmentation) by analyzing several design choices associated with diffusion models (like
score network architectures, loss preconditioning, and sampler design). We did not explore this line of research but note that their approach complements our proposed method, and exploring some of these design choices in the context of \gls{\nsmn} can be an exciting future direction.

Interestingly, \gls{\nsmn} with the ODE setup obtains a better FID score than the SDE setup (FID: 2.21 for SDE vs. 2.10 for ODE) while requiring around four times lesser NFEs. Moreover, when using a solver tolerance of $1e^{-4}$, \gls{\nsmn} achieves an FID score of 2.13, comparable to the best FID of 2.10 while reducing NFEs significantly. \textit{This tradeoff is worse for other SGMs like \gls{CLD} and VP-SDE (Table \ref{table:main_5})}. We report additional SOTA results in Appendix \ref{app:D_3}.

\subsection{Impact of $\Gamma$ and $\nu$ on \mn Sample Quality}
\label{subsec:exp_impact}

\begin{table}[]
\begin{minipage}{0.49\linewidth}
\centering
\scriptsize
\begin{tabular}{@{}lcc@{}}
\toprule
Method                     & NFE  & FID@50k       \\ \midrule
(\textbf{Ours}) PSLD ($\Gamma=0.005$)      & 250  & \textbf{2.01} \\
(\textbf{Ours}) PSLD ($\Gamma=0.005$, ODE) & 244  & 2.56          \\ \midrule
PNDM \citep{liupseudo}                       & 250  & 2.71          \\
DDIM \citep{songdenoising}                      & 250  & 4.44          \\
VPSDE \citep{songscore}                      & 1000 & 2.32          \\
Gamma DDPM \citep{nachmani2021non}                 & 1000 & 2.92          \\
DDPM  \citep{ho2020denoising}                     & 1000 & 3.26          \\
DiffuseVAE \citep{pandeydiffusevae}                 & 1000 & 3.97          \\
VESDE \citep{songscore}                     & 2000 & 3.95          \\
NCSN (w/ denoising) \citep{song2019generative}        & -    & 25.3          \\
NCSNv2 (w/ denoising) \citep{song2020improved}      & -    & 10.23         \\ \bottomrule
\end{tabular}
\caption{PSLD sample quality comparisons for CelebA-64. FID computed using 50k samples.}
\label{table:celeba64_sota}
\end{minipage}
\hfill
\begin{minipage}{0.49\linewidth}
\centering
\scriptsize
\begin{tabular}{@{}ccccc@{}}
\toprule
         & \multicolumn{2}{c}{CIFAR-10 (39M)}                                                                                     & \multicolumn{2}{c}{CelebA-64 (66M)}                                                                                     \\ \midrule
$\Gamma$ & \begin{tabular}[c]{@{}c@{}}FID@50k\\ (EM-QS)\end{tabular} & \begin{tabular}[c]{@{}c@{}}FID@50k \\ (EM-US)\end{tabular} & \begin{tabular}[c]{@{}c@{}}FID@10k \\ (EM-QS)\end{tabular} & \begin{tabular}[c]{@{}c@{}}FID@10k \\ (EM-US)\end{tabular} \\ \midrule
0        & 3.64                                                      & 3.60                                                       & 4.59                                                       & 4.60                                                       \\
0.005    & 3.42                                                      & 3.34                                                       & \textbf{4.17}                                              & 4.37                                                       \\
0.01     & 3.15                                                      & 2.94                                                       & 4.22                                                       & 4.34                                                       \\
0.02     & 3.26                                                      & \textbf{2.80}                                              & 4.43                                                       & 4.52                                                       \\
0.25     & 4.99                                                      & 9.48                                                       & 93.99                                                      & 95.13                                                      \\ \bottomrule
\end{tabular}
\caption{Impact of increasing $\Gamma$ (with fixed $M^{-1}$) on sample quality (NFE=1000). FID computed using 50k and 10k samples for the CIFAR-10 and CelebA-64 datasets, respectively. QS: Quadratic Striding, US: Uniform Striding.}
\label{table:main_1}
\end{minipage}
\end{table}
\begin{figure}[h]
    \centering
    \includegraphics[width=0.6\linewidth]{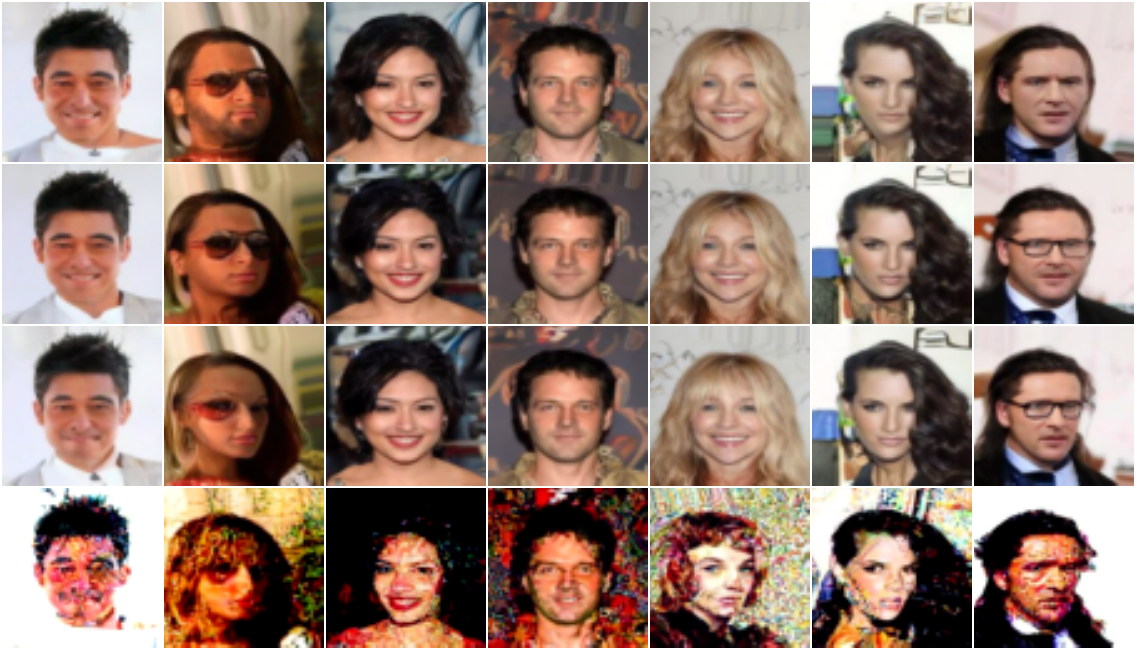}
    \caption{Impact of increasing $\Gamma=\{0, 0.005, 0.02, 0.25\}$ (Top to Bottom) on CelebA-64 sample quality. The best sample quality is achieved at $\Gamma=0.005$ (Second Row) while increasing $\Gamma$ to 0.25 results in loss of high-frequency image features.}
    \label{fig:main_1}
\end{figure}

\noindent
\textbf{Setup and Baselines}: Since adding stochastic noise in both the data and the momentum space is one of the primary aspects of \gls{\nsmn}, we now analyze the impact of the choice of $\Gamma$ and $\nu$ on downstream sample quality. For subsequent experimental results, we use our smaller ablation models (for PSLD and relevant baselines) for comparisons. Table \ref{table:main_1} shows the impact of varying $\Gamma$ on sample quality for the CIFAR-10 and CelebA-64 datasets. Our ablation \gls{CLD} baseline (\gls{\nsmn} with $\Gamma=0, \nu=4$) achieves an FID of 3.60 using the Euler-Maruyama (EM) sampler with Uniform striding (US) and 3.64 using the EM sampler with Quadratic striding (QS). Our results are comparable with the FID of 3.56 obtained by \citep{dockhornscore} for their \gls{CLD} ablation model on CIFAR-10 without using the mixed-score parameterization. Our VP-SDE ablation baseline (not shown in Table \ref{table:main_1}) obtains an FID of 3.19 using EM-US (with 1000 NFEs).\\

\noindent
\textbf{Main Observations}: We observe that \textit{setting $\Gamma$ to a non-zero value within a specific range improves sample quality significantly over \gls{CLD}}. Specifically, our ablation CIFAR-10 model achieves FID scores of 2.94 and 2.80 for $\Gamma=0.01$ and $\Gamma=0.02$ respectively (with EM-US) and outperforms our VP-SDE and \gls{CLD} baselines without using alternative score-network parameterizations like mixed-score which is crucial for competitive performance of \gls{CLD} \citep{dockhornscore}. We make a similar observation for the CelebA-64 dataset on which our model achieves the best FID of 4.17 using EM-QS and outperforms our \gls{CLD} baseline (FID: 4.59). Interestingly, the sample quality worsens for both datasets on increasing $\Gamma$ outside a range. For instance, for CIFAR-10, further increasing $\Gamma$ from 0.02 to 0.04 (not shown in Table \ref{table:main_1}) resulted in an increase in FID from $2.80$ to $2.95$. Consequently, the sample quality for both datasets is the worst at $\Gamma=4.25, \nu=0.25$. We also note that EM-US works better than EM-QS for CIFAR-10 and vice-versa for CelebA-64.

Figure \ref{fig:main_1} further validates our findings qualitatively for the CelebA-64 dataset where for $\Gamma=0.25$, the score network can only recover high-level semantic structures (like gender and glasses, among others) but is unable to recover high-frequency details. Since the diffusion denoiser recovers most high-frequency information in the low-timestep regime, these observations suggest denoising issues near low-timestep indices. We next provide a formal justification for this observation.\\

\noindent
\textbf{Theoretical justification of adding stochasticity in the position space}: Since \gls{\nsmn} involves adding stochasticity in both the data and the momentum space, during training, we need to predict the noise $\bm{\eps}^x_{\theta}(\rvz_t, t)$ and $\bm{\eps}^m_{\theta}(\rvz_t, t)$ in both the data and the momentum space respectively. Therefore, it is unclear why \gls{\nsmn} leads to better sample quality than CLD since predicting both noise components can lead to additional sources of errors during sampling.

However, in the context of the EM sampler, we find (see Appendix \ref{app:D_1}) that setting a \textit{small} non-zero $\Gamma$ can significantly suppress prediction errors from $\bm{\eps}^x_{\theta}(\rvz_t, t)$ at the expense of introducing negligible extra errors from $\bm{\eps}^m_{\theta}(\rvz_t, t)$. Contrarily, using larger values of $\Gamma$ results in scaling the prediction errors from $\bm{\eps}^x_{\theta}(\rvz_t, t)$ by a significant factor, especially in the low-timestep regime, leading to worse sample quality with significant degradations in high-frequency sample details as observed in Figure \ref{fig:main_1}.

Therefore, intuitively, \textit{$\Gamma$ introduces a trade-off between error contribution from both noise predictors $\bm{\eps}^x_{\theta}(\rvz_t, t)$ and $\bm{\eps}^m_{\theta}(\rvz_t, t)$ with small values of $\Gamma$ providing a favorable trade-off which improve overall sample quality}. As a general guideline, we find $\Gamma=0.01$ to work well across datasets. Figure \ref{fig:main_0} shows some qualitative samples generated from \gls{\nsmn} trained on the AFHQv2 \citep{choi2020stargan} dataset with $\Gamma=0.01, \nu=4.01$.

\subsection{Sample Speed vs. Quality Tradeoffs for \mn}

\begin{table}[]
\begin{minipage}[b]{0.56\linewidth}
\centering
\scriptsize
\begin{tabular}{@{}ccccccc@{}}
\toprule
                         &             & \multicolumn{5}{c}{NFE (FID@10k $\downarrow$)}                                  \\ \midrule
Sampler                  & Method      & 50             & 100            & 250           & 500           & 1000          \\ \midrule
\multirow{3}{*}{EM-QS}   & CLD         & 25.01          & 8.91           & 5.97          & 5.61          & 5.7           \\
                         & VP-SDE      & \textbf{17.72} & 7.45           & 5.59          & 5.51          & 5.51          \\
                         & (Ours) PSLD & 19.94          & \textbf{7.33}  & \textbf{5.26} & \textbf{5.20} & \textbf{5.28} \\ \midrule
\multirow{3}{*}{EM-US}   & CLD         & 119.68         & 45.60          & \textbf{9.08} & 5.71          & 5.65          \\
                         & VP-SDE      & \textbf{84.54} & 41.93          & 12.61         & 5.92          & 5.19          \\
                         & (Ours) PSLD & 100.62         & \textbf{39.96} & 11.26         & \textbf{5.45} & \textbf{4.82} \\ \midrule
\multirow{2}{*}{SSCS-QS} & CLD         & 21.31          & 8.37           & 5.82          & 5.75          & 5.69          \\
                         & (Ours) PSLD & \textbf{16.12} & \textbf{7.16}  & \textbf{5.36} & \textbf{5.35} & \textbf{5.27} \\ \midrule
\multirow{2}{*}{SSCS-US} & CLD         & 75.45          & 24.74          & 6.09          & 5.74          & 5.78          \\
                         & (Ours) PSLD & \textbf{72.42} & \textbf{20.46} & \textbf{5.19} & \textbf{4.92} & \textbf{5.29} \\ \bottomrule 
\end{tabular}
\caption{PSLD exhibits better speed vs. sample quality tradeoffs over competing baseline SDEs (CLD and VP-SDE) on CIFAR-10 across four samplers configurations. The rightmost five columns indicate NFEs, with \textbf{bold} indicating the best result for that sampler. QS: Quadratic Striding, US: Uniform Striding. See Appendix \ref{app:D_2} for extended results.}
\label{table:main_2}
\end{minipage}
\hfill 
\begin{minipage}[b]{0.41\linewidth}
\centering
\scriptsize
\begin{tabular}{@{}cccc@{}}
\toprule
Method                                                                              & $\log_{10}{\text{tol}}$ & FID@10k ($\downarrow$) & Avg. NFE \\ \midrule
\multirow{5}{*}{\begin{tabular}[c]{@{}c@{}}CLD \\ (Baseline)\end{tabular}}  & -5               & 5.54    & 280      \\
                                                                                    & -4               & 5.62    & 196      \\
                                                                                    & -3               & 6.54    & 147      \\
                                                                                    & -2               & 9.98    & 86       \\
                                                                                    & -1               & 397.1   & 27       \\\midrule
\multirow{5}{*}{\begin{tabular}[c]{@{}c@{}}\mn \\ ($\Gamma=0.02$)\end{tabular}} & -5               & \textbf{4.79}    & 228      \\
                                                                                    & -4               & 4.84    & 158      \\
                                                                                    & -3               & 5.09    & 111      \\
                                                                                    & -2               & 16.11   & 69       \\
                                                                                    & -1               & 418.779 & 27       \\ \midrule
VPSDE                                                                               & -5               & 5.91    & 123      \\ \bottomrule
\end{tabular}
\caption{PSLD exhibits better speed vs. sample quality tradeoffs over competing baselines on CIFAR-10 using a black-box ODE solver. $\log_{10}{\text{tol}}$ indicates the ODE sampler (RK45) tolerance. \textbf{Bold} indicates best result for that column.}
\label{table:main_3}
\end{minipage}

\end{table}

\label{subsec:exp_svq}

\noindent
\textbf{Sampler Setup}: Since the tradeoff between sample quality and the number of reverse sampling steps required is crucial for any SGM backbone, we now examine this tradeoff for \gls{\nsmn} for the CIFAR-10 dataset (See Appendix \ref{app:D_2} for extended results on the CelebA-64 dataset). We use our VP-SDE and \gls{\nsmn} with $\Gamma=0$ (corresponding to \gls{CLD}) ablation models as comparison baselines. Furthermore, we use combinations of the EM and SSCS samplers with Uniform (US) and Quadratic (QS) timestep striding as different sampler settings to benchmark the performance of all methods. It is worth noting that the SSCS sampler can only be used for augmented SGMs like \gls{CLD} and \gls{\nsmn}. For ODE-based comparisons, we use the probability flow ODE setup with RK45 \citep{DORMAND198019} solver (see Appendix \ref{app:C_5} for more details). Lastly, we measure sample quality using FID computed for 10k samples.

\noindent
\textbf{Main Observations}: Table \ref{table:main_2} shows a comparison between FID scores for our best performing \gls{\nsmn} models (corresponding to $\Gamma=0.01$ and $\Gamma=0.02$) and our VP-SDE and \gls{CLD} baselines from Section \ref{subsec:exp_impact} for the CIFAR-10 dataset across $N \in \{50, 100, 250, 500, 1000\}$ steps. We primarily observe that \textit{\gls{\nsmn} outperforms the VP-SDE and \gls{CLD} baselines across all comparison points} with the most significant differences at lower NFE (Network Function Evaluations) values. For instance, \gls{\nsmn} ($\Gamma=0.02$) achieves the best FID of 16.12 at NFE=50 compared to 17.72 and 21.31 by VP-SDE and \gls{CLD}, respectively. Moreover, for most sampler settings across methods, SSCS performs better in the low NFE regime ($N \leq 250$), while EM performs better for a higher number of NFEs. A similar observation was made in \citep{dockhornscore}.
Similarly, Quadratic striding works much better in the low NFE regime, while Uniform striding works better when using a higher number of NFEs ($N > 500$).

We next compare our best-performing ablation model (\gls{\nsmn} with $\Gamma=0.02$) with our VP-SDE and \gls{CLD} baselines using the probability flow ODE setup across multiple tolerance levels on the CIFAR-10 dataset. Table \ref{table:main_3} compares FID scores (computed on 10k samples) for all methods. Like our SDE setup, \textit{\gls{\nsmn} ($\Gamma$=0.02) outperforms both baselines in sample quality for similar NFE budgets}. Moreover, across the same solver tolerance level, \gls{\nsmn} requires fewer NFEs on average than its \gls{CLD} counterpart while yielding better sample quality. Lastly, we found using black-box solvers to further improve sample quality compared to the SDE baseline at a tolerance level of 1e-5 for both our \gls{\nsmn} and \gls{CLD} models for CIFAR-10 (FID@10k=4.97 for \gls{\nsmn} for ODE vs. FID@10k=4.84 for EM-QS (N=1000) with $\Gamma=0.02$). This observation is consistent with the ODE comparison results presented in Section \ref{sec:exps_sota}.

\begin{figure}
    \centering
    \includegraphics[width=1.0\linewidth]{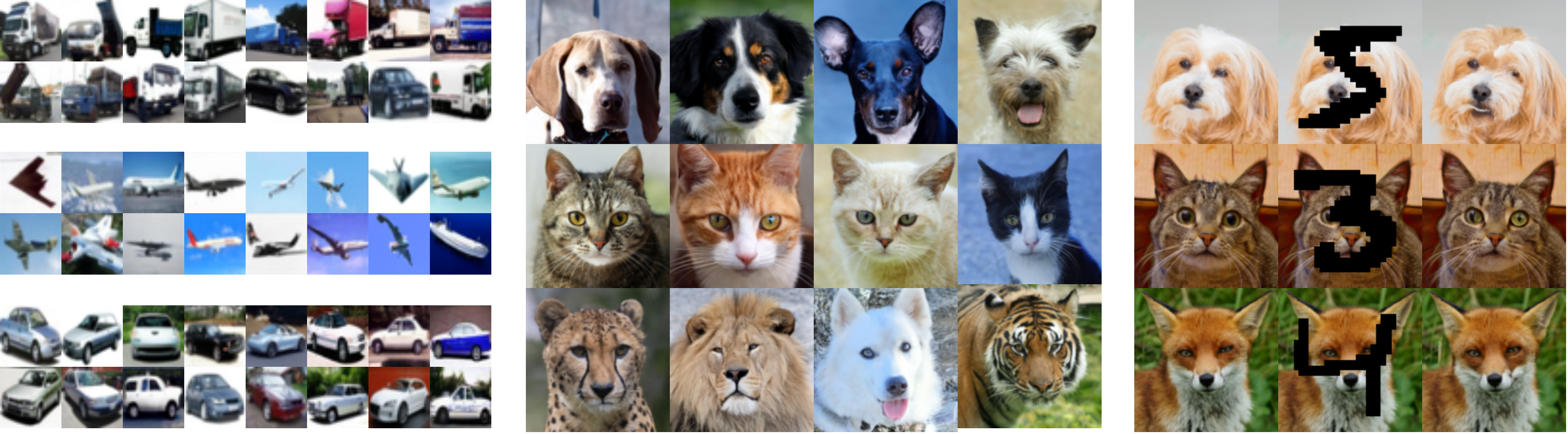}
    \caption{(Left) Class-conditional results on CIFAR-10: \textit{Truck}, \textit{Airplane}, and \textit{Automobile} from Top to Bottom (two rows each). (Middle) Class conditional results on AFHQv2: \textit{Dogs}, \textit{Cats} and \textit{Others} from Top to Bottom (one row each). (Right) Inpainting results on AFHQv2. The columns represent the original, corrupted, and imputed samples, respectively, from left to right.}
    \label{fig:controllable}
\end{figure}
\begin{table}[]
\centering
\scriptsize
\begin{tabular}{@{}ccc@{}}
\toprule
Method                      & FID (Train)             & FID (Test)             \\ \midrule
CLD                         & 1.01              & 7.10             \\
(Ours) PSLD ($\Gamma=0.01$) & \textbf{0.85}     & \textbf{6.93}    \\ \bottomrule
\end{tabular}
\caption{PSLD outperforms CLD on Image inpainting for the AFHQv2 dataset. FID (lower is better) is computed on the full train and test sets.}
\label{table:main_inpaint}
\end{table}

\subsection{Conditional Generation with PSLD}
\noindent
Following prior work \citep{songscore, dhariwal2021diffusion}, given some conditioning information $\rvy$, an unconditional pre-trained score network $\vs_{\theta}(\rvz_t, t)$ can be used for sampling from the distribution $p(\rvz_t|\rvy)$ in \gls{\nsmn}. More specifically,
\begin{align}
    \nabla_{\rvz_t} \log{p(\rvz_t|\rvy)} &= \nabla_{\rvz_t} \log{p(\rvy|\rvz_t)} + \nabla_{\rvz_t} \log{p(\rvz_t)} \\
    &\approx \nabla_{\rvz_t} \log{p(\rvy|\rvz_t)} + \vs_{\theta}(\rvz_t, t)
    \label{eqn:guidance}
\end{align}
We can then use the estimate of $\nabla_{\rvz_t} \log{p(\rvz_t|\rvy)}$ in Eqn. \ref{eqn:guidance} to sample from the following SDE for conditional generation:
\begin{equation}
    d\rvz_t = \left[\vf(\rvz_t) - \mG(t)\mG(t)^T \nabla_{\rvz_t} \log{p(\rvz_t|\rvy)}\right]dt  + \mG(t)d\rvw_t.
    \label{eqn:controllable}
\end{equation}
Figure \ref{fig:controllable} illustrates class conditional samples for the CIFAR-10 and the AFHQv2 datasets obtained by training an additional \textit{time-dependent classifier} $p(\rvy|\rvz_t)$ to compute $\nabla_{\rvz_t} \log{p(\rvy|\rvz_t)}$, followed by sampling from the SDE in Eqn. \ref{eqn:controllable} (full implementation details in Appendix \ref{app:D_4}). Similarly, we can perform data imputation by setting the conditioning signal $\rvy = \bar{\rvz}_0$ where $\bar{\rvz}_0$ is the observed part of the input data $\rvz_0$ (See Figure \ref{fig:controllable}). For image inpainting, \gls{\nsmn} exhibits a better perceptual quality of inpainted samples over CLD on the AFHQv2 dataset (See Table \ref{table:main_inpaint}). We include a complete derivation for inpainting and an analogous framework to \citep{songscore} for solving inverse problems using \gls{\nsmn} with additional conditional synthesis results in Appendix \ref{app:D_4}.

%% file: tex/related.tex
\section{Related Work}
\label{sec:related}
\noindent
\textbf{Advances in Diffusion Models}: Following the seminal work on diffusion (a.k.a score-based) models \citep{sohl2015deep, ho2020denoising, song2019generative, songscore}, there has been much recent progress in advancing unconditional \citep{nichol2021improved, dhariwal2021diffusion, dockhornscore, jing2022subspace, jolicoeur-martineau2021adversarial, vahdat2021score, salimansprogressive, rombach2022high} and conditional \citep{saharia2022image, chenwavegrad, songscore, rombach2022high, pandeydiffusevae} diffusion models for a variety of downstream tasks like text-to-image synthesis \citep{nichol2022glide, ramesh2022hierarchical}, image super-resolution \citep{saharia2022image, li2022srdiff} and video generation \citep{hovideo, yang2022diffusion, yuscaling, ho2022imagen}. Our work is closely related to \gls{CLD} \citep{dockhornscore}, which is motivated by Langevin heat baths in statistical mechanics \citep{alma991006381329705251}. However, our method is not directly motivated by physical interpretation but rather directly constructed from our proposed drift parameterization. Another line of research in \glspl{SGM} is to perform score-based modeling in the latent space \citep{vahdat2021score, rombach2022high, sinha2021d2c} of a powerful autoencoder \citep{vahdat2020nvae, esser2021taming}. Such approaches have been shown to improve the sampling time in SGMs. Therefore, since we propose a novel diffusion model backbone, most existing advances in diffusion models complement \gls{\nsmn}.

\noindent
\textbf{Sampler Design in Diffusion Models}: Improving the speed-vs-quality tradeoff in SGMs is a fundamental area in diffusion model research \citep{songdenoising,bao2022analyticdpm, kongfast, 
liupseudo, zhang2023fast, zhang2022gddim}. One popular approach to speed up diffusion model sampling is DDIM \citep{songdenoising}. \citep{zhang2023fast} show that DDIM can be cast as an exponential integrator and propose further improvements. \citep{zhang2022gddim} further leverage these improvements to propose a \textit{generalized-DDIM} (gDDIM) method for \gls{CLD}. It is worth noting that gDDIM parameterization requires predicting the score w.r.t both the data and the auxiliary variables and is directly compatible with \gls{\nsmn}. Another line of research involves training to speed up diffusion sampling. GENIE \citep{dockhorngenie} proposes to utilize higher-order Taylor methods during training to speed up DDIM sampling. Alternatively, distillation-based approaches distill a teacher into a student diffusion model progressively \citep{salimansprogressive, meng2022distillation} or otherwise \citep{luhman2021knowledge}. Therefore, exploring some of these directions in the context of \gls{\nsmn} would be interesting.

\noindent
\textbf{Auxiliary Diffusion Models}: In a concurrent work, \citep{singhal2023where} define an ELBO for multivariate diffusion models (MDM) and introduce a similar recipe as ours to design new diffusion processes. While \citep{singhal2023where} optimize for likelihood estimates, we primarily focus on sample quality in this work. Both works illustrate a different perspective on the advantages of constructing a generic recipe for designing diffusion processes and, therefore, complementary. Another recent work, Flexible Diffusion \citep{du2022flexible}, exploits the geometry of the data manifold to parameterize the forward process. The proposed framework is complete under linear drift. However, our parameterization makes no such assumptions.

%% file: tex/conclusion.tex
\section{Conclusion}
\label{sec:conclusion}
\noindent
We presented a recipe for constructing forward process parameterization for diffusion processes that guarantees convergence to a prespecified stationary distribution, such as a Gaussian. We use the proposed recipe to construct a novel diffusion process: \fmn 
(\gls{\nsmn}) which achieves excellent sample quality with better speed-vs-quality tradeoffs compared to existing baselines like the VP-SDE and \gls{CLD} on standard image-synthesis benchmarks. We left the exploration of potentially performance-improving design choices such as alternative score network parameterizations and loss weighting  \citep{karraselucidating} as directions for future work. 

While this work only explores stochastic samplers with a single auxiliary "momentum" variable $\rvm$ (of the same dimension as $\rvx$), exploring other design choices of $\mD(\rvz)$ and $\mQ(\rvz)$ \citep{singhal2023where}, which lead to higher-order stochastic samplers (like the Nos\'{e}-Hoover Thermostat) could also be an interesting research direction. Furthermore, our current choices of $\mD(\rvz)$ and $\mQ(\rvz)$ are limited to constant matrices due to relying on denoising score matching. Therefore, the proposed parameterization offers a complementary framework for designing diffusion generative models trained using alternative score-matching techniques. 

Lastly, our proposed recipe is only \emph{complete} under the assumption that both $\rvx$ and $\rvm$ are required to converge to prescribed marginals $p(\rvx)$ and $p(\rvm)$. Without this requirement on $\rvm$, the design space of samplers is potentially larger (as has been pointed out in the Bayesian MCMC literature) and may, e.g., include microcanonical samplers \citep{robnik2023microcanonical, ver2021hamiltonian}. However, the requirements of generative diffusion models are more strict and demand that the forward process's asymptotic joint distribution over $\rvx$ and $\rvm$ has a simple form that enables sampling in constant time. In contrast, Bayesian MCMC only requires the $\rvx$-marginal to converge to the prescribed posterior. We still think that relaxing the requirements on tractability enables potentially promising new samplers for future exploration. \newpage

%% file: tex/additional.tex
\textbf{Acknowledgements}
We thank Gavin Kerrigan, Uros Seljak, and Rajesh Ranganath for insightful discussions. KP acknowledges support from the HPI Research Center
in Machine Learning and Data Science at UC Irvine. 
SM acknowledges support from the National Science Foundation (NSF) under an NSF CAREER Award, award numbers 2003237 and 2007719, by the Department of Energy under grant DE-SC0022331, the IARPA WRIVA program, and by gifts from Qualcomm and Disney.

%% file: tex/appendix/App_A.tex
\tableofcontents
\newpage
\section{A Complete Recipe for SGMs}
\label{app:A}

\subsection{Proof of Theorems}
\label{app:A_1}
\subsubsection{Proof of Stationarity}
Given a positive semi-definite diffusion matrix $\mD(\rvz)$ and a skew-symmetric matrix $\mQ(\rvz)$, we can parameterize the drift $f(\rvz)$ for a stochastic process: $d\rvz = \vf(\rvz) dt + \sqrt{2\mD(\rvz)}d\rvw_t$ as follows:
\begin{equation}
\vf(\rvz) = -(\mD(\rvz) + \mQ(\rvz))\nabla H + \tau(\rvz), \quad\;\quad 
\tau_i(\rvz) = \sum_j \frac{\partial}{\partial \rvz_j}\left(\mD_{ij}(\rvz) + \mQ_{ij}(\rvz)\right)
\end{equation}
Theorem 2.1 then states that the distribution $p_s(z) \propto \exp{(-H(\rvz))}$ will be the stationary distribution for the stochastic process as defined above. 
\begin{proof}
Using the Fokker-Planck formulation for the stochastic dynamics, we have:
\begin{equation}
    \frac{\partial p_t(\rvz)}{\partial t} = -\sum_i \frac{\partial}{\partial z_i}\left(\vf_i(\rvz)p_t(\rvz)\right) + \sum_{i,j} \frac{\partial^2}{\partial \rvz_i\rvz_j} \left(\mD_{ij}(\rvz)p_t(\rvz)\right)
    \label{eqn:fpk}
\end{equation}
Furthermore, we have:
\begin{equation}
    \vf_i(\rvz) = \tau_i(\rvz) -\sum_j \left(\mD_{ij}(\rvz) + \mQ_{ij}(\rvz)\right)\nabla H_j(\rvz)
\end{equation}
Therefore,
{
\small
\begin{align}
    \sum_i \frac{\partial}{\partial \rvz_i}\left(\vf_i(\rvz)p_t(\rvz)\right) &= \sum_i \frac{\partial}{\partial z_i}\left[\tau_i(\rvz)p_t(\rvz) -\sum_j \left(\mD_{ij}(\rvz) + \mQ_{ij}(\rvz)\right)\nabla H_j(\rvz)p_t(\rvz)\right] \\
    &= \sum_i \frac{\partial}{\partial \rvz_i}\left[\sum_j \frac{\partial}{\partial \rvz_j}\left(\mD_{ij}(\rvz) + \mQ_{ij}(\rvz)\right)p_t(\rvz) -\sum_j \left(\mD_{ij}(\rvz) + \mQ_{ij}(\rvz)\right)\nabla H_j(\rvz)p_t(\rvz)\right] \\
    &= \sum_{i,j} \frac{\partial}{\partial \rvz_i}\Bigg[\frac{\partial}{\partial \rvz_j}\left(\mD_{ij}(\rvz) + \mQ_{ij}(\rvz)\right)p_t(\rvz)\Bigg] -\sum_{i,j} \frac{\partial}{\partial \rvz_i}\Bigg[\left(\mD_{ij}(\rvz) + \mQ_{ij}(\rvz)\right)\nabla H_j(\rvz)p_t(\rvz)\Bigg] \\
    &=\sum_{i,j} \frac{\partial}{\partial \rvz_i}\Bigg[\frac{\partial}{\partial \rvz_j}\left(\mD_{ij}(\rvz) + \mQ_{ij}(\rvz)\right)p_t(\rvz)\Bigg] -\underbrace{\sum_{i,j} \frac{\partial}{\partial \rvz_i}\Bigg[\left(\mD_{ij}(\rvz) + \mQ_{ij}(\rvz)\right)\nabla H_j(\rvz)p_t(\rvz)\Bigg]}_{=F(\rvz)}
\end{align}
}
Substituting the above result in the Fokker-Planck formulation in Eqn. \ref{eqn:fpk}, we have:
\begin{align}
    \frac{\partial p_t(\rvz)}{\partial t} &= F(\rvz) - \sum_{i,j} \frac{\partial}{\partial \rvz_i}\Bigg[\frac{\partial}{\partial \rvz_j}\left(\mD_{ij}(\rvz) + \mQ_{ij}(\rvz)\right)p_t(\rvz) - \frac{\partial}{\partial \rvz_j} \left(\mD_{ij}(\rvz)p_t(\rvz)\right)\Bigg] \\
    &= F(\rvz) - \sum_{i,j} \frac{\partial}{\partial \rvz_i}\Bigg[\frac{\partial}{\partial \rvz_j}\left(\mQ_{ij}(\rvz)\right)p_t(\rvz) - \mD_{ij}(\rvz)\frac{\partial}{\partial \rvz_j} \left(p_t(\rvz)\right)\Bigg]\\
    &= F(\rvz) - \sum_{i,j} \frac{\partial}{\partial \rvz_i}\Bigg[\frac{\partial}{\partial \rvz_j}\left(\mQ_{ij}(\rvz)p_t(\rvz)\right) - \left(\mD_{ij}(\rvz) + \mQ_{ij}(\rvz)\right)\frac{\partial}{\partial \rvz_j} \left(p_t(\rvz)\right)\Bigg] \\
    &= F(\rvz) + \underbrace{\sum_{i,j} \frac{\partial}{\partial \rvz_i}\Bigg[\left(\mD_{ij}(\rvz) + \mQ_{ij}(\rvz)\right)\frac{\partial p_t(\rvz)}{\partial \rvz_j}\Bigg]}_{=G(\rvz)} - \sum_{i,j}\frac{\partial^2}{\partial \rvz_i\partial \rvz_j}\left(\mQ_{ij}(\rvz)p_t(\rvz)\right) \\
    &= F(\rvz) + G(\rvz) - \sum_{i,j}\frac{\partial^2}{\partial \rvz_i\partial \rvz_j}\left(\mQ_{ij}(\rvz)p_t(\rvz)\right)
\end{align}
Since $\mQ(\rvz)$ is a skew-symmetric matrix, $\sum_{i,j}\frac{\partial^2}{\partial \rvz_i\partial \rvz_j}\left(\mQ_{ij}(\rvz)p_t(\rvz)\right) = 0$. Therefore,
\begin{align}
    \frac{\partial p_t(\rvz)}{\partial t} &= F(\rvz) + G(\rvz) \\
    &= \sum_{i,j} \frac{\partial}{\partial \rvz_i}\Bigg[\left(\mD_{ij}(\rvz) + \mQ_{ij}(\rvz)\right)\left(\nabla H_j(\rvz)p_t(\rvz) + \frac{\partial p_t(\rvz)}{\partial \rvz_j}\right)\Bigg]\\
    &= \sum_i \frac{\partial}{\partial \rvz_i}\Bigg[\left(\mD_i(\rvz) + \mQ_i(\rvz)\right)\left(\nabla H(\rvz)p_t(\rvz) + \frac{\partial p_t(\rvz)}{\partial \rvz}\right)\Bigg] \\
    &= \nabla \cdot \Bigg[\left(\mD_i(\rvz) + \mQ_i(\rvz)\right)\left(\nabla H(\rvz)p_t(\rvz) + \frac{\partial p_t(\rvz)}{\partial \rvz}\right)\Bigg]
\end{align}
Therefore, we have the following parameterization for the Fokker-Planck formulation for the defined stochastic dynamics:
\begin{equation}
    \frac{\partial p_t(\rvz)}{\partial t} = \nabla \cdot \Bigg[\left(\mD_i(\rvz) + \mQ_i(\rvz)\right)\left(\nabla H(\rvz)p_t(\rvz) + \frac{\partial p_t(\rvz)}{\partial \rvz}\right)\Bigg]
\end{equation}
Substituting $p_s(\rvz) \propto \exp{(-H(\rvz))}$ in the above result implies $\frac{\partial p_t(\rvz)}{\partial t} = 0$. This implies that $p_s(\rvz) \propto \exp{(-H(\rvz))}$ is the form of the stationary distribution for the drift parameterization $\vf(\rvz) = -(\mD(\rvz) + \mQ(\rvz))\nabla H + \tau(\rvz)$. An alternative version of this proof can be found in \citet{ma2015complete}
\end{proof}

\subsubsection{Proof of Completeness}
\label{app:A_2}
We now state the proof for Theorem 2.2 which states that for every stochastic dynamics $d\rvz = \vf(\rvz) dt + \sqrt{2\mD(\rvz)}d\rvw_t$ with the desired stationary distribution $p_s(\rvz) \propto \exp{(-H(\rvz))}$, there exists a positive semi-definite $\mD(\rvz)$ and a skew-symmetric $\mQ(\rvz)$ such that $\vf(\rvz) = -(\mD(\rvz) + \mQ(\rvz))\nabla H + \bm{\tau}(\rvz)$ holds. We directly include the proof from \citet{ma2015complete} for completeness.

\begin{proof}
We have the following result:
\begin{align}
    \vf_i(\rvz)p_s(\rvz) &= \bm{\tau}_i(\rvz)p_s(\rvz) -\sum_j \left(\mD_{ij}(\rvz) + \mQ_{ij}(\rvz)\right)\nabla H_j(\rvz)p_s(\rvz) \\
    &= \sum_j \frac{\partial}{\partial \rvz_j}\left(\mD_{ij}(\rvz) + \mQ_{ij}(\rvz)\right)p_s(\rvz) -\sum_j \left(\mD_{ij}(\rvz) + \mQ_{ij}(\rvz)\right)\nabla H_j(\rvz)p_s(\rvz) \\
    &= \sum_j \frac{\partial}{\partial \rvz_j}\Bigg[\left(\mD_{ij}(\rvz) + \mQ_{ij}(\rvz)\right)p_s(\rvz)\Bigg]
\end{align}
which implies,
\begin{equation}
    \sum_j \frac{\partial}{\partial \rvz_j}\left(\mQ_{ij}(\rvz)p_s(\rvz)\right) = \vf_i(\rvz)p_s(\rvz) - \sum_j \frac{\partial}{\partial \rvz_j}\left(\mD_{ij}(\rvz)p_s(\rvz)\right)
    \label{eqn:p1}
\end{equation}
Furthermore, from the Fokker-Planck formalism,
\begin{align}
    \frac{\partial p_t(\rvz)}{\partial t} &= -\sum_i \frac{\partial}{\partial z_i}\left(\vf_i(\rvz)p_t(\rvz)\right) + \sum_{i,j} \frac{\partial^2}{\partial \rvz_i\rvz_j} \left(\mD_{ij}(\rvz)p_t(\rvz)\right) \\
    &= -\sum_i \frac{\partial}{\partial z_i}\Bigg[\vf_i(\rvz)p_t(\rvz) - \sum_j \frac{\partial}{\partial \rvz_j} \left(\mD_{ij}(\rvz)p_t(\rvz)\right)\Bigg]
\end{align}
For $p_t(\rvz) = p_s(\rvz)$, we have,
\begin{equation}
    \sum_i \frac{\partial}{\partial z_i}\Bigg[\vf_i(\rvz)p_t(\rvz) - \sum_j \frac{\partial}{\partial \rvz_j} \left(\mD_{ij}(\rvz)p_t(\rvz)\right)\Bigg] = 0
    \label{eqn:p2}
\end{equation}

Denoting the Fourier transform of $\mQ(\rvz)p_s(\rvz)$ as $\hat{\mQ}(\vk)$ and the Fourier transform of $\vf_i(\rvz)p_t(\rvz) - \sum_j \frac{\partial}{\partial \rvz_j} \left(\mD_{ij}(\rvz)p_t(\rvz)\right)$ by $\hat{\mF}(\vk)$, then from Eqns. \ref{eqn:p1} and \ref{eqn:p2} we have the following equations in the Fourier-space:
\begin{align}
    2\pi i \hat{\mQ}\vk = \hat{\mF} \\
    \vk^T \hat{\mF} = 0
\end{align}
Therefore, it implies that the matrix $\hat{\mQ}$ is a projection matrix from $\vk$ to the span of $\hat{\mF}$. Consequently, the matrix $\hat{\mQ}$ can be constructed as: $\hat{\mQ} = (2\pi i)^{-1}\frac{\hat{\mF}\vk^T}{\vk^T\vk} - (2\pi i)^{-1}\frac{\vk\hat{\mF}^T}{\vk^T\vk}$. This construction also shows that $\hat{\mQ}$ is skew-symmetric. Moreover, the skew-symmetric $\mQ$ can be obtained by computing the Inverse-Fourier transform of $(p_s(\rvz))^{-1}\hat{\mQ}$
\end{proof}

\subsection{Existing SGMs parameterized using the SGM recipe}
\label{app:A_3}
In this section, we provide examples of SGMs that can be cast under the recipe proposed in Section 2.2. It is worth noting that under the completeness framework proposed in Section 2.2, given a positive semi-definite diffusion matrix $\mD(\rvz)$, a skew-symmetric curl matrix $\mQ(\rvz)$ and the Hamiltonian $H(\rvz)$ corresponding to a specified target distribution $p_s(\rvz)$, the forward process SDE can be parameterized in terms of the target distribution as follows:
\begin{equation}
    H(\rvz) = U(\rvx) + \frac{\rvm^T M^{-1} \rvm}{2}, \quad \nabla H(\rvz) = \begin{pmatrix} \nabla U(\rvx) \\ M^{-1}\rvm\end{pmatrix}
\end{equation}
\begin{equation}
    \vf(\rvz) = -(\mD(\rvz) + \mQ(\rvz))\begin{pmatrix}\nabla U(\rvx) \\M^{-1}\rvm\end{pmatrix} + \tau(\rvz)
    \label{eqn:app_recipe}
\end{equation}
\begin{equation}
    d\rvz = \vf(\rvz) dt + \sqrt{2\mD(\rvz)}d\rvw_t \label{eqn:3_3} \\
\end{equation}

We now recast several existing SGMs under this framework:

\subsubsection{Non-augmented SGMs}
\label{app:A_3_1}
For SGMs with a non-augmented form, we assume auxiliary variables $\rvm_t=0$ with the equilibrium distribution given by $p_s(\rvz) = \mathcal{N}(\bf{0}_d, \mI_d)$. The Hamiltonian and its gradient can then be specified as follows:
\begin{align}
    H(\rvz) &= \frac{\rvx^T \rvx}{2} \,,\qquad
    \nabla H(\rvz) = \rvx
\end{align}
For the choice of $\mD_{\text{VP}}(\rvz) = \frac{\beta_t}{2}\mI_d$ and $\mQ_{\text{VP}}(\rvz) = \mathbf{0}_d$, the drift for the forward SDE defined in Eqn. \ref{eqn:app_recipe} reduces to the following form:
\begin{equation}
    d\rvx = -\frac{\beta_t}{2}\rvx dt + \sqrt{\beta_t}d\rvw_t
    \label{eqn:vpsde}
\end{equation}
where $\beta_t$ is a time-dependent constant. The forward SDE in Eqn. \ref{eqn:vpsde} is the same as the VP-SDE proposed in \cite{songscore}. From our recipe, the stationary distribution for the VPSDE should be $p_s(\rvx) = \mathcal{N}(\mathbf{0}_d, \mI_d)$. Indeed the perturbation kernel for the VP-SDE is specified as follows:
{
\begin{equation}
    p(\rvx_t|\rvx_0) = \mathcal{N}(\rvx_0 e^{-\frac{1}{2}\int_0^t \beta(s) ds}, (1 - e^{-\int_0^t \beta(s) ds})^2\mI_d)
\end{equation}
}

which converges to a standard Gaussian distribution as $t \to \infty$. This example suggests that the proposed recipe can be used to establish the validity of the convergence of a forward process with a specified stationary distribution $p_s(\rvz)$ without deriving the perturbation kernel or relying on physical intuition.
Interestingly, the Variance-Exploding (VE) SDE \citep{songscore} is one example that cannot be cast in our framework. This would mean that it will not asymptotically converge to the standard Gaussian distribution at equilibrium. Indeed, this can be confirmed from the analytical form of the perturbation kernel of the VE-SDE given by:
\begin{equation}
    p(\rvx_t|\rvx_0) = \mathcal{N}(\rvx_0, [\sigma^2(t) - \sigma^2(0)]\mI_d)
\end{equation}
As $t \to \infty$, the variance of the perturbation kernel of the VE-SDE grows unbounded and therefore does not converge to the equilibrium distribution $\mathcal{N}(\mathbf{0}_d, \mI_d)$. This should not be surprising since the VE-SDE, for the specified Hamiltonian, could not be recast in the completeness framework, to begin with.

\subsubsection{Augmented SGMs}
\label{app:A_3_2}
For SGMs with an augmented state-space (data state space $\rvx_t$ + auxiliary variables $\rvm_t$), we assume the equilibrium distribution $p_s(\rvz) = \mathcal{N}(\bf{0}_d, \mI_d)\mathcal{N}(\bf{0}_d, M\mI_d)$. The Hamiltonian and its gradient can then be specified as follows:
\begin{align}
    H(\rvz) &= \frac{\rvx^T \rvx}{2} + \frac{\rvm^T M^{-1} \rvm}{2} \,,\qquad
    \nabla H(\rvz) = \begin{pmatrix} \rvx \\ M^{-1}\rvm\end{pmatrix}
\end{align}
For this choice of $H$, the forward SDE representative of \mn can be obtained by choosing the following $\mD$ and $\mQ$ matrices:
\begin{equation}
    \mD_{\text{\nsmn}} = \frac{\beta}{2}\left(\begin{pmatrix} \Gamma & 0 \\ 0 & M\gamma \end{pmatrix} \otimes \mI_d\right)\quad\quad \mQ_{\text{\nsmn}} = \frac{\beta}{2}\left(\begin{pmatrix} 0 & -1 \\ 1 & 0 \end{pmatrix}\otimes \mI_d\right)
\end{equation}

Similarly, the forward SDE representative of CLD \citep{dockhornscore} can be obtained by choosing:
\begin{equation}
    \mD_{\text{CLD}} = \beta\left(\begin{pmatrix} 0 & 0 \\ 0 & \Gamma \end{pmatrix} \otimes \mI_d\right) \quad\quad \mQ_{\text{CLD}} = \beta\left(\begin{pmatrix} 0 & -1 \\ 1 & 0 \end{pmatrix}\otimes \mI_d\right)
\end{equation}

Since both \mn and CLD can be shown to converge asymptotically to $p_s(\rvz)$ from the analytical form of their perturbation kernels $p(\rvz_t|\rvz_0)$, the result from our completeness framework is valid. More importantly, given a forward process for an SGM, our recipe can be used to validate if the SGM converges to a specified equilibrium distribution without the need for analytically determining the perturbation kernel (which is usually non-trivial).

%% file: tex/appendix/App_B.tex
\section{\fmn}
\label{app:B}
In this section, we elaborate on several aspects of \nsmn, which were discussed briefly in the main text. Moreover, we work with the following form of the forward process for \nsmn:
\begin{equation}
    \begin{pmatrix}d\rvx_t \\ d\rvm_t\end{pmatrix} = \left(\frac{\beta_t}{2}\begin{pmatrix} -\Gamma & M^{-1}\\ -1 & - \nu \end{pmatrix} \otimes \mI_d\right) \,\begin{pmatrix}
        \rvx_t \\
        \rvm_t
    \end{pmatrix} dt + \left(\begin{pmatrix} \sqrt{\Gamma\beta_t} &0 \\ 0 &\sqrt{M\nu\beta_t} \end{pmatrix} \otimes \mI_d \right)d\rvw_t,
    \label{eqn:b_1}
\end{equation}

It is worth noting that the form of the forward process defined in Eqn. \ref{eqn:b_1} is more general than Eqn. 12 in the sense that we consider a time-dependent $\beta_t$ here for our discussions. We can then reason about the forward SDE in Eqn. 12 by fixing $\beta_t$ to a time-independent quantity $\beta$ for all subsequent analyses.
\subsection{Critical Damping in \nsmn}
\label{app:B_1}
Assuming $\beta_t=1$ and $\rvx_t, \rvm_t \in \mathbb{R}$ for simplicity, the equations of motion for the deterministic dynamics can be specified as follows:
\begin{equation}
    \frac{dx_t}{dt} = -\Gamma x_t + M^{-1}m_t \label{eqn:b_2}
\end{equation}
\begin{equation}
    \frac{dm_t}{dt} = -x_t - \nu m_t
\end{equation}
From Eqn. \ref{eqn:b_2}, we have:
\begin{equation}
m_t=M\left(\frac{d x_{t}}{d t}+\Gamma x_{t}\right)
\end{equation}
Furthermore, taking the derivative of both sides in Eqn. \ref{eqn:b_2}, we have:
\begin{align}
\frac{d^{2} x_{t}}{d t^{2}} &= -\Gamma \frac{d x_{t}}{d t}+M^{-1} \frac{d m_{t}}{d t} \\
&= -\Gamma \frac{d x_{t}}{d t}+M^{-1}\left[-x_{t}-\nu m_{t}\right]\\
&= -\Gamma \frac{d x_{t}}{d t}+ M^{-1}\left[-x_{t}-\nu M\left(\frac{d x_{t}}{d t}+\Gamma x_{t}\right)\right]
\end{align}
\begin{align}
&= -\Gamma \frac{d x_{t}}{d t} -M^{-1} x_{t}-\nu\left(\frac{d x_{t}}{d t}+\Gamma x_{t}\right) \\
&= -\Gamma \frac{d x_{t}}{d t} -M^{-1} x_{t}-\nu \frac{d x_{t}}{d t}-\Gamma \nu x_{t}\\
&= -(\Gamma + \nu) \frac{d x_{t}}{d t} -M^{-1} x_{t}-\Gamma \nu x_{t}
\end{align}
We, therefore, have the following dynamical equation in terms of the position:
\begin{equation}
    \frac{d^{2} x_{t}}{d t^{2}} + (\Gamma + \nu) \frac{d x_{t}}{d t} + (M^{-1} +\Gamma \nu) x_{t} = 0
\end{equation}
Assuming the exponential \textit{ansatz} $\rvx_t = \exp{(-\lambda t)}$ and plugging into the above ODE, we have the following result:

\begin{equation}
    \exp{(-\lambda t)}\left[\lambda^2 -(\Gamma + \nu)\lambda + (M^{-1} +\Gamma \nu) \right] = 0
\end{equation}
which implies,
\begin{equation}
    \lambda^2 -(\Gamma + \nu)\lambda + (M^{-1} +\Gamma \nu) = 0
\end{equation}
\begin{equation}
    \lambda = \frac{(\Gamma + \nu) \pm \sqrt{(\Gamma + \nu)^2 - 4M^{-1} - 4\Gamma\nu}}{2}
\end{equation}
\begin{equation}
    \lambda = \frac{(\Gamma + \nu) \pm \sqrt{(\Gamma - \nu)^2 - 4M^{-1}}}{2}
\end{equation}

Corresponding to the value of $\nu, \Gamma$ and $M$, we can now have the following damping conditions:\\\\
\textbf{(i)} $(\Gamma - \nu)^2 < 4M^{-1}$ corresponds to \textit{Underdamped dynamics}

\textbf{(ii)} $(\Gamma - \nu)^2 = 4M^{-1}$ corresponds to \textit{Critical damping}

\textbf{(iii)} $(\Gamma - \nu)^2 > 4M^{-1}$ corresponds to \textit{Overdamped dynamics}

Moreover when $\Gamma=0$ and $\bar{\nu} = M\nu$, we get: $\bar{\nu}^2 = 4M$ which is the critical damping condition proposed in \citet{dockhornscore}. Therefore, similar to \citet{dockhornscore}, we work in the Critical Damping regime specified by the condition $(\Gamma - \nu)^2 = 4M^{-1}$

\subsection{\mn Training}
\label{app:B_2}
\subsubsection{Overall Training Framework in \nsmn}
\label{app:B_2_1}
Following the derivation in \citet{dockhornscore}, the maximum likelihood training formulation for score matching can be specified as follows. Let $p_0$, $q_0$ be two densities with corresponding marginal densities $p_t$ and $q_t$ (for forward diffusion using \mn defined in Eqn. \ref{eqn:b_1}) at time t. As shown in \citet{song2021maximum}, the KL-Divergence between $p_0$ and $q_0$ can then be expressed as a mixture of score-matching losses over multiple time scales as follows: 
\begin{equation}
\begin{split}
    \infdiv{p_0}{q_0} &= \infdiv{p_0}{q_0} - \infdiv{p_T}{q_T} + \infdiv{p_T}{q_T} \\
    &= - \int_0^T \frac{\partial\infdiv{p_t}{q_t}}{\partial t}dt + \infdiv{p_T}{q_T} 
\end{split}
\label{eqn:kl_objective}
\end{equation}
Following the derivation from \citet{song2021maximum}, the Fokker-Planck equation describing the time evolution of the probability density function of the SDE in Eqn. \ref{eqn:b_1} can be expressed as follows:
\begin{equation}
\begin{split}
    \frac{\partial p_t(\rvz_t)}{\partial t} &= \nabla_{\rvz_t} \cdot \left[\tfrac{1}{2} \left(G(t) G(t)^\top \otimes \mI_d\right) \nabla_{\rvz_t} p_t(\rvz_t) - p_t(\rvz_t) (f(t) \otimes \mI_d) \rvz_t \right] \\
    &= \nabla_{\rvz_t} \cdot \left[\vh_p(\rvz_t, t) p_t(\rvz_t)\right]
\end{split}
\end{equation}
where
\begin{equation}
\vh_p(\rvz_t, t) \coloneqq \tfrac{1}{2} \left(G(t) G(t)^\top  \otimes \mI_d\right) \nabla_{\rvz_t} \log p_t(\rvz_t) - (f(t) \otimes \mI_d) \rvz_t
\end{equation}
\begin{equation}
    f(t) =  \left(\frac{\beta_t}{2}\begin{pmatrix} -\Gamma & M^{-1}\\ -1 & - \nu \end{pmatrix} \right)
\end{equation}
\begin{equation}
    G(t) = \begin{pmatrix} \sqrt{\Gamma\beta_t} &0 \\ 0 &\sqrt{M\nu\beta_t} \end{pmatrix}
\end{equation}
Further assuming that $\log p_t(\rvz_t)$ and $\log q_t(\rvz_t)$ are smooth functions with at most polynomial growth at infinity, we have
\begin{align}
    \lim_{\rvz_t \to \infty} \vh_p(\rvz_t, t) p_t(\rvz_t) = \lim_{\rvz_t \to \infty} \vh_q(\rvz_t, t) q_t(\rvz_t) = 0.
\end{align}
Using the above fact, we can compute the time-derivative of the Kullback--Leibler divergence between $p_t$ and $q_t$ as 
\begin{equation}
\begin{split}
    \frac{\partial\infdiv{p_t}{q_t}}{\partial t} &= \frac{\partial}{\partial t} \int p_t(\rvz_t) \log \frac{p_t(\rvz_t)}{q_t(\rvz_t)} \, d\rvz_t \\
    &= \int \frac{\partial p_t(\rvz_t)}{\partial t} \log \frac{p_t(\rvz_t)}{q_t(\rvz_t)} \, d\rvz_t - \int \frac{p_t(\rvz_t)}{q_t(\rvz_t)}\frac{\partial q_t(\rvz_t)}{\partial t}d\rvz_t \\
    &= - \int p_t(\rvz_t) \Big[\vh_p(\rvz_t, t) -  \vh_q(\rvz_t, t)\Big]^\top \Big[\nabla_{\rvz_t} \log p_t(\rvz_t) - \nabla_{\rvz_t} \log q_t(\rvz_t)\Big] \, d\rvz_t \\
    &= - \frac{1}{2} \int p_t(\rvz_t) \Big[\nabla_{\rvz_t} \log p_t(\rvz_t) - \nabla_{\rvz_t} \log q_t(\rvz_t)\Big]^\top \Big(G(t) G(t)^\top \otimes \mI_d\Big) \\ & \qquad\qquad \Big[\nabla_{\rvu_t} \log p_t(\rvz_t) - \nabla_{\rvz_t} \log q_t(\rvz_t) \Big] \, d\rvz_t \\
    &= -\frac{1}{2} \int p_t(\rvz_t) \Big[\Gamma\beta_t\Vert \nabla_{\rvx_t} \log p_t(\rvz_t) - \nabla_{\rvx_t} \log q_t(\rvz_t) \Vert_2^2 + \\&\qquad\qquad M\nu\beta_t\Vert \nabla_{\rvm_t} \log p_t(\rvz_t) - \nabla_{\rvm_t} \log q_t(\rvz_t) \Vert_2^2\Big] \, d\rvz_t
\end{split}
\label{eqn:b2_1}
\end{equation}
Assuming our generative prior $p(x_T)$ matches the equilibrium state of the forward process closely i.e. $\infdiv{p_T}{q_T} \approx 0$ and substituting the result in Eqn. \ref{eqn:b2_1} in Eqn. \ref{eqn:kl_objective}, we get the following score-matching objective corresponding to the maximum-likelihood objective in Eqn. \ref{eqn:kl_objective} as follows:
\begin{equation}
\begin{split}
    \infdiv{p_0}{q_0} = \frac{1}{2}\mathbb{E}_{t \sim \mathcal{U}(0,T)} \mathbb{E}_{\rvz_t \sim p_t(\rvz_t)}&\Big[\underbrace{\Gamma\beta_t\big\Vert \nabla_{\rvx_t} \log p_t(\rvz_t) - \nabla_{\rvx_t} \log q_t(\rvz_t) \big\Vert_2^2}_{\text{Data-Space}} + \\ &\qquad\underbrace{M\nu\beta_t\big\Vert \nabla_{\rvm_t} \log p_t(\rvz_t) - \nabla_{\rvm_t} \log q_t(\rvz_t) \big\Vert_2^2}_{\text{Momentum-Space}}\Big]
    \label{eqn:b2_2}
\end{split}
\end{equation}

In general, the above score-matching loss can be re-formulated using arbitrary loss weightings $\lambda_1(t)$ and $\lambda_2(t)$ as follows:
\begin{equation}
\begin{split}
    \infdiv{p_0}{q_0} = \frac{1}{2}\mathbb{E}_{t \sim \mathcal{U}(0,T)} \mathbb{E}_{\rvz_t \sim p_t(\rvz_t)}&\Big[\lambda_1(t)\big\Vert \nabla_{\rvx_t} \log p_t(\rvz_t) - \nabla_{\rvx_t} \log q_t(\rvz_t) \big\Vert_2^2 + \\& \qquad\lambda_2(t)\big\Vert \nabla_{\rvm_t} \log p_t(\rvz_t) - \nabla_{\rvm_t} \log q_t(\rvz_t) \big\Vert_2^2\Big]
\end{split}
\end{equation}
Choosing the same weighting for both loss components i.e. $\lambda_1(t) = \lambda_2(t) = \lambda(t)$, the score-matching objective in Eqn. \ref{eqn:b2_2} can be simplified as follows:
\begin{equation}
    \mathcal{L}_{\text{SM}} = \frac{1}{2}\mathbb{E}_{t \sim \mathcal{U}(0,T)} \mathbb{E}_{\rvz_t \sim p_t(\rvz_t)}\Big[\lambda(t)\big\Vert \nabla_{\rvz_t} \log p_t(\rvz_t) - \nabla_{\rvz_t} \log q_t(\rvz_t) \big\Vert_2^2\Big]
\end{equation}
Approximating the score $\nabla_{\rvz_t} \log q_t(\rvz_t)$ using a parametric estimator $\vs_{\theta}(\rvz_t, t)$ and following \citet{6795935}, it can be shown that the $\mathcal{L}_{\text{SM}}$ objective is equivalent to the following Denoising Score Matching (DSM) objective:
\begin{equation}
    \mathcal{L}_{\text{DSM}} = \frac{1}{2}\mathbb{E}_{t \sim \mathcal{U}(0,T)}\mathbb{E}_{\rvz_0 \sim p(\rvz_0)} \mathbb{E}_{\rvz_t \sim p_t(\rvz_t|\rvz_0)}\Big[\lambda(t)\big\Vert \nabla_{\rvz_t} \log p_t(\rvz_t | \rvz_0) - \vs_{\theta}(\rvz_t, t) \big\Vert_2^2\Big]
\end{equation}
Moreover, \citet{dockhornscore} propose to use the following objective a.k.a. Hybrid Score Matching (HSM), which is equivalent to the DSM objective (upto a constant independent of $\theta$):
\begin{equation}
    \mathcal{L}_{\text{HSM}} = \frac{1}{2}\mathbb{E}_{t \sim \mathcal{U}(0,T)} \mathbb{E}_{\rvx_0 \sim p(\rvx_0)}\mathbb{E}_{\rvz_t \sim p_t(\rvz_t|\rvx_0)}\Big[\lambda(t)\big\Vert \nabla_{\rvz_t} \log p_t(\rvz_t | \rvx_0) - \vs_{\theta}(\rvz_t, t) \big\Vert_2^2\Big]
    \label{eqn:b2_3}
\end{equation}
The perturbation kernels $p(\rvz_t|\rvz_0)$ and $p(\rvz_t|\rvx_0)$ can be computed analytically for an SDE with affine drift (See Appendix \ref{app:B_3} for the exact analytical forms of the perturbation kernel for \nsmn). Following \citet{dockhornscore}, we use the Hybrid Score Matching (HSM) objective throughout this work. We next discuss the computation of the analytical form of the target score $\nabla_{\rvz_t} \log p_t(\rvz_t | \rvx_0)$ (or $\nabla_{\rvz_t} \log p_t(\rvz_t | \rvz_0)$ for DSM) and the parameterization of our score network $\vs_{\theta}(\rvz_t, t)$.

\subsubsection{Analytical Score Computation and Parameterization}
\label{app:B_2_2}
In cases when the perturbation kernels are multivariate Gaussian distributions of the form $\mathcal{N}(\vmu_t, \mSigma_t)$, the target score $\nabla_{\rvz_t} \log p_t(\rvz_t | \rvx_0)$  can be computed analytically as follows:
\begin{align}
    \nabla_{\rvz_t}\log{p(\rvz_t|\rvx_0)} &= -\mSigma_t^{-1}(\rvz_t - \vmu_t)\\
    &=-\mL_t^{-T}\mL_t^{-1}(\mL_t \bm{\epsilon}) = -\mL_t^{-T}\bm{\epsilon}
    \label{eqn:b2_4}
\end{align}
where $\bm{\epsilon} \sim \mathcal{N}(\mathbf{0}_{2d}, \mI_{2d})$ and $\mSigma_t = \mL_t\mL_t^T$ is the Cholesky decomposition. Moreover, for $\mSigma_t = \left(\begin{pmatrix}
    \Sigma_{t}^{xx} & \Sigma_{t}^{xm} \\ \Sigma_{t}^{xm} & \Sigma_{t}^{mm}
\end{pmatrix}\otimes \mI_d\right)$ (as is the case in \nsmn), the Cholesky decomposition can be computed analytically as follows:
\begin{equation}
    \mL_t = \left(\begin{pmatrix}
    L_{t}^{xx} & 0 \\ L_{t}^{xm} & L_{t}^{mm}
\end{pmatrix}\otimes \mI_d\right)
\end{equation}
\begin{equation}
    L_t = \begin{pmatrix}
    L_{t}^{xx} & 0 \\ L_{t}^{xm} & L_{t}^{mm}
\end{pmatrix} = \begin{pmatrix}
    \sqrt{\Sigma^{xx}_t} & 0 \\ \frac{\Sigma^{xm}_t}{\sqrt{\Sigma^{xx}_t}} & \sqrt{\frac{\Sigma^{xx}_t \Sigma^{mm}_t - \left(\Sigma^{xm}_t\right)^2}{\Sigma^{xx}_t}}
\end{pmatrix}
\end{equation}
Consequently,
\begin{equation}
\begin{split}
    \mL^{-T}_t &= L^{-T}_t \otimes \mI_d \\
    &= \begin{pmatrix} \sqrt{\Sigma^{xx}_t} & \frac{\Sigma^{xm}_t}{\sqrt{\Sigma^{xx}_t}} \\ 0 & \sqrt{\frac{\Sigma^{xx}_t \Sigma^{mm}_t - \left(\Sigma^{xm}_t\right)^2}{\Sigma^{xx}_t}}\end{pmatrix}^{-1} \otimes \mI_d \\
    &= \begin{pmatrix} \frac{1}{\sqrt{\Sigma^{xx}_t }} &\frac{-\Sigma^{xm}_t}{\sqrt{\Sigma^{xx}_t}\sqrt{\Sigma^{xx}_t \Sigma^{mm}_t - \left(\Sigma^{xm}_t\right)^2}} \\ 0 &\sqrt{\frac{\Sigma^{xx}_t}{\Sigma^{xx}_t \Sigma^{mm}_t - \left(\Sigma^{xm}_t\right)^2}}\end{pmatrix} \otimes \mI_d.
\end{split}
\end{equation}
Plugging the analytical form of $\mL_t^{-T}$ into Eqn. \ref{eqn:b2_4}, we get the following analytical form of the target score $\nabla_{\rvz_t} \log p_t(\rvz_t | \rvx_0)$:
\begin{align}
    \nabla_{\rvz_t} \log p_t(\rvz_t | \rvx_0) &= -\mL_t^{-T}\bm{\epsilon} \\
    &= -\left(\begin{pmatrix}
    l_{t}^{xx} & l_t^{xm} \\ 0 & l_{t}^{mm}
\end{pmatrix} \otimes \mI_d\right) \begin{pmatrix}
    \bm{\epsilon}_x \\ \bm{\epsilon}_m
\end{pmatrix} \\
    &= -\begin{pmatrix}
    l_{t}^{xx}\bm{\epsilon_x} + l_t^{xm} \bm{\epsilon_m} \\ l_{t}^{mm} \bm{\epsilon}_m
\end{pmatrix} \label{eqn:b2_5}
\end{align}
where $l_t^{xx} = \frac{1}{\sqrt{\Sigma^{xx}_t }}$, $l_t^{xm} = \frac{-\Sigma^{xm}_t}{\sqrt{\Sigma^{xx}_t}\sqrt{\Sigma^{xx}_t \Sigma^{mm}_t - \left(\Sigma^{xm}_t\right)^2}}$ and $l_t^{mm} = \sqrt{\frac{\Sigma^{xx}_t}{\Sigma^{xx}_t \Sigma^{mm}_t - \left(\Sigma^{xm}_t\right)^2}}$. While one can directly model the score as defined in Eqn. \ref{eqn:b2_5}, we instead parameterize the score network as $\vs_{\theta}(\rvz_t, t) = -\mL_t^{-T}\bm{\epsilon}_{\theta}(\rvz_t, t)$.

\subsubsection{Putting it all together}
\label{app:B_2_3}
Plugging the analytical form of $\nabla_{\rvz_t} \log p_t(\rvz_t | \rvx_0)$ and our score network parameterization $\vs_{\theta}(\rvz_t, t) = -\mL_t^{-T}\bm{\epsilon}_{\theta}(\rvz_t, t)$ in the HSM objective in Eqn. \ref{eqn:b2_3}, we get the following objective:
\begin{align}
    \mathcal{L}_{\text{HSM}} &= \frac{1}{2}\mathbb{E}_{t \sim \mathcal{U}(0,T)} \mathbb{E}_{\rvx_0 \sim p(\rvx_0)}\mathbb{E}_{\rvz_t \sim p_t(\rvz_t|\rvx_0)}\Big[\lambda(t)\big\Vert \nabla_{\rvz_t} \log p_t(\rvz_t | \rvx_0) - \vs_{\theta}(\rvz_t, t) \big\Vert_2^2\Big]\\
    &= \frac{1}{2}\mathbb{E}_{t \sim \mathcal{U}(0,T)} \mathbb{E}_{\rvx_0 \sim p(\rvx_0)} \mathbb{E}_{\bm{\epsilon} \sim \mathcal{N}(\bm{0}_{2d}, \mI_{2d})}\Big[\lambda(t)\big\Vert \mL^{-T}_t\bm{\epsilon} - \mL_t^{-T}\bm{\epsilon}_{\theta}(\vmu_t + \mL_t\bm{\epsilon}, t) \big\Vert_2^2\Big]\\
    &\leq \frac{1}{2}\mathbb{E}_{t \sim \mathcal{U}(0,T)} \mathbb{E}_{\rvx_0 \sim p(\rvx_0)} \mathbb{E}_{\bm{\epsilon} \sim \mathcal{N}(\bm{0}_{2d}, \mI_{2d})}\Big[\lambda(t)\big\Vert \mL_t^{-T} \big\Vert_2^2\big\Vert \bm{\epsilon} - \bm{\epsilon}_{\theta}(\vmu_t + \mL_t\bm{\epsilon}, t) \big\Vert_2^2\Big] \label{eqn:b2_6}
\end{align}
It is worth noting that since our original HSM objective is upper bounded by the objective in Eqn. \ref{eqn:b2_6}, minimizing the latter also minimizes $\mathcal{L}_{\text{HSM}}$. Since we optimize for sample quality, following prior work \citep{songscore, dockhornscore}, we choose $\lambda(t) = \frac{1}{\Vert \mL_t^{-T} \Vert_2^2}$ to cancel the weighting induced by $\Vert \mL_t^{-T} \Vert_2^2$. Our final training objective reduces to the following \textit{noise-prediction} formulation: 

\begin{equation}
    \mathcal{L}(\theta) = \frac{1}{2}\mathbb{E}_{t \sim \mathcal{U}(0,T)} \mathbb{E}_{\rvx_0 \sim p(\rvx_0)}\mathbb{E}_{\bm{\epsilon} \sim \mathcal{N}(\bm{0}_{2d}, \mI_{2d})}\Big[\big\Vert \bm{\epsilon} - \bm{\epsilon}_{\theta}(\vmu_t + \mL_t\bm{\epsilon}, t) \big\Vert_2^2\Big] 
\end{equation}

It is worth noting that in our training setup, we need to predict the full $2d$-dimensional $\bm{\epsilon}$. This is in contrast to the training setup in CLD, where we only need to predict the last-d components i.e. $\bm{\epsilon}_{d:2d}$ of the noise vector $\bm{\epsilon}$. This difference in training arises due to different formulations of the diffusion coefficient in \mn and CLD. Indeed, setting $\Gamma=0$ in Eqn. \ref{eqn:b2_2} would result in a similar training objective as in CLD.
\subsection{Perturbation Kernel in \nsmn}
\label{app:B_3}
We now present the analytical form of the perturbation kernels $p(\rvz_t|\rvz_0)$ and $p(\rvz_t|\rvx_0)$ for \nsmn, which are required for training using DSM or HSM respectively.
\subsubsection{Mean and Variance of $p(\rvz_t|\rvz_0)$}
\label{app:B_3_1}
Since the drift and the diffusion coefficients in Eqn. \ref{eqn:b_1} are affine, the perturbation kernel $p(\rvz_t|\rvz_0)$ will be a multivariate Gaussian distribution $\mathcal{N}(\vmu_t, \mSigma_t)$. Following \citet{sarkka2019applied}, $\vmu_t$ and $\mSigma_t$, evolve as the following ODEs:
\begin{equation}
    \frac{d\vmu_t}{dt} = \mF(t)\vmu_t
\end{equation}
\begin{equation}
    \frac{d\mSigma_t}{dt} = \mF(t)\mSigma_t + \mSigma_t\mF^T(t) + \mG(t)\mG(t)^T
\end{equation}

where $\mF(t) = \left(\frac{\beta_t}{2}\begin{pmatrix} -\Gamma & M^{-1}\\ -1 & - \nu \end{pmatrix} \otimes \mI_d\right)$ and $\mG(t) = \left(\begin{pmatrix} \sqrt{\Gamma\beta_t} &0 \\ 0 &\sqrt{M\nu\beta_t} \end{pmatrix} \otimes \mI_d \right)$ for \nsmn. Under critical damping i.e. $M^{-1} = \frac{(\Gamma - \nu)^2}{4}$, solving the ODEs for the mean and variance yields the following form of $p(\rvz_t|\rvz_0) = \gN(\vmu_t, \mSigma_t)$:

\begin{equation}
    \vmu_t = \begin{pmatrix}\vmu_t^x \\ \vmu_t^m\end{pmatrix} = \begin{pmatrix} A_1 \gB(t)\vx_0 + A_2\gB(t)\vm_0 + \vx_0\\ C_1\gB(t)\vx_0 + C_2\gB(t)\vm_0 + \vm_0\end{pmatrix}e^{-(\frac{\nu + \Gamma}{4}) \gB(t)}
\end{equation}
where $\mathcal{B}(t) = \int_0^t \beta(s)ds$ and coefficients:
\begin{equation}
    A_1 = \frac{\nu - \Gamma}{4} \quad\quad A_2 = \frac{(\Gamma - \nu)^2}{8}
\end{equation}
\begin{equation}
    C_1 = \frac{-1}{2} \quad\quad C_2 = \frac{\Gamma - \nu}{4}
\end{equation}

The variance $\Sigma_t$ for the perturbation kernel $p(\rvz_t)\rvz_0$ is given by:

\begin{equation}
    \mSigma_t = \left(\begin{pmatrix} \Sigma^{xx}_t & \Sigma^{xm}_t \\ \Sigma^{xm}_t & \Sigma^{mm}_t \end{pmatrix}e^{-(\frac{\Gamma + \nu}{2}) \gB(t)}\right) \otimes \mI_d
\end{equation}
where,
\begin{align}
    \Sigma^{xx}_t &= A_1 \gB^2(t)\Sigma^{xx}_0 + A_2 \gB^2(t)\Sigma^{mm}_0 + A_3 \gB(t)\Sigma^{xx}_0 + A_4 \gB^2(t) + A_5 \gB(t) + (e^{2\lambda \gB(t)} - 1) + \Sigma^{xx}_0\\
    \Sigma^{xm}_t &= C_1 \gB^2(t)\Sigma^{xx}_0 + C_2 \gB^2(t)\Sigma^{mm}_0 + C_3 \gB(t)\Sigma^{xx}_0 + C_4 \gB(t)\Sigma^{mm}_0 + C_5 \gB^2(t)\\
    \Sigma^{mm}_t &= D_1 \gB^2(t)\Sigma^{xx}_0 + D_2 \gB^2(t)\Sigma^{mm}_0 + D_3 \gB(t)\Sigma^{mm}_0 + D_4 \gB^2(t) + D_5 \gB(t) + M(e^{2\lambda B(t)} - 1) + \Sigma^{mm}_0
\end{align}
where $\mSigma_0 = \begin{pmatrix}
    \Sigma_0^{xx} & 0 \\0 & \Sigma_0^{mm}
\end{pmatrix}$, $\mathcal{B}(t) = \int_0^t \beta(s)ds$ and coefficients:
\begin{equation}
    A_1 = \frac{M^{-1}}{4} \quad\quad A_2 = \frac{M^{-2}}{4} \quad\quad A_3 = \frac{\nu - \Gamma}{2} \quad\quad A_4 = \frac{-M^{-1}}{2} \quad\quad A_5 = \frac{\Gamma - \nu}{2}
\end{equation}
\begin{equation}
    C_1 = \frac{\Gamma - \nu}{8} \quad\quad C_2 = \frac{(\Gamma - \nu)^3}{32} \quad\quad C_3 = \frac{-1}{2} \quad\quad C_4 = \frac{M^{-1}}{2} \quad\quad C_5 = \frac{\nu - \Gamma}{4}
\end{equation}
\begin{equation}
    D_1 = \frac{1}{4} \quad\quad D_2 = \frac{M^{-1}}{4} \quad\quad D_3 = \frac{\Gamma - \nu}{2} \quad\quad D_4 = \frac{-1}{2} \quad\quad D_5 = \frac{M(\nu - \Gamma)}{2}
\end{equation}

It is worth noting that, when $\Gamma=0$, $\bar{\nu} = M\nu$ and $\bar{\beta}(t) = \frac{\beta(t)}{2}$ such that $\gB(t) = 2\bar{\gB}(t)$ where $\bar{\gB}(t) = \int_0^t \bar{\beta}(s)ds$, we have the following form of the mean $\mu_t$:
\begin{equation}
    \mu_t = \begin{pmatrix} 2\bar{\nu}^{-2} \bar{\gB}(t)\vx_0 + 4\bar{\nu}^{-2}\bar{\gB}(t)\vm_0 + \vx_0\\ -\bar{\gB}(t)\vx_0 - 2\bar{\nu}^{-1}\bar{\gB}(t)\vm_0 + \vm_0\end{pmatrix}e^{-2\bar{\nu}^{-1} \bar{\gB}(t)}
    \label{eqn:b2_7}
\end{equation}
The expression for $\mu_t$ in Eqn. \ref{eqn:b2_7} is exactly the same as the mean of the perturbation kernel for CLD (Refer to Appendix B.1 in \citet{dockhornscore}). A similar analysis holds for the variance $\Sigma_t$ which provides more insight into CLD being a special case of \nsmn. Similar to CLD, at $t=0$, we have $\vmu_0 = \begin{pmatrix}
    \rvx_0 \\ \rvm_0
\end{pmatrix}$, where $\rvx_0 \sim p(\rvx_0)$ (a.k.a the data generating distribution) and $\rvm_0 \sim \gN(\bm{0}_d, M\gamma\mI_d)$, where $\gamma$ is a scalar hyperparameter. Similarly, for DSM training, both $\Sigma_0^{xx}$ and $\Sigma_0^{mm}$ can be set to 0 (since $\rvz_t = [\rvx_t, \rvm_t]^T$ is a sample based estimate).

\subsubsection{Mean and Variance of $p(\rvz_t|\rvx_0)$}
\label{app:B_3_2}
Since the data generating distribution for $m_0$ and the DSM perturbation kernel $p(\rvz_t|\rvz_0)$ are multivariate Gaussians, we can marginalize out the initial momentum variables $\rvm_0$ from $p(\rvz_t|\rvz_0)$ to obtain the perturbation kernel for HSM as $p(\rvz_t | \rvx_0) = \int p(\rvz_t|\rvx_0, \rvm_0)p(\rvm_0) d\rvm_0$. Consequently, the perturbation kernel $p(\rvz_t|\rvx_0)$ can be obtained by setting $m_0=\bf{0}_d$ and $\Sigma_0^{mm} = M\gamma$ in the expressions of $\vmu_t$ and $\mSigma_t$ for $p(\rvz_t|\rvz_0)$.

\subsubsection{Convergence}
\label{app:B_3_3}
As $t \to \infty$, the mean $\mu_t$ converges to $\bm{0}_{2d}$ since the multiplicative term $e^{-(\frac{\nu + \Gamma}{4}) \gB(t)}$ goes to 0. Similarly, the covariance for the perturbation kernel converges to the following case:
\begin{align}
    \Sigma_{e}^{xx} &= \lim_{t \to \infty} \Sigma_t^{xx}e^{-(\frac{\Gamma + \nu}{2}) \gB(t)} = 1 \\
    \Sigma_{e}^{xm} &= \lim_{t \to \infty} \Sigma_t^{xm}e^{-(\frac{\Gamma + \nu}{2}) \gB(t)} = 0 \\
    \Sigma_{e}^{mm} &= \lim_{t \to \infty} \Sigma_t^{mm}e^{-(\frac{\Gamma + \nu}{2}) \gB(t)} = M 
\end{align}
Therefore, the perturbation kernel converges to the following steady-state distribution $p_{\text{EQ}}(\rvz) = \gN(\rvx;\bm{0}_d, \mI_d)\gN(\rvm;\bm{0}_d, M\mI_d)$. It is worth noting that this is the exact equilibrium distribution that we specified in our SGM recipe to construct the forward process for \nsmn.
\subsection{\mn Sampling}
\label{app:B_4}
The reverse SDE analogous to the forward SDE defined in Eqn. \ref{eqn:b_1} can be formulated as follows \citep{songscore}:
\begin{equation}
    d\bar{\rvz}_t = \bar{\vf}(\bar{\rvz}_t)dt + \mG(T-t)d\bar{\rvw}_t
\end{equation}
\begin{equation}
    \bar{\vf}(\bar{\rvz}_t) = \frac{\beta_t}{2}\begin{pmatrix}
        \Gamma \bar{\rvx}_t - M^{-1}\bar{\rvm}_t + 2\Gamma \vs_{\theta}(\bar{\rvz}_t, T-t)|_{0:d}\\
        \bar{\rvx}_t + \nu \bar{\rvm}_t + 2M\nu \vs_{\theta}(\bar{\rvz}_t, T-t)|_{d:2d})
    \end{pmatrix}, \qquad \mG(T-t) = \begin{pmatrix} \sqrt{\Gamma\beta_t} &0 \\ 0 &\sqrt{M\nu\beta_t} \end{pmatrix} \otimes \mI_d
\end{equation}

where $\bar{\rvz}_t = \rvz_{T - t}$, $\bar{\rvx}_t = \rvx_{T - t}$, $\bar{\rvm}_t = \rvm_{T - t}$. Given an estimate of the score $\vs_{\vtheta}(\rvz_t, T-t)$, one can simulate the above SDE to generate data from noise. Given $\bar{\rvz}_0 = \big(\bar{\rvx}_0,  \bar{\rvm}_0\big)^T \sim p_{\text{EQ}}(\rvz) = \gN(\rvx;\bm{0}_d, \mI_d)\gN(\rvm;\bm{0}_d, M\mI_d)$, we now discuss update steps for different samplers in context of \nsmn.

\subsubsection{Euler-Maruyama (EM) Sampler}
\label{app:B_4_1}
The EM update step for the reverse SDE corresponding to \mn are as follows:
\begin{align}
    \begin{pmatrix}
        \bar{\rvx}_{t'} \\ \bar{\rvm}_{t'} 
    \end{pmatrix} = \begin{pmatrix}
        \bar{\rvx}_t \\ \bar{\rvm}_t
    \end{pmatrix} + \frac{\beta_t\delta t}{2}\begin{pmatrix}
        \Gamma \bar{\rvx}_t - M^{-1}\bar{\rvm}_t + 2\Gamma \vs_{\theta}(\bar{\rvz}_t, T-t)|_{0:d}\\
        \bar{\rvx}_t + \nu \bar{\rvm}_t + 2M\nu \vs_{\theta}(\bar{\rvz}_t, T-t)|_{d:2d})
    \end{pmatrix} + \begin{pmatrix}
        \sqrt{\Gamma\beta_t\delta t}\bm{\epsilon}_{t'}^x \\ \sqrt{M\nu\beta_t\delta t} \bm{\epsilon}_{t'}^m
    \end{pmatrix}
\end{align}
where $\bm{\epsilon}_t = [\bm{\epsilon}_t^x, \bm{\epsilon}_t^m]^T \sim \gN(\bm{0}_{2d}, \mI_{2d})$ and $t' = t + \delta t$ where $\delta t$ is the step size for a single update.

\subsubsection{Symmetric Splitting CLD Sampler (SSCS)}
\label{app:B_4_2}
Inspired by the application of splitting-based integrators in molecular dynamics \citep{alma991006381329705251}, \citet{dockhornscore} proposed the SSCS sampler with the following symmetric splitting scheme:
\begin{equation}
    \begin{pmatrix}
        d\bar{\rvx}_t \\
        d\bar{\rvm}_t
    \end{pmatrix} = \underbrace{\frac{\beta_t}{2}\begin{pmatrix}
        - \Gamma \bar{\rvx}_t -M^{-1}\bar{\rvm}_t  \\
        \bar{\rvx}_t - \nu\bar{\rvm}_t
    \end{pmatrix} dt + \mG(T-t)d\bar{\rvw}_t}_{A} + \underbrace{\beta_t\begin{pmatrix}
         \Gamma \bar{\rvx}_t + \Gamma\vs_{\theta}(\bar{\rvz}_t, T-t)|_{0:d} \\
        \nu\bar{\rvm}_t + M\nu\vs_{\theta}(\bar{\rvz}_t, T-t)|_{d:2d}
    \end{pmatrix}dt}_{S}
    \label{eqn:mod_sscs}
\end{equation}
\citet{dockhornscore} then approximate the flow map for the original SDE by the application of the following symmetric splitting schedule \cite{Trotter1959, Strang1968}:
\begin{equation}
    e^{t(\mathcal{L}_A + \mathcal{L}_S)} \approx \Big[e^{\frac{\delta t}{2}\mathcal{L}_A^*} e^{\delta t\mathcal{L}_S^*} e^{\frac{\delta t}{2}\mathcal{L}_A^*} \Big]^N + \mathcal{O}(N \delta t^3)
\end{equation}
where $N = \frac{t}{\delta t}$. The solution $\bar{\rvz}_t$ for the reverse SDE at any time t can then be obtained by the application of the flow map approximation of $e^{t(\mathcal{L}_A + \mathcal{L}_S)}$ to $\bar{\rvz}_0$. Since we use the same splitting formulation as \citet{dockhornscore}, the modified SSCS sampler for \mn is still a first-order integrator sampler (Also see Appendix D in \citet{dockhornscore} for more analysis of the SSCS sampler as proposed for CLD). However, since the form of the analytical splitting component in Eqn. \ref{eqn:mod_sscs} is different from CLD (due to a non-zero $\Gamma$), we next discuss the solution for this analytical form.

\textbf{Analytical splitting-term:} We have the following analytical splitting term:
\begin{align}
    \begin{pmatrix}
        d\bar{\rvx}_t \\
        d\bar{\rvm}_t
    \end{pmatrix} = \frac{\beta_t}{2}\begin{pmatrix}
        - \Gamma \bar{\rvx}_t -M^{-1}\bar{\rvm}_t  \\
        \bar{\rvx}_t - \nu\bar{\rvm}_t
    \end{pmatrix} dt + \left(\begin{pmatrix} \sqrt{\Gamma\beta_t} &0 \\ 0 &\sqrt{M\nu\beta_t} \end{pmatrix} \otimes \mI_d\right) d\bar{\rvw}_t
    \label{eqn:analytical}
\end{align}

The solution for this analytical SDE is similar to the derivation of the perturbation kernel in Appendix \ref{app:B_3}. However, there are two key differences. Firstly, we need to integrate between time-intervals $(t, t+\delta t)$ as opposed to from $(0, t)$ for the perturbation kernel. Secondly, since we are sampling, we set the initial covariances $\Sigma_{xx}^t$ and $\Sigma_{mm}^t$ to zero. The analytical solution for the SDE in Eqn. \ref{eqn:analytical} can then be specified as follows:
\begin{equation}
\bar{\rvz}_t \sim \gN(\bar{\vmu}_t, \bar{\mSigma}_t)
\end{equation}
where
\begin{equation}
    \vmu_(\bar{\rvx}_t, \bar{\rvm}_t, t, t') = \begin{pmatrix} A_1 \gB(t, t')\bar{\vx}_t + A_2\gB(t, t')\bar{\vm}_t + \bar{\vx}_t\\ C_1\gB(t, t')\bar{\vx}_t + C_2\gB(t, t')\bar{\vm}_t + \bar{\vm}_t\end{pmatrix}e^{-(\frac{\nu + \Gamma}{4}) \gB(t, t')}
\end{equation}
\begin{equation}
    A_1 = \frac{\nu - \Gamma}{4} \quad\quad A_2 = \frac{-(\Gamma - \nu)^2}{8}
\end{equation}
\begin{equation}
    C_1 = \frac{1}{2} \quad\quad C_2 = \frac{\Gamma - \nu}{4}
\end{equation}

The solution for the covariance is given by the following expression:
\begin{equation}
    \mSigma(t, t') = \left(\begin{pmatrix} \Sigma^{xx}_{t, t'} & \Sigma^{xm}_{t, t'} \\ \Sigma^{xm}_{t, t'} & \Sigma^{mm}_{t, t'} \end{pmatrix}e^{-(\frac{\Gamma + \nu}{2}) \gB(t, t')}\right) \otimes \mI_d
\end{equation}
where,
\begin{align}
    \Sigma^{xx}_t &= -\frac{(\Gamma - \nu)^2}{8} \gB^2(t, t') + \frac{\left(\Gamma - \nu\right)}{2} \gB(t, t') + (e^{-(\frac{\Gamma + \nu}{2}) \gB(t, t')} - 1)\\
    \Sigma^{xm}_t &= \frac{(\Gamma - \nu)}{4} \gB^2(t, t')\\
    \Sigma^{mm}_t &= -\frac{1}{2} \gB^2(t, t') + \frac{M(\Gamma - \nu)}{2} \gB(t, t') + M(e^{-(\frac{\Gamma + \nu}{2}) \gB(t, t')} - 1)
\end{align}
where $\gB(t, t') = -\int_t^{t'} \beta(s) ds$ and $t' = t + \delta t$. Indeed, setting $\Gamma = 0$ recovers the original SSCS algorithm proposed in \citet{dockhornscore}. Therefore, given $\bar{\rvz}_t = \Big(\bar{\rvx}_t, \bar{\rvm}_t\Big)^T$, the flow map update for the analytical splitting term $e^{\frac{\delta t}{2}\mathcal{L}_A^*}$ is given by:
\begin{equation}
    \bar{\rvz}_{t'} \sim \gN(\vmu(\bar{\rvx}_t, \bar{\rvm}_t, t, t'), \mSigma(t, t'))
\end{equation}
where $t' = t + \frac{\delta t}{2}$.

\textbf{Score-based splitting term:} The flow map update for the score-based splitting term $e^{\delta t\mathcal{L}_S^*}$ is given by an Euler update as follows:
\begin{equation}
    \begin{pmatrix}
        \bar{\rvx}_{t'} \\ \bar{\rvm}_{t'} 
    \end{pmatrix} = \begin{pmatrix}
        \bar{\rvx}_t \\ \bar{\rvm}_t 
    \end{pmatrix} + \delta t\beta_t\begin{pmatrix}
         \Gamma \bar{\rvx}_t + \Gamma\vs_{\theta}(\bar{\rvz}_t, T-t)|_{0:d} \\
        \nu\bar{\rvm}_t + M\nu\vs_{\theta}(\bar{\rvz}_t, T-t)|_{d:2d}
    \end{pmatrix}
\end{equation}

Combining the two splitting terms together, a more generic form of the SSCS algorithm can be specified as follows:

\begin{algorithm}[H]
\small
\caption{{\textit{Modified SSCS} (Terms in \textcolor{blue}{blue} indicate differences from the SSCS sampler proposed in \citet{dockhornscore})} %
}
\begin{algorithmic}

\State {\bfseries Input:} Trajectory length T, Score function $\vs_\vtheta(\rvz_t,T-t)$, \textcolor{blue}{\mn parameters $\Gamma$, $\nu$, $\beta_t$, $M=\frac{(\Gamma - \nu)^2}{4}$}, number of sampling steps $N$, step sizes $\{\delta t_n\geq0\}_{n=0}^{N-1}$ spanning the interval (0, $T - \eps$).
\State {\bfseries Output:} $\bar{\rvz}_T$ = ($\bar{\rvx}_T$, $\bar{\rvm}_T$) %
\State
\State $\bar{\rvx}_0\sim \mathcal{N}(\bm{0}_d,\mI_d)$, $\bar{\rvm}_0\sim\mathcal{N}(\bm{0}_d,M\mI_d)$, $\bar{\rvz}_0=(\bar{\rvx}_0,\bar{\rvm}_0)$
\Comment{Draw initial prior samples from $p_\textrm{EQ}(\rvu)$}
\State $t=0$ \Comment{Initialize time}
\For{$n=0$ {\bfseries to} $N-1$}
\State $\textcolor{blue}{\bar{\rvz}_{n+\frac{1}{2}} \sim \gN(\vmu(\bar{\rvx}_n, \bar{\rvm}_n, t, t+ \frac{\delta t_n}{2}), \mSigma(t, t + \frac{\delta t_n}{2}))}$ \Comment{First half-step: Apply $\exp\{\frac{\delta t_n}{2}\lop^*_{A}\}$}

\State $\bar{\rvz}_{n+\frac{1}{2}} \leftarrow \bar{\rvz}_{n+\frac{1}{2}} + \delta t_n
\beta_t\textcolor{blue}{\begin{pmatrix}
         \Gamma \bar{\rvx}_{n + \frac{1}{2}} + \Gamma\vs_{\theta}(\bar{\rvz}_{n + \frac{1}{2}}, T-t)|_{0:d} \\
        \nu\bar{\rvm}_{n + \frac{1}{2}} + M\nu\vs_{\theta}(\bar{\rvz}_{n + \frac{1}{2}}, T-t)|_{d:2d}
    \end{pmatrix}}$ \Comment{Full step: Apply $\exp\{\delta t_n\lop^*_{S}\}$}
\State $\textcolor{blue}{\bar{\rvz}_{n+1} \sim \gN(\vmu(\bar{\rvx}_{n+\frac{1}{2}}, \bar{\rvm}_{n+\frac{1}{2}}, t, t+ \frac{\delta t_n}{2}), \mSigma(t, t + \frac{\delta t_n}{2}))}$ \Comment{Second half-step: Apply $\exp\{\frac{\delta t_n}{2}\lop^*_{A}\}$}
\State $t \leftarrow t + \delta t_n$ \Comment{Update time}
\EndFor
\State $\bar{\rvz}_N \leftarrow \bar{\rvz}_N + \epsilon\textcolor{blue}{\frac{\beta_t}{2}\begin{pmatrix}
        \Gamma \bar{\rvx}_{n+1} - M^{-1}\bar{\rvm}_{n+1} + 2\Gamma \vs_{\theta}(\bar{\rvz}_{n+1}, \eps)|_{0:d}\\
        \bar{\rvx}_{n+1} + \nu \bar{\rvm}_{n+1} + 2M\nu \vs_{\theta}(\bar{\rvz}_{n+1}, \eps)|_{d:2d})
    \end{pmatrix}}$ \Comment{Denoising}
\State $(\bar{\rvx}_N, \bar{\rvm}_N)=\bar{\rvz}_N$  \Comment{Extract output data and momentum samples}
\end{algorithmic}
\end{algorithm}

\subsubsection{Probability Flow ODE}
\label{app:B_4_3}
Following \citet{songscore}, the probability flow ODE for \mn can be specified as follows:
\begin{equation}
    d\bar{\rvz}_t = \bar{\vf}(\bar{\rvz}_t)dt
\end{equation}
\begin{equation}
    \bar{\vf}(\bar{\rvz}_t) = \frac{\beta_t}{2}\begin{pmatrix}
        \Gamma \bar{\rvx}_t - M^{-1}\bar{\rvm}_t + \Gamma \vs_{\theta}(\bar{\rvz}_t, T-t)|_{0:d}\\
        \bar{\rvx}_t + \nu \bar{\rvm}_t + M\nu \vs_{\theta}(\bar{\rvz}_t, T-t)|_{d:2d})
    \end{pmatrix}
\end{equation}
The Probability-Flow ODE can be solved using any fixed/adaptive step-size black-box ODE solvers like RK45 \citep{DORMAND198019}

%% file: tex/appendix/App_C.tex
\section{Implementation Details}
\label{app:C}
\subsection{Datasets and Preprocessing}
We use CIFAR-10 \citep{krizhevsky2009learning} (50k images) and CelebA-64 ($\approx$ 200k images) \citep{liu2015faceattributes} datasets for both quantitative and qualitative analysis. We use the AFHQv2 \citep{choi2020stargan} dataset ($\approx$ 14k images) only for qualitative analysis. Unless specified otherwise, we always use the CelebA dataset at 64x64 resolution and the AFHQv2 dataset at 128 x 128 resolution. During training, all datasets are preprocessed to a numerical range of [-1, 1]. Following prior work, we use random horizontal flips to train all models (ablation and SOTA) across datasets as a data augmentation strategy.

\subsection{Score Network Architecture}
\label{app:C_2}
Table \ref{table:app_1} illustrates our score model architectures for different datasets. Our network architectures are largely based on the design of the DDPM++/NCSN++ score networks introduced in \citet{songscore}. Apart from minor design choices, the DDPM++/NCSN++ score-network architectures are primarily based on the U-Net \citep{ronneberger2015u} model. We further highlight several key aspects of our score network architectures across different datasets as follows:

\textbf{CIFAR-10}: We use a smaller version (39M) of the DDPM++ architecture (with the number of residual blocks per resolution set to two) for ablation studies (for both VP-SDE and \nsmn) while we use the NCSN++ architecture \citep{songscore} for training larger models used for SOTA comparisons. Moreover, when training larger models, like \citet{karraselucidating} we remove the layers at 4x4 resolution and re-distribute capacity to the layers at the 16x16 resolution. This results in model sizes of 55M/97M parameters corresponding to four and eight residual blocks per resolution, respectively, with channel multipliers [2,2,2]. Moreover, when training larger models (55M/97M), we slightly increase the dropout rate from 0.1 to 0.15. We observed that these changes improved performance slightly while reducing model sizes and enabling faster training.

\textbf{CelebA-64}: Similar to CIFAR-10, we use a DDPM++ score model architecture for ablation experiments. 
while we use a NCSN++ architecture for SOTA comparisons. Moreover, we remove the 4x4 layers from our ablation model for SOTA analysis and increase the channel multiplier for the 32x32 layers from 1 to 2. 
The dropout rate is set to 0.1 due to a larger dataset size for CelebA-64. This setting results in a model size of approximately 66M for the ablation experiments and 62M for SOTA comparisons.

\textbf{AFHQv2}: We use the original DDPM++ architecture for training our AFHQv2 model. Additionally, we increase the dropout rate to 0.2, given a relatively smaller dataset size. This setting results in a model size of approximately 68M parameters for qualitative analysis.

\begin{table}[]
\centering
\begin{tabular}{lccccc}

                                                                    \toprule     & \multicolumn{2}{c}{CIFAR-10}  & \multicolumn{2}{c}{CelebA-64}   & AFHQv2          \\\midrule  
Hyperparameter                                                           & SOTA          & Ablation      & SOTA          & Ablation        & Qualitative     \\ \midrule 
Base channels                                                            & 128           & 128           & 128           & 128             & 128             \\
Channel multiplier                                                       & {[}2,2,2{]}   & {[}1,2,2,2{]} & {[}1,2,2,2{]} & {[}1,1,2,2,2{]} & {[}1,2,2,2,3{]} \\
\# Residual blocks                                                       & 4,8           & 2             & 4             & 4               & 2               \\
Non-Linearity                                                            & Swish         & Swish         & Swish         & Swish           & Swish           \\
Attention resolution                                                     & {[}16{]}      & {[}16{]}      & {[}16{]}      & {[}16{]}        & {[}16{]}        \\
\# Attention heads                                                       & 1             & 1             & 1             & 1               & 1               \\
Dropout                                                                  & 0.15          & 0.1           & 0.1           & 0.1             & 0.2             \\
Finite Impulse Response (FIR) \citep{pmlr-v97-zhang19a} & True          & False         & True          & False           & False           \\
FIR kernel                                                               & {[}1,3,3,1{]} & N/A           & {[}1,3,31{]}  & N/A             & N/A             \\
Progressive Input                                                        & Residual      & None          & Residual      & None            & None            \\
Progressive Combine                                                      & Sum           & Sum           & Sum           & Sum             & Sum             \\
Embedding type                                                           & Fourier       & Positional    & Fourier       & Positional      & Positional      \\
Sigma scaling                                                            & False         & False         & False         & False           & False           \\
Model size                                                               & 55M/97M       & 39M           & 62M           & 66M             & 68M             \\ \bottomrule 
\end{tabular}
\caption{Score Network hyperparameters for \nsmn.}
\label{table:app_1}
\end{table}

\subsection{SDE Parameters}
\label{app:C_3}
\textbf{\nsmn:} For \mn (including the CLD baseline), unless specified otherwise, we set the mass parameter $M^{-1}=4$ and $\beta=8$. The choice of these parameters is motivated by empirical results presented in \citet{dockhornscore}. We add a stabilizing numerical epsilon value of $1e^{-9}$ in the diagonal entries of the Cholesky decomposition of $\mSigma_t$ when sampling the perturbed data-point $\rvz_t \sim \gN(\vmu_t, \mSigma_t)$ during training. The data-generating distribution is set to $p_0(\rvz) = \gN(\bm{0}_d, \mI_d)\gN(\bm{0}_d, M\gamma\mI_d)$ where $\gamma=0.04$. For SOTA analysis, we experiment with $\Gamma \in \{0.01, 0.02\}$ for CIFAR-10 and $\Gamma = 0.005$ for the CelebA-64 datasets. We chose these values of $\Gamma$ and $\nu$ based on the best-performing (in terms of FID) ablation models for these datasets (See Table 4 in the main text). Lastly, for training our AFHQv2 model for qualitative analysis, we set $\Gamma=0.01$. All the other SDE parameters remain the same.

\textbf{VP-SDE:} For our VP-SDE baseline, following \citet{songscore}, we set $\beta_{\text{min}} = 0.1$ and $\beta_{\text{max}} = 20.0$

\subsection{Training}
\label{app:C_4}
Table \ref{table:app_2} summarizes the different training hyperparameters across datasets and evaluation settings (ablation and SOTA). Additionally, we use the Hybrid Score Matching (HSM) objective (See Appendix \ref{app:B_2_1}) for all augmented state-space models (\mn and CLD); for the VP-SDE baseline, we use the Denoising Score Matching (DSM) objective. Throughout this work, we optimize for sample quality and thus use the \textit{epsilon-prediction} loss during training (See Appendix \ref{app:B_2_1}).

\begin{table}[]
\centering
\begin{tabular}{@{}lccccc@{}}
\toprule
                     & \multicolumn{2}{c}{CIFAR-10} & \multicolumn{2}{c}{CelebA-64} & AFHQv2                          \\ \midrule 
                     & SOTA         & Ablation      & SOTA         & Ablation       & \multicolumn{1}{l}{Qualitative} \\ \midrule 
Random Seed          & 0            & 0             & 0            & 0              & 0                               \\
\# iterations        & 800k         & 800k          & 800k         & 320k           & 400k                            \\
Optimizer            & Adam         & Adam          & Adam         & Adam           & Adam                            \\
Grad Clip. cutoff    & 1.0          & 1.0           & 1.0          & 1.0            & 1.0                             \\
Learning rate (LR)   & 2e-4         & 2e-4          & 2e-4         & 2e-4           & 1e-4                            \\
LR Warmup steps      & 5000         & 5000          & 5000         & 5000           & 5000                            \\
FP16                 & False        & False         & False        & False          & False                           \\
EMA Rate             & 0.9999       & 0.9999        & 0.9999       & 0.9999         & 0.9999                          \\
Effective Batch size & 128          & 128           & 128          & 128            & 64                              \\
\# GPUs              & 8            & 4             & 8            & 4              & 8                               \\
Train eps cutoff     & 1e-5         & 1e-5          & 1e-5         & 1e-5           & 1e-5                            \\ \bottomrule 
\end{tabular}
\caption{Training hyperparameters for \nsmn}
\label{table:app_2}
\end{table}

\subsection{Evaluation}
\label{app:C_5}

\textbf{SDE Sampling:} As is common in prior works \citep{songscore, dockhornscore}, we use the integration interval $(1e^{-3}, 1.0)$ for solving the reverse SDE/ODE for sample generation. Unless specified otherwise, we use 1000 sampling steps when using numerical SDE solvers. When using numerical black-box ODE solvers, we use the RK-45 \citep{DORMAND198019} solver
with the same absolute and relative tolerance levels. We use the ODE solver at different tolerance levels for ablations studies (Table 6 in the main text) and tolerance levels of $1e^{-5}$ and $1e^{-4}$ for reporting SOTA results on CIFAR-10. Similarly, we use a tolerance level of $1e^{-5}$ for reporting ODE solver performance on CelebA-64. We use the \texttt{odeint} function from the \texttt{torchdiffeq}\citep{torchdiffeq} package when using the black-box ODE solver for sampling.

\begin{table}[]
\footnotesize
\centering
\begin{minipage}{0.52\linewidth}
\begin{tabular}{@{}cccc@{}}
\toprule
                          &                      & CIFAR-10 & AFHQ-v2 \\ \midrule
\multirow{9}{*}{Training} & Random Seed          & 0        & 0       \\
                          & \# iterations        & 200k     & 70k     \\
                          & Optimizer            & Adam     & Adam    \\
                          & LR   & 2e-4     & 2e-4    \\
                          & Warmup steps      & 5000     & 5000    \\
                          & FP16                 & False    & False   \\
                          & Batch size & 256      & 64      \\
                          & \# GPUs              & 4        & 4       \\
                          & Train eps cutoff     & 1e-5     & 1e-5    \\ \midrule
\multirow{4}{*}{SDE}      & $M^{-1}$             & 4.0      & 4.0     \\
                          & $\Gamma$             & 0.01     & 0.01    \\
                          & $\nu$                & 4.01     & 4.01    \\
                          & $\beta$              & 8.0      & 8.0     \\ \midrule 
\end{tabular}
\caption{Classifier Training Hyperparameters}
\label{table:app_clf_tr}
\end{minipage}
\hfill
\begin{minipage}{0.47\linewidth}
\begin{tabular}{lcc}
\toprule
                                 & CIFAR-10      & AFHQ-v2       \\ \midrule
Base channels                                  & 128           & 128           \\
Num. classes                                   & 10            & 3             \\
Channel multiplier                             & {[}1,2,3,4{]} & {[}1,2,3,4{]} \\
\# Residual blocks                             & 4             & 4             \\
Non-Linearity                                  & Swish         & Swish         \\
Attention resolution                           & {[}16, 8{]}   & {[}16, 8{]}   \\
\# Attention heads                             & 1             & 1             \\
Dropout                                        & 0.1           & 0.1           \\
FIR \citep{pmlr-v97-zhang19a} & False         & False         \\
Progressive Input                              & None          & None          \\
Progressive Combine                            & Sum           & Sum           \\
Embedding type                                 & Positional    & Positional    \\
Sigma scaling                                  & False         & False         \\
Model size                                     & 56.7M         & 57.8M         \\ \bottomrule
\end{tabular}
\caption{Classifier Network Hyperparameters}
\label{table:app_clf_na}
\end{minipage}
\end{table}

\textbf{Timestep Selection during Sampling}: We use \textit{Uniform} (US) and \textit{Quadratic} (QS) striding for timestep discretization in this work. In uniform striding, given an NFE budget N, we discretize the integration interval ($\eps$, T) into N equidistant parts, which are then used for score function evaluations. In quadratic striding \citep{dockhornscore, songdenoising}, the evaluation timepoints are given by:
\begin{equation}
    \tau_i = \left(\frac{i}{N}\right)^2 \;\forall i \in [0, N)
\end{equation}
This ensures more number of score function evaluations in the lower timestep regime (i.e. $t$, which is close to the data). This kind of timestep selection is particularly useful when the NFE budget is limited (See Table 5).

\textbf{Last-Step Denoising}: Similar to prior works \citep{songscore, dockhornscore, jolicoeur-martineau2021adversarial}, we perform a single denoising EM step (without noise injection) at the very last step of our sampling routine for both SDE and ODE solvers. Formally, we perform the following update:
\begin{align}
    \begin{pmatrix}
        \rvx_0 \\ \rvm_0 
    \end{pmatrix} = \begin{pmatrix}
        \rvx_{\eps} \\ \rvm_{\eps}
    \end{pmatrix} + \frac{\beta_t\eps}{2}\begin{pmatrix}
        \Gamma \rvx_{\eps} - M^{-1}\rvm_{\eps} + 2\Gamma \vs_{\theta}(\rvz_{\eps}, \eps)|_{0:d}\\
        \rvx_{\eps} + \nu \rvm_{\eps} + 2M\nu \vs_{\theta}(\rvz_{\eps}, \eps)|_{d:2d})
    \end{pmatrix}
\end{align}
where $\eps = 1e^{-3}$. Such a denoising step has been found useful in removing additional noise, thereby improving FID scores \citep{jolicoeur-martineau2021adversarial}.

\textbf{Evaluation Metrics:} For most ablation experiments involving the analysis of speed vs sample quality trade-offs between different models, we report the FID \citep{heusel2017gans} score on 10k samples for computational convenience. For SOTA comparisons, we report FID for 50k samples for both CIFAR-10 and CelebA-64 datasets. When reporting extended SOTA results for CIFAR-10, we also report the Inception Score (IS) \citep{salimans2016improved} metric. We use the \texttt{torch-fidelity}\citep{obukhov2020torchfidelity} package for computing all FID and IS scores reported in this work. When reporting average NFE (number of function evaluations) in Table 6, we average the NFE values over a batch size of 16 samples for 10k samples in aggregate and take a ceiling of the resulting value.

\subsection{Classifier Architecture and Training}
\label{app:C_6}
For class conditional synthesis (Appendix \ref{app:D_4}), we append the downsampling part of the UNet architecture with a classification head and use the resulting model as our classifier architecture. Table \ref{table:app_clf_na} shows different hyperparameters of our classifier model architecture. For classifier training, we set $\Gamma=0.01$ for the AFHQv2 and the CIFAR-10 datasets. The remaining SDE parameters remain unchanged from our previous setting. Table \ref{table:app_clf_tr} lists different hyperparameters for classifier training.

%% file: tex/appendix/App_D.tex
\section{Additional Results}
\label{app:D}

\subsection{Impact of $\Gamma$ and $\nu$ on \mn Sample Quality}
\label{app:D_1}

Table \ref{table:app_3} shows the impact of varying $\Gamma$ and $\nu$ (with a fixed $M^{-1}$) on the \mn sample quality using the EM-sampler with quadratic striding (EM-QS) for the CIFAR-10 dataset. Extending Table 4, we additionally present FID scores for $\Gamma \in \{4.0,8.0\}$ in Table \ref{table:app_3}. As we increase the value of $\Gamma$ to 4.0 or 8.0, the FID scores further increase to 11.43 and 14.15, respectively, confirming our observations in Section 4.2. Figure \ref{fig:app_3} further illustrates the qualitative impact of increasing $\Gamma$ on CIFAR-10 sample quality. For the setting with $\Gamma=8.0$, using EM with uniform striding (EM-US) introduces evident noise artifacts in the generated samples. However, such artifacts are less pronounced when using EM with quadratic striding ((EM-QS)) instead. This suggests potential denoising problems for low timestep indices during sampling as quadratic striding focuses more score network evaluations in the low timestep regime, which might lead to lesser artifacts. This is similar to our observations in Section 4.2 (See Figure \ref{fig:app_4} for more qualitative results on CelebA-64). We now provide a formal justification for these observations.
\begin{table}[]
\centering
\begin{tabular}{@{}cccc@{}}
\toprule
$\Gamma$ & $\nu$ & $M^{-1} = \frac{(\Gamma - \nu)^2}{4}$ & \begin{tabular}[c]{@{}l@{}}FID@50k $\downarrow$ \\ (EM-QS)\end{tabular} \\ \midrule
0        & 4     & 4   & 3.64                                                                    \\
0.005    & 4.005 & 4   & 3.42                                                                    \\
0.01     & 4.01  & 4   & 3.15                                                                   \\
0.02     & 4.02  & 4   & 3.26                                                                   \\
0.25     & 4.25  & 4   & 4.99                                                                   \\
4        & 0     & 4   & 11.43                                                                  \\
8        & 4     & 4   & 14.15                                                                  \\ \bottomrule
\end{tabular}
\caption{Extended results for impact of the choice of $\Gamma$ on sample quality for CIFAR-10. FID (lower is better) reported on 50k samples.}
\label{table:app_3}
\end{table}

Given an input sample $\bar{\rvz}_t = (\bar{\rvx}_t, \bar{\rvm}_t)$ at time t, consider the following update rule for the EM-sampler for \mn with a uniform spacing interval of $\delta t$ between successive steps:
\begin{align}
    \begin{pmatrix}
        \bar{\rvx}_{t'} \\ \bar{\rvm}_{t'} 
    \end{pmatrix} = \begin{pmatrix}
        \bar{\rvx}_t \\ \bar{\rvm}_t
    \end{pmatrix} + \frac{\beta\delta t}{2}\begin{pmatrix}
        \Gamma \bar{\rvx}_t - M^{-1}\bar{\rvm}_t + 2\Gamma \vs_{\theta}(\bar{\rvz}_t, T-t)|_{0:d}\\
        \bar{\rvx}_t + \nu \bar{\rvm}_t + 2M\nu \vs_{\theta}(\bar{\rvz}_t, T-t)|_{d:2d})
    \end{pmatrix} + \begin{pmatrix}
        \sqrt{\Gamma\beta\delta t}\bm{\epsilon}_{t'}^x \\ \sqrt{M\nu\beta\delta t} \bm{\epsilon}_{t'}^m
    \end{pmatrix}
\end{align}
To simplify notation, let us denote $\bar{\beta} = \frac{\beta\delta t}{2}$, $\vs_{\theta}^x(\bar{\rvz}_t) = \vs_{\theta}(\bar{\rvz}_t, T-t)|_{0:d}$ and $\vs_{\theta}^m(\bar{\rvz}_t) = \vs_{\theta}(\bar{\rvz}_t, T-t)|_{d:2d}$. Therefore, for the next timestep $t'$, we have:
\begin{align}
    \brvx_{t'} &= \brvx_t + \bar{\beta}\Big(\Gamma \brvx_t - M^{-1}\brvm_t + 2\Gamma\vs_{\theta}^x(\bar{\rvz}_t)\Big) + \sqrt{\Gamma\beta\delta t}\bm{\epsilon}_{t'}^x \label{eqn:em_1}\\
    \brvm_{t'} &= \brvm_t + \bar{\beta} \Big(\brvx_t + \nu \brvm_t + 2M\nu\vs_{\theta}^m(\bar{\rvz}_t) \Big) + \sqrt{M\nu\beta\delta t}\bm{\epsilon}_{t'}^m
    \label{eqn:em_2}
\end{align}
Similarly for the next consecutive time-step $t^{''}$, we have the following EM-update rule:
\begin{equation}
    \brvx_{t''} = \brvx_{t'} + \bar{\beta}\Big(\Gamma \brvx_{t'} - M^{-1}\brvm_{t'} + 2\Gamma\vs_{\theta}^x(\bar{\rvz}_{t'})\Big) + \sqrt{\Gamma\beta\delta t}\bm{\epsilon}_{t''}^x
    \label{eqn:update_t}
\end{equation}
Substituting the update expressions for $\brvx_{t'}$ and $\brvm_{t'}$ from Eqns. \ref{eqn:em_1}-\ref{eqn:em_2} in the update rule for $\brvx_{t''}$, we have the following modified update rule for $\brvx_{t''}$:
\begin{equation}
    \brvx_{t''} = \vf(\brvx_t, \brvm_t) + \hat{\vs}_{\theta} + \bm{\eta}
\end{equation}
where $\vf$ is a function of $(\brvx_t, \brvm_t)$, $\bm{\eta}$ is the aggregate stochastic noise. More importantly, the score term $\hat{\vs}_{\theta}$ is given as follows:
\begin{align}
    \hat{\vs}_{\theta} &= 2\bar{\beta}\Gamma\vs_{\theta}^x(\brvz_t) + 2 \bar{\beta}^2 \Big[\Gamma^2 \vs_{\theta}^x(\brvz_t) - \nu\vs_{\theta}^m(\brvz_t)\Big] + 2 \Gamma \bar{\beta}\vs_{\theta}^x(\brvz_{t'}) \\
    &= 2\bar{\beta}\Gamma\vs_{\theta}^x(\brvz_t) + 2 \bar{\beta}^2 \Big[\begin{pmatrix}\vs_{\theta}^x(\brvz_t) \\ \vs_{\theta}^m(\brvz_t)\end{pmatrix}^T\begin{pmatrix}
        \Gamma^2 \\ - \nu
    \end{pmatrix}\Big] + 2 \Gamma \bar{\beta}\vs_{\theta}^x(\brvz_{t'}) \\
    &= 2\bar{\beta}\Gamma\vs_{\theta}^x(\brvz_t) + 2 \bar{\beta}^2 \Big[\vs_{\theta}(\brvz_t)^T\begin{pmatrix}
        \Gamma^2 \\ - \nu
    \end{pmatrix}\Big] + 2 \Gamma \bar{\beta}\vs_{\theta}^x(\brvz_{t'}) \label{eqn:em_3}
\end{align}
In this work, we parameterize the score $\vs_{\theta}(\brvz_t) = -\mL_t^{-T}\bm{\eps}_{\theta}(\brvz_t)$ where $\mL_t$ is Cholesky factorization matrix of the covariance matrix $\mSigma_t$ of the perturbation kernel at time t. Substituting this parameterization in Eqn. \ref{eqn:em_3}, we get,
\begin{align}
    \hat{\vs}_{\theta} &= 2\bar{\beta}\Gamma\vs_{\theta}^x(\brvz_t) + 2 \bar{\beta}^2 \Big[-\bm{\eps}^T_{\theta}(\brvz_t)\mL_t^{-1}\begin{pmatrix}
        \Gamma^2 \\ - \nu
    \end{pmatrix}\Big] + 2 \Gamma \bar{\beta}\vs_{\theta}^x(\brvz_{t'}) \\
    &= 2\bar{\beta}\Gamma\vs_{\theta}^x(\brvz_t) - 2 \bar{\beta}^2 \Big[\Gamma^2 (l_t^{xx} \bm{\eps}_{\theta}^x(\brvz_t) + l_t^{xm}\bm{\eps}_{\theta}^m(\brvz_t)) - \nu l_t^{mm}\bm{\eps}^m_{\theta}(\brvz_t)\Big] + 2 \Gamma \bar{\beta}\vs_{\theta}^x(\brvz_{t'})\\
    &= 2\bar{\beta}\Gamma\vs_{\theta}^x(\brvz_t) - 2 \bar{\beta}^2 \Big[\underbrace{\Gamma^2 l_t^{xx}}_{=\lambda_1} \bm{\eps}_{\theta}^x(\brvz_t) + \underbrace{(\Gamma^2l_t^{xm} - \nu l_t^{mm})}_{=\lambda_2}\bm{\eps}_{\theta}^m(\brvz_t) \Big] + 2 \Gamma \bar{\beta}\vs_{\theta}^x(\brvz_{t'})
    \label{eqn:em_4}
\end{align}
where,
\begin{equation}
    l_t^{xx} = \frac{1}{\sqrt{\Sigma^{xx}_t }}
\end{equation}
\begin{equation}
    l_t^{xm} = \frac{-\Sigma^{xm}_t}{\sqrt{\Sigma^{xx}_t}\sqrt{\Sigma^{xx}_t \Sigma^{mm}_t - \left(\Sigma^{xm}_t\right)^2}}
\end{equation}
\begin{equation}
    l_t^{mm} = \sqrt{\frac{\Sigma^{xx}_t}{\Sigma^{xx}_t \Sigma^{mm}_t - \left(\Sigma^{xm}_t\right)^2}}
\end{equation}
Assuming the input $\brvz_t$ is a sample from the underlying flow map of the reverse SDE (a very strong assumption), the score term $\hat{\vs}_{\theta}$ in Eqn. \ref{eqn:update_t} is the primary source of introducing errors (since the neural network-based score prediction will be offset by some error from the true underlying score). Without loss of generality, we further assume that the update timepoints $t, t' \text{and}\, t''$ lie in the low timestep regime. Furthermore, for notational convenience, we denote $\lambda_1 = \Gamma^2 l_t^{xx}$ and $\lambda_2 = (\Gamma^2l_t^{xm} - \nu l_t^{mm}$) as the scaling factors for the second term in Eqn. \ref{eqn:em_4}. Thus, the error introduced due to the neural network predictors $\bm{\eps}^x_{\theta}(\brvz_t)$ and $\bm{\eps}^m_{\theta}(\brvz_t)$ will be scaled by $\lambda_1$ and $\lambda_2$ respectively. Therefore, for achieving lower sampler discretization errors, it might be desirable to have low magnitudes of $\lambda_1$ and $\lambda_2$. We now qualitatively analyze the magnitude of these coefficients for different ranges of values of $\Gamma$ and $\nu$.

\begin{figure}
\centering
    \includegraphics[width=1.0\linewidth]{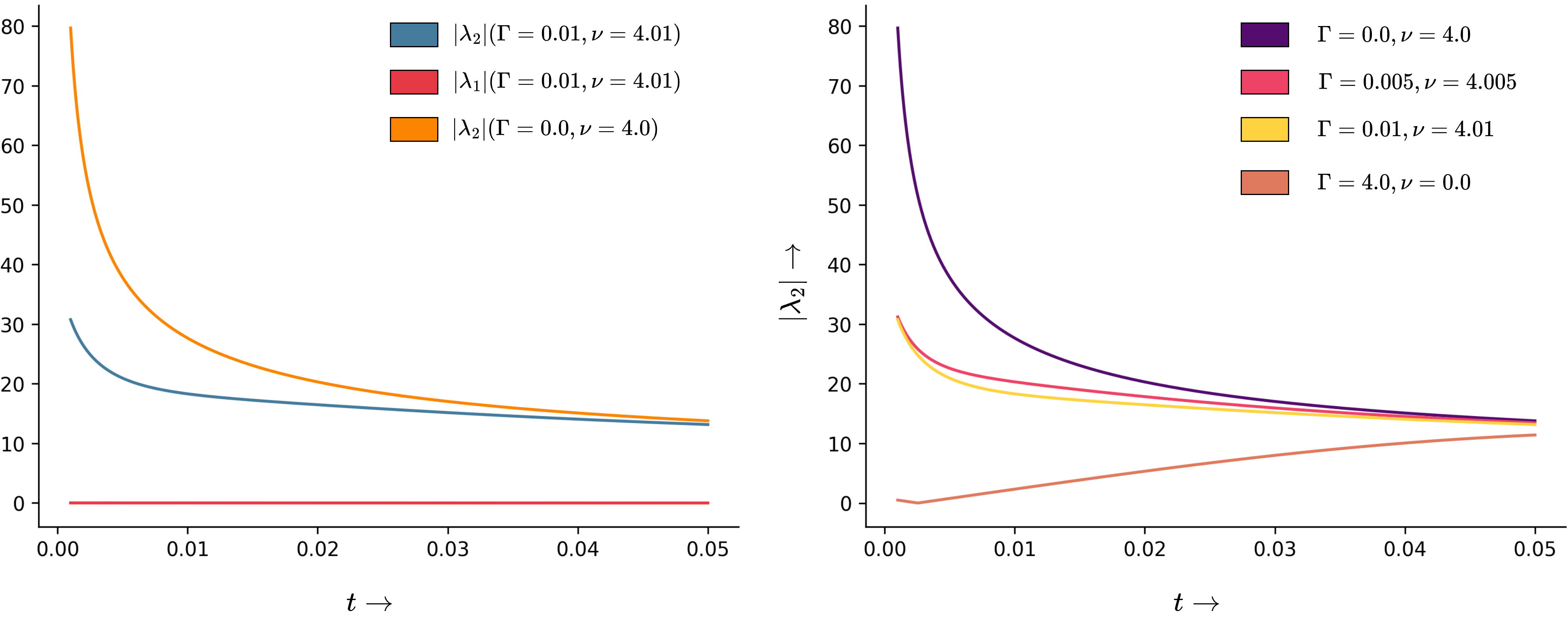}
    \caption{(a) Comparison between $|\lambda_1|$ and $|\lambda_2|$ corresponding to $\Gamma=0.0$ and $\Gamma=0.01$ in the low-timestep regime for a fixed $M^{-1}=4$. (b) Variation of $|\lambda_2|$ for different values of $(\Gamma, \nu)$}
    \label{fig:app_1}
\end{figure}

\textbf{Case-1: Effect of using a non-zero $\Gamma$}: We first analyze the impact of using a non-zero $\Gamma$ value on the magnitude of $\lambda_1$ and $\lambda_2$. Figure \ref{fig:app_1}a illustrates the impact of the choice of $\Gamma$ and $\nu$ on the coefficients $\lambda_1$ and $\lambda_2$ in the low-timestep regime. When $\Gamma=0$, the error in the score term $\hat{\vs}_{\theta}$ will be only due to the term $|\lambda_2|\bm{\eps}_{\theta}^m(\brvz_t)$. As illustrated in Figure \ref{fig:app_1}a, the value of $|\lambda_2|$ (when $\Gamma=0$) is very high in the low-timestep regime and, therefore might negatively impact the sample quality since any errors in the estimation of $\bm{\eps}_{\theta}^m(\brvz_t)$ would be scaled by a large factor.

Interestingly, for the setting $\Gamma=0.01, \nu=4.01$, the value of $|\lambda_2|$ reduces significantly, thus reducing the error scaling factor. It is worth noting that using a non-zero $\Gamma$ also simultaneously enables error contribution from other terms in $\hat{\vs}_{\theta}$ involving $\Gamma$ (especially $|\lambda_1| \bm{\eps}_{\theta}^x(\brvz_t)$). However, as illustrated in Figure \ref{fig:app_1}a, the value of $|\lambda_1|$ is extremely small as compared to $|\lambda_2|$ making the additional error introduced insignificant. Due to this reason, the overall error introduced by the score term $\hat{\vs}_{\theta}$ is more when $\Gamma=0$ as compared to the setting with a (small) non-zero $\Gamma$ value. This explains why using a small value of $\Gamma$ yields better sample quality than our CLD baseline (See Table 4)

\begin{figure}
\centering
    \includegraphics[width=1.0\linewidth]{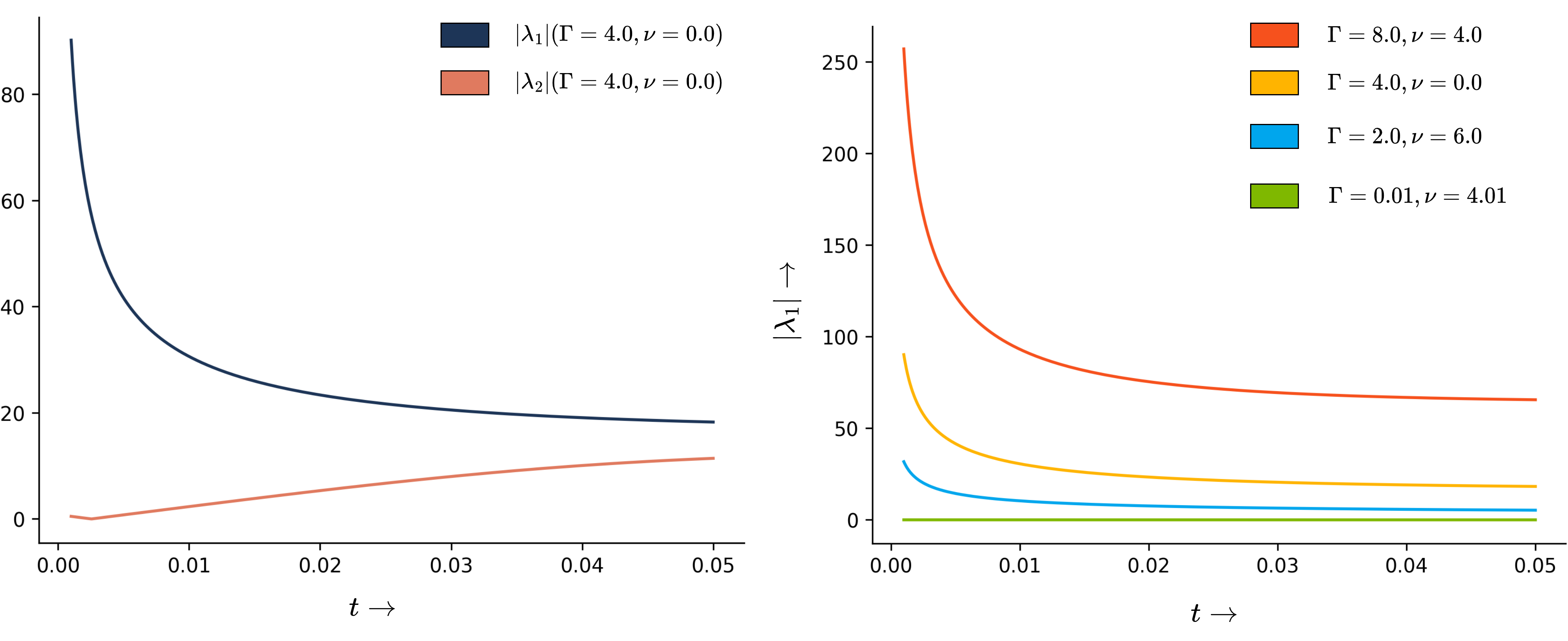}
    \caption{(a) Comparison between $|\lambda_1|$ and $|\lambda_2|$ corresponding to $\Gamma=4.0$ and $\nu=0.0$ in the low-timestep regime. (b) Variation of $|\lambda_1|$ for different values of $(\Gamma, \nu)$}
    \label{fig:app_2}
\end{figure}

\begin{figure}
\centering
    \includegraphics[width=1.0\linewidth]{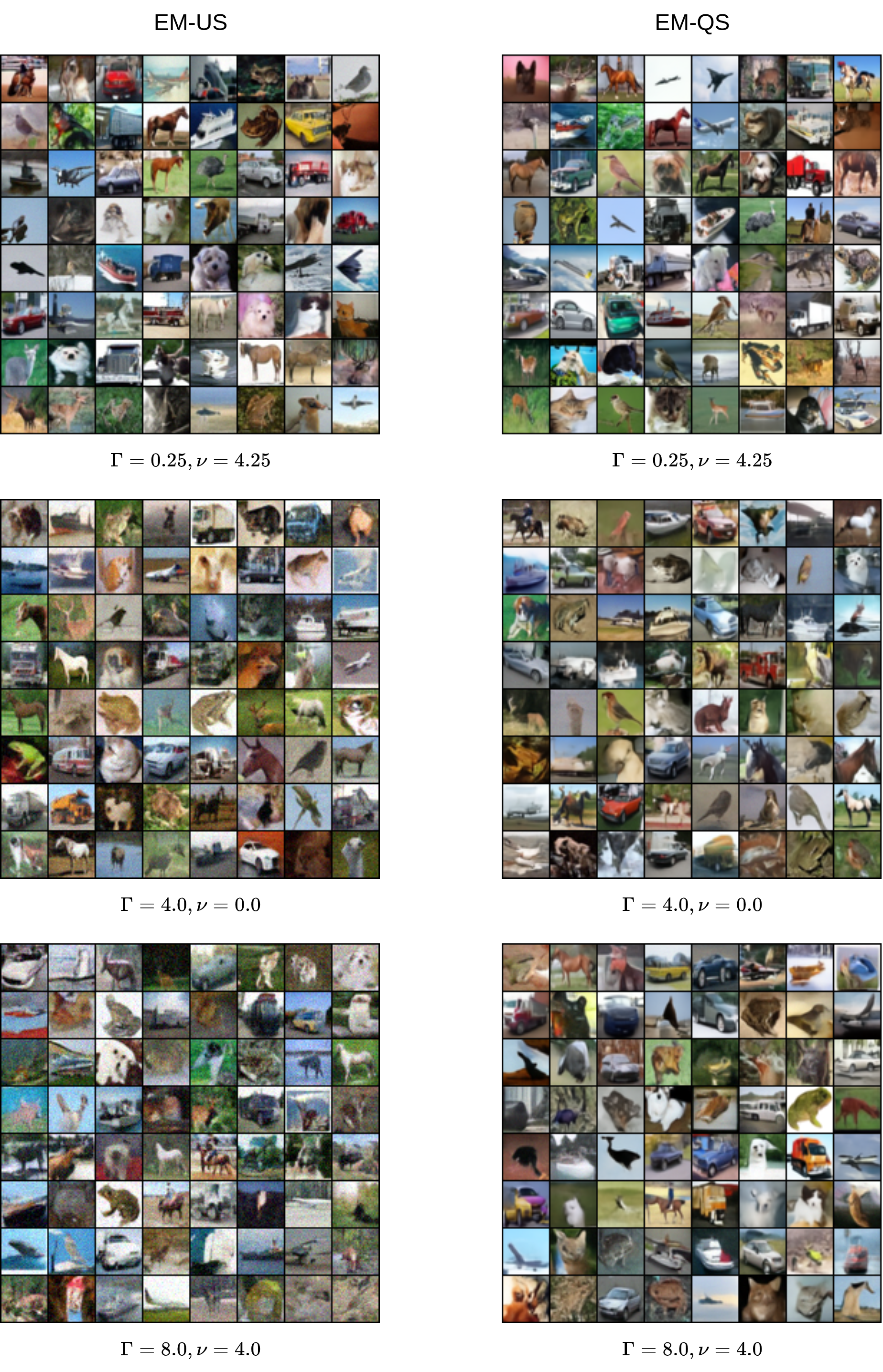}
    \caption{Qualitative illustration of the impact of $\Gamma$ on CIFAR-10 sample quality. Samples get progressively worse when increasing $\Gamma$. (Uncurated) Samples in the left and right columns were generated using EM-US and EM-QS samplers.}
    \label{fig:app_3}
\end{figure}

\begin{figure}
\centering
    \includegraphics[width=0.95\linewidth]{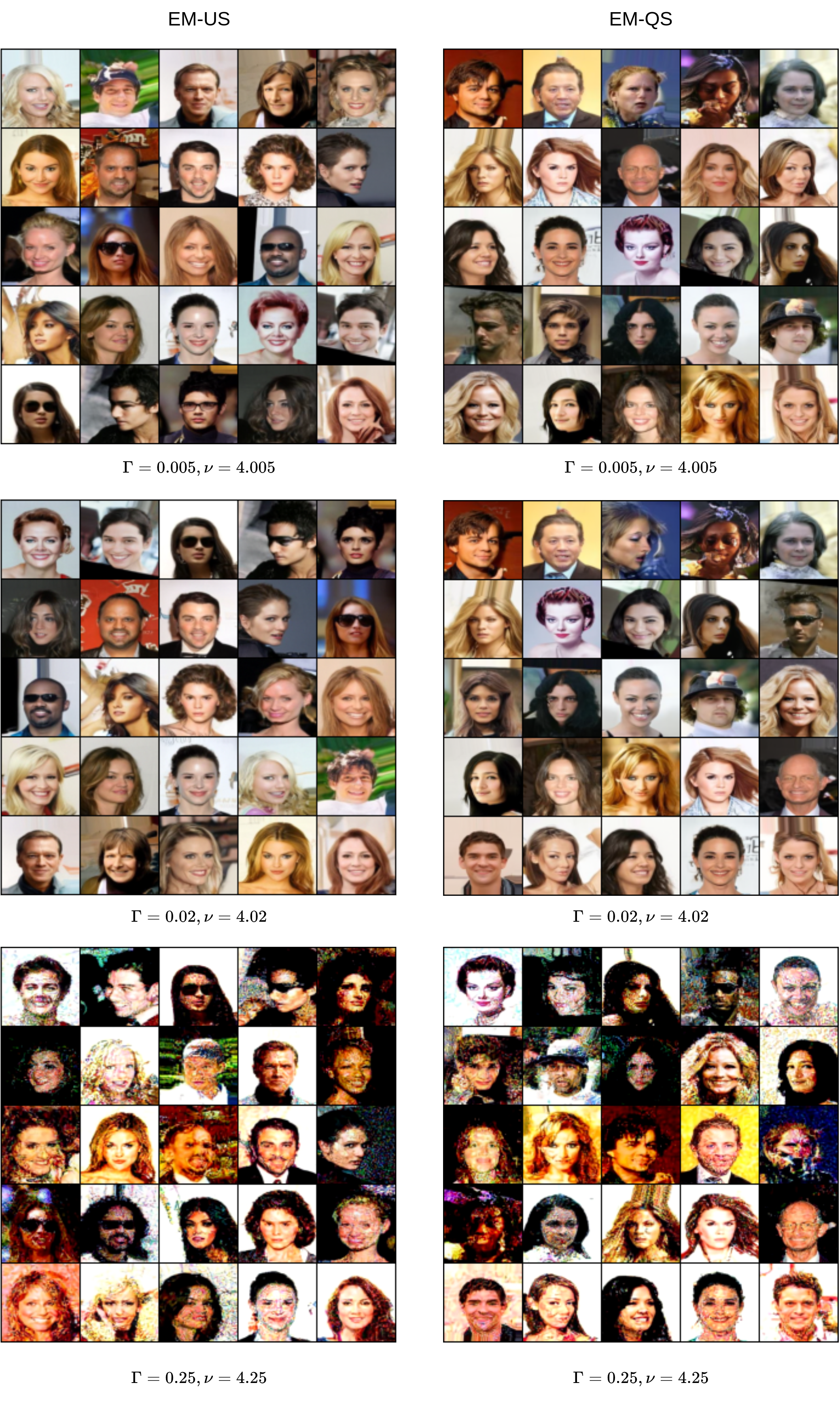}
    \caption{Qualitative illustration of the impact of $\Gamma$ on CelebA-64 sample quality. Samples get progressively worse when increasing $\Gamma$. (Uncurated) Samples in the left and right columns were generated using EM-US and EM-QS samplers.}
    \label{fig:app_4}
\end{figure}

\textbf{Case-2: Effect of using a large $\Gamma$}: Figure \ref{fig:app_1}b illustrates the variation of $|\lambda_2|$ for some more values of $\Gamma$. Interestingly for $\Gamma=4.0$, the value of $|\lambda_2|$ decreases to almost 0 in the low-timestep regime. However, for $\Gamma=4.0$, the value of $|\lambda_1|$ increases significantly (Figure \ref{fig:app_2}a), therefore, leading to large error scaling factors in the term $|\lambda_1| \bm{\eps}_{\theta}^x(\brvz_t)$. This finding justifies our observation in Figure 2, where a value of $\Gamma=0.25$ makes sample quality significantly worse for the CelebA-64 dataset and is unable to recover high-frequency details. Figure \ref{fig:app_2}b further illustrates the variation of $|\lambda_1|$ for different $(\Gamma, \nu)$ pairs in the low-timestep regime.

From the above analysis, it seems that \textit{the choice of $\Gamma$ provides an important trade-off between balancing the errors produced due to the terms $\bm{\eps}_{\theta}^x(\brvz_t)$ and $\bm{\eps}_{\theta}^m(\brvz_t)$} in Eqn. \ref{eqn:em_4}. Therefore, the choice of $\Gamma$ is crucial for sample quality in \nsmn.

\begingroup
\setlength{\tabcolsep}{12pt}
\begin{table}[t]
\centering
\footnotesize
\begin{tabular}{@{}ccccccc@{}}
\toprule
                         &                             & \multicolumn{5}{c}{NFE (FID@10k $\downarrow$)}                                  \\ \midrule
Sampler                  & Method                      & 50             & 100            & 250           & 500           & 1000          \\ \midrule
\multirow{4}{*}{EM-QS}   & CLD                         & 25.01          & 8.91           & 5.97          & 5.61          & 5.7           \\
                         & VP-SDE                      & \textbf{17.72} & 7.45           & 5.59          & 5.51          & 5.51          \\
                         & (Ours) PSLD ($\Gamma=0.01$) & 23.96          & 8.12           & 5.41          & \textbf{5.13} & \textbf{5.24} \\
                         & (Ours) PSLD ($\Gamma=0.02$) & 19.94          & \textbf{7.33}  & \textbf{5.26} & 5.20          & 5.28          \\ \midrule
\multirow{4}{*}{EM-US}   & CLD                         & 119.68         & 45.60          & 9.08          & 5.71          & 5.65          \\
                         & VP-SDE                      & \textbf{84.54} & 41.93          & 12.61         & 5.92          & 5.19          \\
                         & (Ours) PSLD ($\Gamma=0.01$) & 109.01         & 40.22          & \textbf{9.07} & \textbf{5.25} & 4.95          \\
                         & (Ours) PSLD ($\Gamma=0.02$) & 100.62         & \textbf{39.96} & 11.26         & 5.45          & \textbf{4.82} \\ \midrule
\multirow{3}{*}{SSCS-QS} & CLD                         & 21.31          & 8.37           & 5.82          & 5.75          & 5.69          \\
                         & (Ours) PSLD ($\Gamma=0.01$) & 18.41          & 7.42           & 5.41          & \textbf{5.28} & 5.29          \\
                         & (Ours) PSLD ($\Gamma=0.02$) & \textbf{16.12} & \textbf{7.16}  & \textbf{5.36} & 5.35          & \textbf{5.27} \\ \midrule
\multirow{3}{*}{SSCS-US} & CLD                         & 75.45          & 24.74          & 6.09          & 5.74          & 5.78          \\
                         & (Ours) PSLD ($\Gamma=0.01$) & 76.6           & 21.25          & \textbf{5.18} & 5.10          & 5.33          \\
                         & (Ours) PSLD ($\Gamma=0.02$) & \textbf{72.42} & \textbf{20.46} & 5.19          & \textbf{4.92} & \textbf{5.29} \\ \bottomrule 
\end{tabular}
\caption{Extended Speed vs. Sample quality comparisons using the SDE setup for CIFAR-10. FID computed for 10k samples. Values in \textbf{bold} indicate the best result for that column.}
\label{table:app_ext_cifar10}
\end{table}
\endgroup

\subsection{Additional Speed vs. Sample Quality Comparisons}
\label{app:D_2}

\begingroup
\setlength{\tabcolsep}{12pt}
\begin{table}[]
\centering
\footnotesize
\begin{tabular}{@{}ccccccc@{}}
\toprule
                          &                             & \multicolumn{5}{c}{NFE (FID@10k $\downarrow$)}                                                 \\ \midrule
Sampler                   & Method                      & 50             & 100            & 250           & 500           & 1000                         \\ \midrule
                          & CLD                         & 73.61          & 14.78          & 4.77          & 4.48          & 4.59                         \\
\multirow{-2}{*}{EM-QS}   & (Ours) PSLD ($\Gamma=0.005$) & \textbf{44.36} & \textbf{6.99}  & \textbf{3.77} & \textbf{3.92} & \textbf{4.17}                \\ \midrule
                          & CLD                         & 122.63         & 54.67          & 11.66         & 4.97          & 4.6                          \\
\multirow{-2}{*}{EM-US}   & (Ours) PSLD ($\Gamma=0.005$) & \textbf{99.05} & \textbf{44.06} & \textbf{8.9}  & \textbf{4.42} & \textbf{4.37}                \\ \midrule
                          & CLD                         & 44.83          & 10.7           & 4.82          & 4.73          & 4.74 \\
\multirow{-2}{*}{SSCS-QS} & (Ours) PSLD ($\Gamma=0.005$) & \textbf{34.3}  & \textbf{8.13}  & \textbf{4.16} & \textbf{4.11} & \textbf{4.09}                \\ \midrule
                          & CLD                         & 105.16         & 45.54          & 6.75          & \textbf{4.06} & 4.18                         \\
\multirow{-2}{*}{SSCS-US} & (Ours) PSLD ($\Gamma=0.005$) & \textbf{97.8}  & \textbf{35.59} & \textbf{4.65} & 4.08          & \textbf{4.05}    \\ \bottomrule           
\end{tabular}
\caption{Speed vs. Sample Quality comparison using the SDE setup for CelebA-64. FID computed for 10k samples. Values in \textbf{bold} indicate the best result for that column.}
\label{table:app_5}
\end{table}
\endgroup

\begin{table}[h]
\begin{minipage}{0.49\linewidth}
\scriptsize
\centering
\begin{tabular}{@{}ccccc@{}}
\toprule
Model                     & Size & NFE  & FID $\downarrow$  & IS $\uparrow$   \\ \midrule
\mn ($\Gamma$=0.01)       & 55M        & 1000 & 2.34 & 9.57 \\
\mn ($\Gamma$=0.02)       & 55M        & 1000 & 2.3  & 9.68 \\
\mn ($\Gamma$=0.01, deep) & 97M        & 1000 & 2.26 & 9.71 \\
\mn ($\Gamma$=0.02, deep) & 97M        & 1000 & \textbf{2.21} & \textbf{9.74} \\ \bottomrule
\end{tabular}
\caption{CIFAR-10 sample quality (SDE). FID (lower is better) and IS (higher is better) were computed on 50k samples.}
\label{table:app_6}
\end{minipage}
\hfill
\begin{minipage}{0.49\linewidth}
\scriptsize
\centering
\begin{tabular}{@{}ccccc@{}}
\toprule
Model                     & Size & NFE & FID $\downarrow$  & IS $\uparrow$   \\ \midrule
\mn ($\Gamma$=0.01)       & 55M        & 243 & 2.41 & 9.63 \\
\mn ($\Gamma$=0.02)       & 55M        & 232 & 2.4  & 9.84 \\
\mn ($\Gamma$=0.01, deep) & 97M        & 246 & \textbf{2.10} & 9.79 \\
\mn ($\Gamma$=0.02, deep) & 97M        & 231 & 2.31 & 9.91 \\
\mn ($\Gamma$=0.01, deep) & 97M        & 159 & 2.13 & 9.76 \\
\mn ($\Gamma$=0.02, deep) & 97M        & 159 & 2.34 & \textbf{9.93} \\ \bottomrule
\end{tabular}
\caption{CIFAR-10 sample quality (ODE). FID (lower is better) and IS (higher is better) were computed on 50k samples.}
\label{table:app_4}
\end{minipage}
\end{table}

\textbf{CIFAR-10}: We extend the Speed vs. Sample Quality results in Table 5 to include results for PSLD with $\Gamma=0.01$ in Table \ref{table:app_ext_cifar10}\\

\textbf{CelebA-64}:
Similar to our setup for CIFAR-10 (Section 4.3), we benchmark the speed vs quality tradeoffs of \mn ($\Gamma=0.005$) against our CLD ablation baseline for the CelebA-64 dataset (See Table \ref{table:app_5}). Similar to CIFAR-10, \mn outperforms our CLD baseline across all timesteps. The performance difference is most notable in the low-timestep regime (FID of 6.99 for \mn with EM-QS vs 10.7 for CLD with SSCS-QS). However, there are two notable differences in our observations when compared to CIFAR-10: 

\textbf{(i)} Firstly, quadratic striding works best for the CelebA-64 dataset. This contrasts with CIFAR-10, where a uniform striding schedule works better.

\textbf{(ii)} More interestingly, \mn achieves the best performance of FID=3.77 at N=250 steps, and sample quality degrades on further increasing the number of steps. This contrasts our results for CIFAR-10, where \mn achieves the best performance at T=1000.

\subsection{Extended SOTA Results}
\label{app:D_3}

\begin{figure}
\centering
    \includegraphics[width=1.0\linewidth]{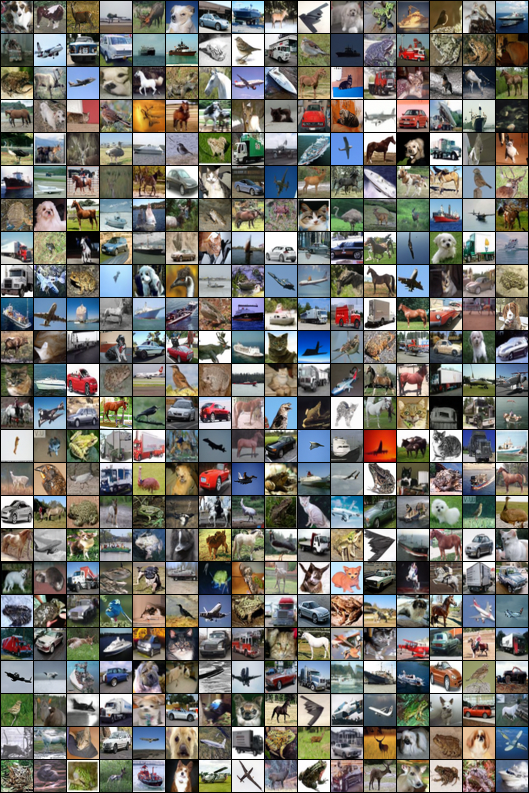}
    \caption{Uncurated samples from our SOTA \mn ($\Gamma=0.02, \nu=4.02$) model using SDE sampling (FID=2.21, NFE=1000)}
    \label{fig:app_5}
\end{figure}

\begin{figure}
\centering
    \includegraphics[width=1.0\linewidth]{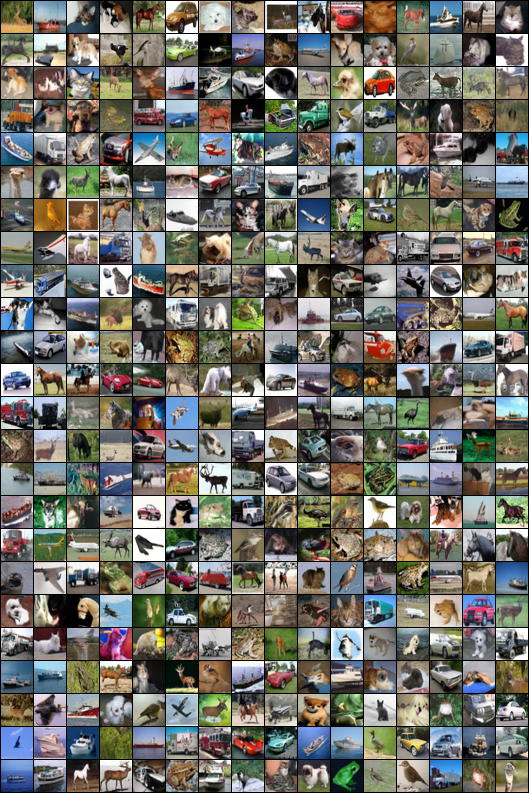}
    \caption{Uncurated samples from our SOTA \mn ($\Gamma=0.01, \nu=4.01$) model using ODE sampling (FID=2.10, NFE=246)}
    \label{fig:app_6}
\end{figure}

\begin{figure}
\centering
    \includegraphics[width=1.0\linewidth]{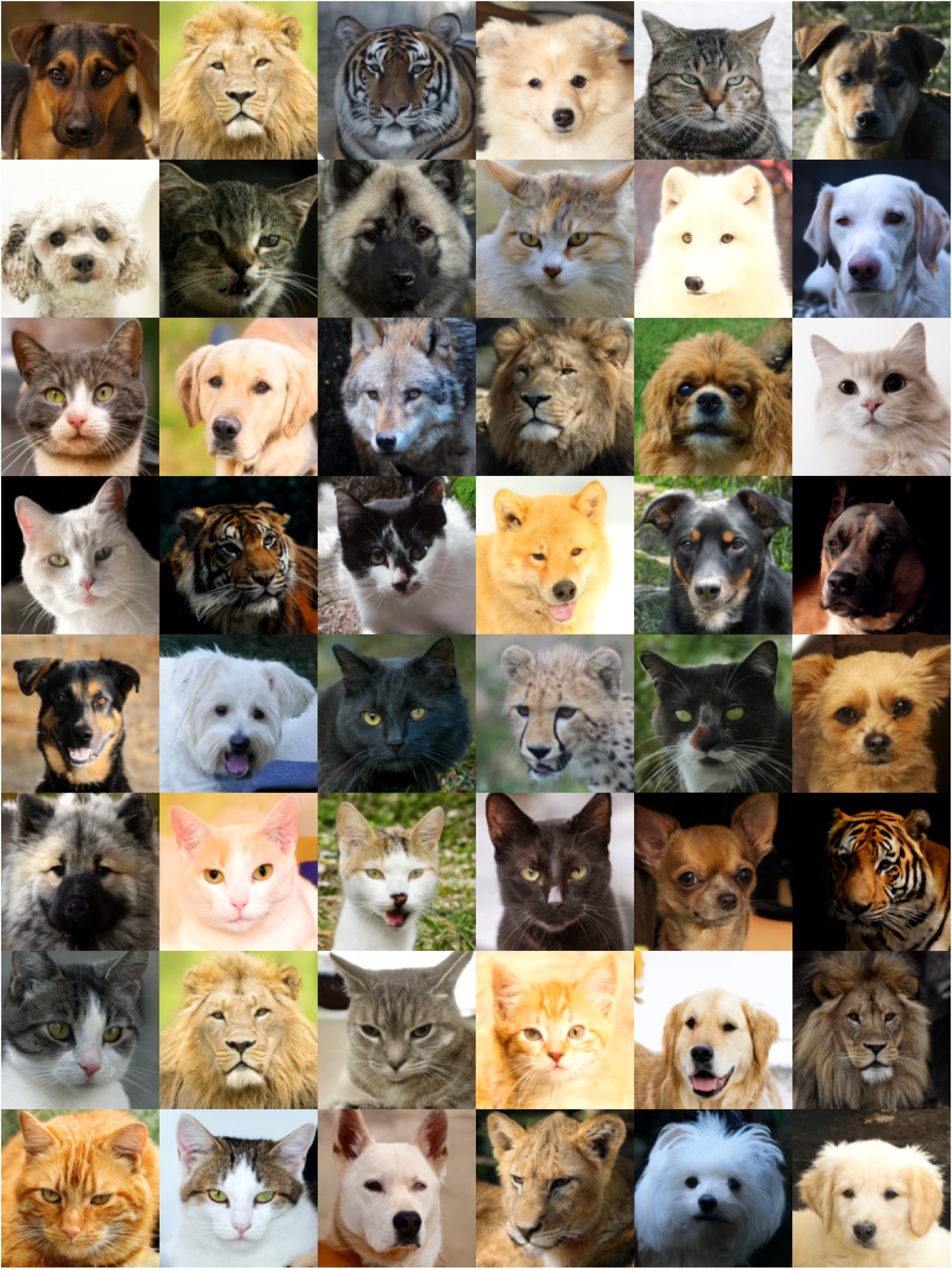}
    \caption{Random unconditional AFHQv2 samples at 128x128 resolution from our \mn ($\Gamma=0.01$) model using the EM-QS sampler with N=1000.}
    \label{fig:app_7}
\end{figure}

\textbf{Extended Qualitative Results}: We provide qualitative samples from our SOTA CIFAR-10 models using the SDE and ODE setups in Figures \ref{fig:app_5} and \ref{fig:app_6} respectively. We provide some additional samples from the AFHQv2 dataset at the 128 x 128 resolution in Figure \ref{fig:app_7}.

\textbf{Extended Quantitative Results}:
Table \ref{table:app_6} shows the FID and IS scores for all models using the SDE sampling setup. \mn with $\Gamma=0.02$ attains the best IS score of 9.74. When using the ODE sampling setup (Table \ref{table:app_4}), \mn with $\Gamma=0.02$ achieves the best IS score of 9.93.

\subsection{Conditional Synthesis using PSLD}
\label{app:D_4}

\begin{figure}
\centering
    \includegraphics[width=1.0\linewidth]{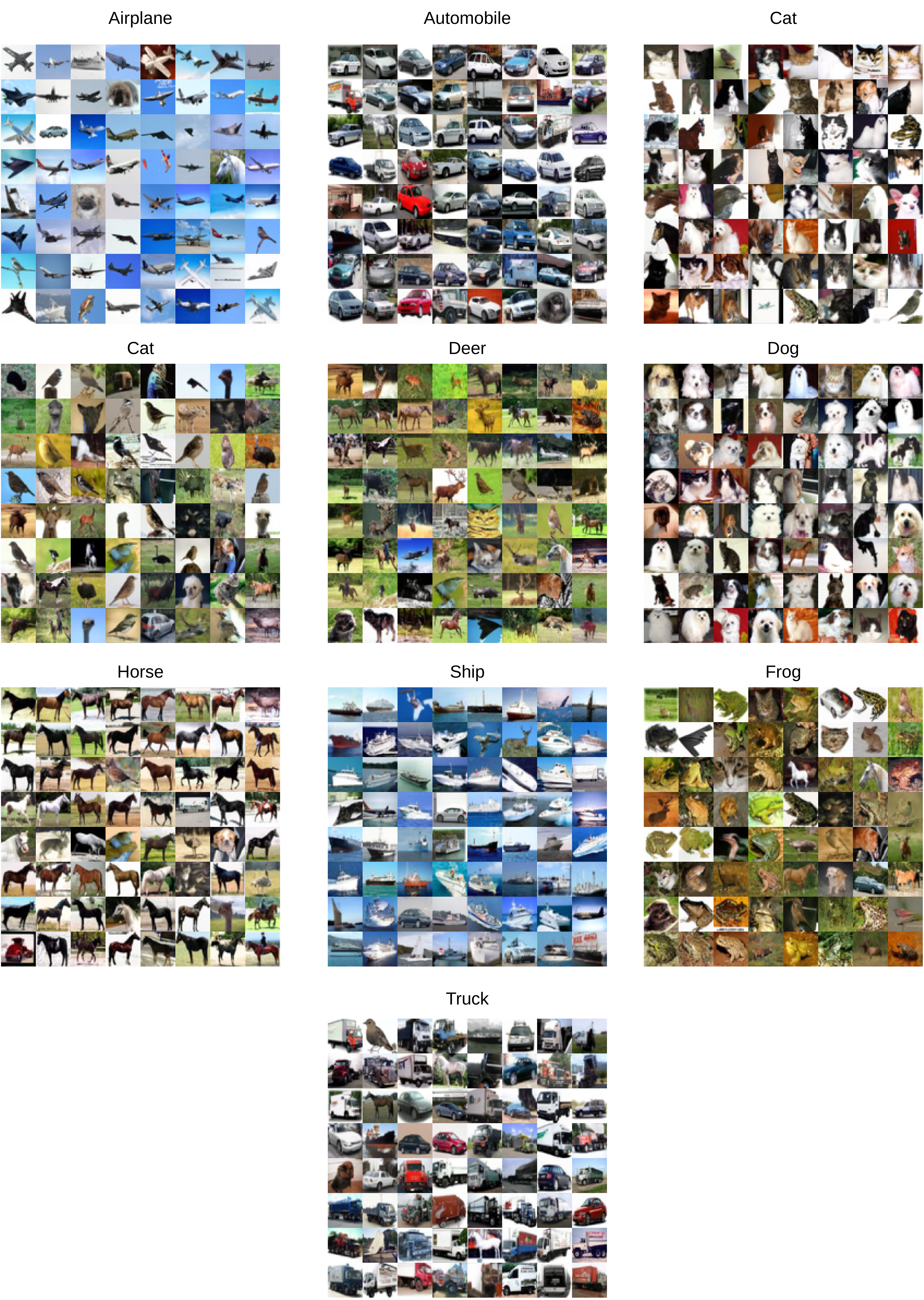}
    \caption{Randomly sampled class conditional results on the CIFAR-10 dataset using the EM-US sampler (N=1000). Guidance weight $\lambda=5.0$. Using large guidance weight reduces diversity but improves sample quality.}
    \label{fig:app_10}
\end{figure}

\begin{figure}
\centering
    \includegraphics[width=0.95\linewidth]{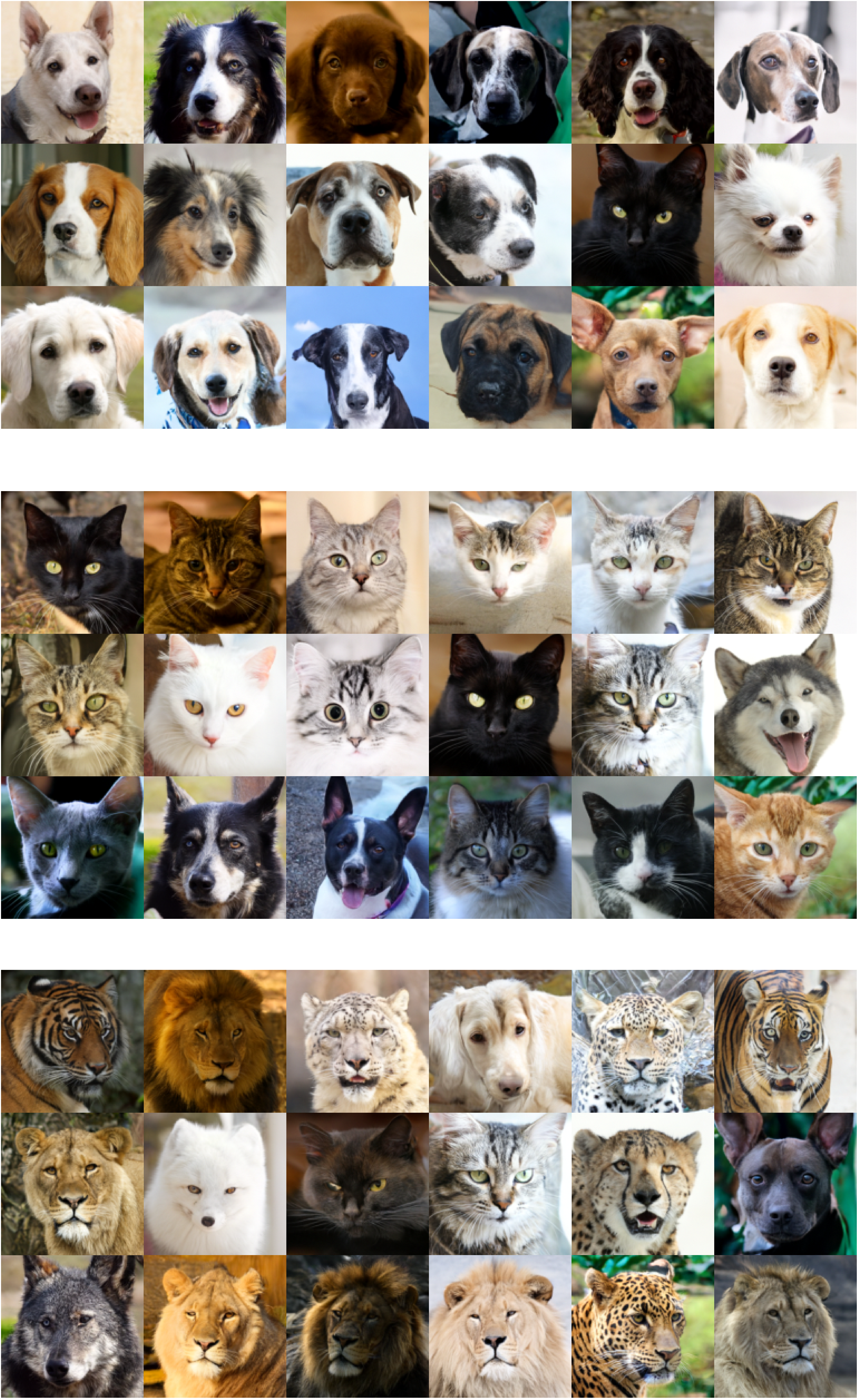}
    \caption{Randomly sampled class conditional results on the AFHQv2 dataset using the EM-US sampler (N=1000). Guidance weight $\lambda=10.0$. (Top to Bottom) Each of the three rows correspond to \textit{Dog}, \textit{Cats} and \textit{Others}.}
    \label{fig:app_9}
\end{figure}

\textbf{Class-Conditional Synthesis}: As discussed in Section 4.4, given class label information $\rvy$, an unconditional pre-trained score network $\vs_{\theta}(\rvz_t, t)$ can be used for sampling from the class conditional distribution $p(\rvz_t|\rvy)$ in \nsmn. More specifically, we need to simulate the following reverse SDE:
\begin{equation}
    d\rvz_t = \left[\vf(\rvz_t) - \mG(t)\mG(t)^T \nabla_{\rvz_t} \log{p(\rvz_t|\rvy)}\right]dt  + \mG(t)d\rvw_t
    \label{eqn:cc_controllable}
\end{equation}
The \textit{conditional} score $\nabla_{\rvz_t} \log{p(\rvz_t|\rvy)}$ can be further decomposed as follows:
\begin{align}
    \nabla_{\rvz_t} \log{p(\rvz_t|\rvy)} &= \nabla_{\rvz_t} \log{p(\rvy|\rvz_t)} + \nabla_{\rvz_t} \log{p(\rvz_t)}\\
    &\approx \underbrace{\nabla_{\rvz_t} \log{p(\rvy|\rvz_t)}}_{\text{Classifier Gradient}} + \underbrace{\vs_{\theta}(\rvz_t, t)}_{\text{Score}}
    \label{eqn:cc_guidance}
\end{align}
For practical scenarios, it is common to scale the contribution of the classifier gradient by a factor of $\lambda > 1$. Thus,
\begin{equation}
    \nabla_{\rvz_t} \log{p(\rvz_t|\rvy)} = \lambda\nabla_{\rvz_t} \log{p(\rvy|\rvz_t)} + \vs_{\theta}(\rvz_t, t)
\end{equation}
The above technique of approximating the conditional score $\nabla_{\rvz_t} \log{p(\rvz_t|\rvy)}$ is called as \textit{classifier-guidance} \citep{dhariwal2021diffusion, songscore}. The classifier $p(\rvy|\rvz_t)$ is trained by minimizing a time-dependent cross-entropy loss as follows:
\begin{equation}
    \mathcal{L}_{\text{clf}}(\phi) = \mathbb{E}_{t \sim \gU(0,1)}\mathbb{E}_{\rvx_0, \rvy \sim p_{\text{data}(\rvx_0, \rvy)}}\mathbb{E}_{\rvz_t \sim p(\rvz_t|\rvx_0)}\left[-\sum_k \mathds{1}(y=y_k)\log{C^k_{\phi}(\rvz_t, t)}\right]
\end{equation}
where $C^k_{\phi}(\rvz_t, t)$ is a time-dependent classifier that takes as input a perturbed data point $\rvz_t$ and outputs class prediction probabilities. We perform class conditional synthesis for the CIFAR-10 (10 classes) and the AFHQ-v2 datasets. For the AFHQ-v2 dataset, we use the classes \textit{Cats}, \textit{Dogs}, and \textit{Others} from the train split for classifier training (See Appendix \ref{app:C_6} for complete implementation details). We provide additional class conditional samples for CIFAR-10 in Figure \ref{fig:app_10} and for AFHQ-v2 in Figure \ref{fig:app_9}.

\begin{figure}
\centering
    \includegraphics[width=1.0\linewidth]{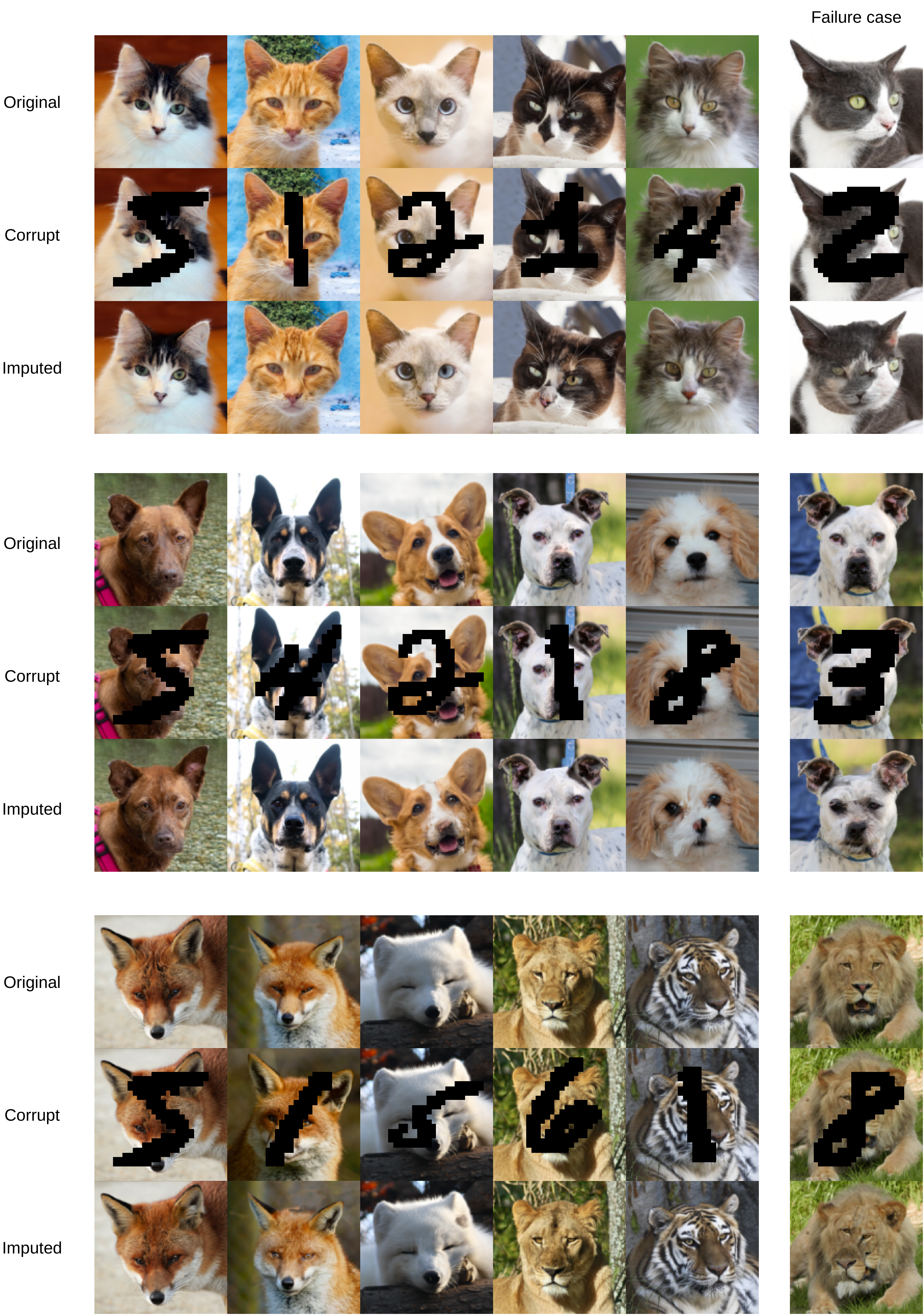}
    \caption{Additional imputation results on the AFHQv2 dataset (test split) using the EM-US sampler (N=1000). The Rightmost column indicates some failure cases.}
    \label{fig:app_8}
\end{figure}

\textbf{Image Inpainting}: 
Following \citep{songscore}, we can partition the input $\rvx_0$ into known ($\hat{\rvx}_0$) and unknown ($\bar{\rvx}_0$) components respectively. We can now define the diffusion for the unknown component in the augmented space as follows:
\begin{equation}
    d \bar{\rvz}_t = \bar{\rvf}(\rvz_t)dt + \bar{\mG}(t)d\rvw_t
\end{equation}
where $\bar{\rvf}(\rvz_t) = \rvf(\bar{\rvz}_t)$ i.e. the drift applied to the missing components of $\rvz_t$. Similarly, $\bar{\mG}(t)$ corresponds to the diffusion coefficient applied to the corresponding components of Brownian motion $d\rvw_t$. The corresponding reverse-SDE (conditioned on the observed signal $\hat{\rvx}_0$) can be specified as:
\begin{equation}
    d \bar{\rvz}_t = \left[\bar{\rvf}(\rvz_t) - \bar{\mG}(t)\bar{\mG}^T(t)\nabla_{\bar{\rvz}_t}\log{p(\bar{\rvz}_t|\hat{\rvx}_0)}\right]dt + \bar{\mG}(t)d\rvw_t
\end{equation}
Following the derivation in \citep{songscore}, it can be shown that:
\begin{align}
    \nabla_{\bar{\rvz}_t}\log{p(\bar{\rvz}_t|\hat{\rvx}_0)} &\approx \nabla_{\bar{\rvz}_t}\log{p(\bar{\rvz}_t | \hat{\rvz}_t)} \\
    &= \nabla_{\bar{\rvz}_t}\log{p([\bar{\rvz}_t ;\hat{\rvz}_t])}
\end{align}
where $\hat{\rvz}_t \sim p(\hat{\rvz}_t|\hat{\rvx}_0)$ is a noisy augmented state sampled from the perturbation kernel given an observed signal $\hat{\rvx}_0$. We provide additional imputation results in Figure \ref{fig:app_8}

\textbf{General Inverse Problems}: Similar to imputation, we can utilize PSLD for solving general inverse problems. Given a conditioning signal $\rvy$, we have,
\begin{align}
    \nabla_{\rvz_t} \log p_t(\rvz_t \mid \rvy) = \nabla_{\rvz_t} \log \int p_t(\rvz_t \mid \rvy_t, \rvy) p(\rvy_t \mid \rvy) d \rvy_t,
\end{align}
Further assuming that $p(\rvy_t \mid \rvy)$ is tractable and $p_t(\rvz_t \mid \rvy_t, \rvy) \approx p_t(\rvz_t \mid \rvy_t)$, we have
\begin{align}
    \nabla_{\rvz_t} \log p_t(\rvz_t \mid \rvy) &\approx \nabla_{\rvz_t} \log \int p_t(\rvz_t \mid \rvy_t) p(\rvy_t \mid \rvy) d \rvy_t\\
    &\approx \nabla_{\rvz_t} \log p_t(\rvz_t \mid \hat{\rvy}_t)\\
    &= \nabla_{\rvz_t} \log p_t(\rvz_t) + \nabla_{\rvz_t} \log p_t(\hat{\rvy}_t \mid \rvz_t)\\
    &\approx \vs_{\vtheta^*}(\rvz_t, t) + \nabla_{\rvz_t} \log p_t(\hat{\rvy}_t \mid \rvz_t), \label{eqn:cond_approx}
\end{align}
where $\hat{\rvy}_t \sim p(\rvy_t \mid \rvy)$. Thus \mn can be used for conditional synthesis like previous SGMs \citep{songscore} while achieving better speed-vs-quality tradeoffs and better overall sample quality. Therefore, \mn provides an attractive baseline for further developments in SGMs.

